\documentclass[journal]{IEEEtran}
\usepackage{amsmath,amssymb,bm}
\usepackage{array}

\usepackage{graphicx}
\usepackage[caption=false,font=footnotesize]{subfig} % IEEEtran推荐使用subfig
\usepackage{booktabs}      % 专业表格线
\usepackage{multirow}      % 表格多行
\usepackage{flafter}       % 浮动体控制

\usepackage{algorithm}
\usepackage{algpseudocode}

\usepackage{xcolor}
\usepackage{url}
\usepackage{adjustbox}
\usepackage{dcolumn}

\usepackage{cite}
\begin{document}

\title{JOGS: Joint Optimization of Pose Estimation and 3D Gaussian Splatting}

\author{Xianben Yang$^{\ast}$\IEEEmembership{},
Yuxuan Li$^{\ast}$\IEEEmembership{},
Tao Wang\IEEEmembership{},
Yi Jin\IEEEmembership{},
 Yidong Li \IEEEmembership{}and
  Haibin Ling \IEEEmembership{}

\thanks{$^{\ast}$Xianben Yang and Yuxuan Li contributed equally to this work.}
% \thanks{Xianben Yang, Yuxuan Li, Tao Wang, Yi Jin and Yidong Li are with the Key Laboratory of Big Data \& Artificial Intelligence in Transportation (Beijing Jiaotong University), Ministry of Education, China, and the School of Computer Science and Technology, Beijing Jiaotong University, Beijing 100044, China. Haibin Ling is a Chair Professor in the Department of Artificial Intelligence at Westlake University, Hangzhou 310030, China. Corresponding author: Tao Wang (email: twang@bjtu.edu.cn).}}
\thanks{Xianben Yang, Yuxuan Li, Tao Wang, Yi Jin and Yidong Li are with the Key Laboratory of Big Data \& Artificial Intelligence in Transportation (Beijing Jiaotong University), Ministry of Education, China, and also with the School of Computer Science and Technology, Beijing Jiaotong University, Beijing 100044, China. }
\thanks{Haibin Ling is with the Department of Artificial Intelligence, Westlake University, Hangzhou 310030, China.}
\thanks{Corresponding author: Tao Wang(e-mail: twang@bjtu.edu.cn).}
}

\markboth{Journal of \LaTeX\ Class Files,~Vol.~14, No.~8, August~2021}%
{Shell \MakeLowercase{\textit{et al.}}: A Sample Article Using IEEEtran.cls for IEEE Journals}


\maketitle

\begin{abstract}
Traditional novel view synthesis methods heavily rely on external camera pose estimation tools such as COLMAP, which often introduce computational bottlenecks and propagate errors.  To address these challenges, we propose a unified framework that jointly optimizes 3D Gaussian points and camera poses without requiring pre-calibrated inputs.  Our approach iteratively refines 3D Gaussian parameters and updates camera poses through a novel co-optimization strategy, ensuring simultaneous improvements in scene reconstruction fidelity and pose estimation accuracy.
The key innovation lies in decoupling the joint optimization into two interleaved phases: first, updating 3D Gaussian parameters via differentiable rendering with fixed poses, and second, refining camera poses using a customized 3D optical flow algorithm that incorporates geometric and photometric constraints.  This formulation progressively reduces projection errors, particularly in challenging scenarios with large viewpoint variations and sparse feature distributions, where traditional methods struggle.  Extensive evaluations on multiple datasets demonstrate that our approach significantly outperforms existing COLMAP-free techniques in reconstruction quality, and also surpasses the standard COLMAP-based baseline in general.
\end{abstract}


\section{Introduction}
\IEEEPARstart{R}{ecent} advancements in the field of computer vision have led to significant progress in 3D scene reconstruction and rendering. In particular, the introduction of \textit{3D Gaussian Splatting }(3DGS)~\cite{10.1145/3592433} technology has provided an efficient and realistic technique for scene representation and rendering. 3DGS explicitly models the scene using a group of Gaussian ellipsoids. This provides rapid and accurate rendering, clearly exhibiting its benefits in real-time situations. Due to its explicit representation and efficient rendering capabilities, 3DGS has been widely applied in various fields~\cite{11247877,11275889,10964679,10636772,11194205,11329503}, especially in scenarios requiring efficient processing and realistic rendering~\cite{wu2024recentadvances3dgaussian}. 

Accurate pose estimation is extremely important~\cite{Fu_2024_CVPR} for most novel view synthesis methods including 3DGS. Most existing 3DGS methods do not include a pose estimation component, but rely on external inputs (\textit{e.g.}, COLMAP~\cite{schoenberger2016sfm,schoenberger2016mvs}).
The separation of pose estimation and 3DGS optimization may lead to suboptimal solutions. On the other hand, the reliance on external input may limit its application to certain scenarios~\cite{bian2022nopenerf,Fu_2024_CVPR}.
To solve these issues, recent research suggests many 3DGS solutions that do not require inputs of camera poses.   For example, CFGS~\cite{Fu_2024_CVPR} uses a combined optimization of camera parameters and Gaussian points. This method transforms the camera pose registration problem into an image optimization task between two consecutive frames. It achieves excellent reconstruction results in continuous and dense image streams, but its performance degradations when the constraints are violated. ZeroGS~\cite{chen2024zerogstraining3dgaussian} and InstantSplat~\cite{fan2024instantsplat} do not require pre-supplied camera poses, but they need to load a pre-trained model in advance for pose estimation, and usually work in very sparse views.
These methods either require the integration of additional information or pre-trained models, or can only operate with strong constraints on the input images, which substantially limits their applicability.

In this paper, to address these challenges, we propose \textbf{JOGS}  that \textbf{J}ointly \textbf{O}ptimizes the camera poses and the 3D  \textbf{G}aussian \textbf{S}platting representations without requiring pre-calibrated inputs.  In addition to the losses used in the standard 3DGS algorithm~\cite{10.1145/3592433}, our framework introduces a reprojection loss to penalize the inconsistencies between different views. 
After an initial coarse pipeline setup, the camera poses are subsequently optimized in conjunction with the 3DGS parameters using the \textit{alternating direction method} (ADM) algorithm~\cite{8186925}.
In each iteration, the 3DGS parameters are updated following the standard 3DGS algorithm. For the refinement of camera poses, we propose a \textit{lucas-kanade 3D optical flow} (LK3D) algorithm, which leverages Gaussian points and image reprojection errors by integrating image gradients with transformation-based projection error relationships. This alternating optimization strategy significantly improves pose accuracy and achieves stable convergence even under large camera viewpoint movements or sparse feature distributions. 

For validation, we compare the proposed method with several state-of-the-art methods on three public datasets, including Tanks and Temples~\cite{Knapitsch2017}, LLFF-NeRF~\cite{mildenhall2019llff} and 
Shiny~\cite{Wizadwongsa2021NeX}. The experimental results show that our method outperforms the baselines in novel view synthesis, and achieves high reconstruction quality in different scenarios. The source code of our method will be released upon paper publication.

In summary, we make the following contributions:

(1) we propose a unified framework for joint optimization of 3DGS parameters and camera poses, which does not rely on external tools such as COLMAP;  

(2) we propose an LK3D algorithm to optimize camera poses based on the reprojection errors between 3D Gaussian points and image pixels, which is independent of the sequential relationship between images and is able to effectively fine-tune the camera pose; and

(3) we validate the effectiveness of our method in different datasets, which exhibits robust reconstruction quality across all scenarios.

\section{Related Work}
\label{sec:related}

\begin{figure*}[t!]
    \centering
    \captionsetup{font=small}
    \includegraphics[width=0.98\textwidth]{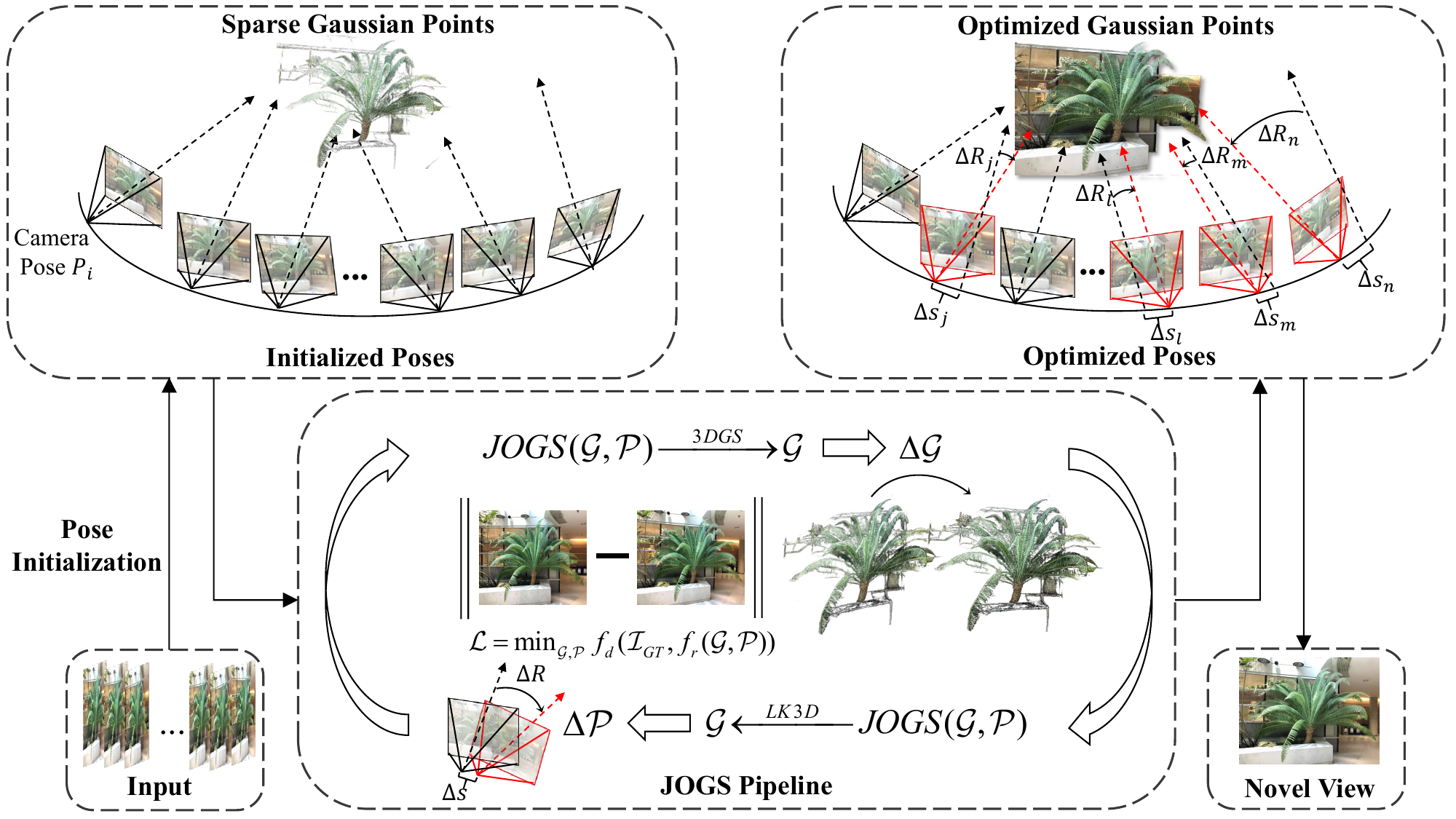}
    \caption{\textbf{Method Overview.} Our JOGS framework jointly optimizes Pose Estimation and 3D Gaussian Splatting. It starts with a simple SfM initialization, then iteratively updates 3D Gaussian splatting parameters $\mathcal{G}$ and refines camera poses $\mathcal{P}$, ensuring simultaneous improvements in scene reconstruction fidelity and pose estimation accuracy. The updating of Gaussian points follows a standard 3DGS pipeline, while the refinement of camera poses is done by the proposed LK3D algorithm.}
    \label{fig:overview}
    \vspace{-2mm}
\end{figure*}

\subsection{Novel View Synthesis}
\label{subsec:Novel View Synthesis}
The task aims to generate photorealistic renderings of target scenes from unknown viewpoints using a limited set of input images.  
Recent advancements in neural rendering~\cite{10.1145/3503250,10229247,10856404,11129026} have significantly improved NVS in terms of reconstruction quality and efficiency. The seminal work on \textit{Neural Radiance Fields} (NeRF)~\cite{10.1145/3503250} introduced a paradigm shift in NVS by representing scenes as continuous implicit neural radiance fields, encoded via \textit{multilayer perceptrons} (MLPs). Subsequent studies extended NeRF along several directions. For example, Mip-NeRF 360~\cite{barron2022mipnerf360} improve core rendering fidelity and scene representation. Some works~\cite{park2021nerfies,li2020neural} enhanced dynamic scene modeling. Some methods~\cite{garbin2021fastnerf,SunSC22} optimized computational efficiency to accelerate training. However, NeRF continues to face challenges, including prolonged training times, high hardware demands and limited editability. 
Recently, the emergence of 3DGS~\cite{10.1145/3592433} has achieved breakthroughs by utilizing explicit differentiable representations, striking a balance between rendering quality and efficiency. Extensive research has been conducted on 3DGS, covering areas such as scene rendering quality and realism~\cite{Yu2023MipSplatting,scaffoldgs}, 3DGS acceleration~\cite{radl2024stopthepop}, geometry reconstruction~\cite{liu2024atomgs,Huang2DGS2024}, dynamic scenes~\cite{Huang_2024_CVPR} and few-shot reconstruction~\cite{Vista3D,peng2024neurips}. Nevertheless, most existing methods still rely on camera poses and sparse point clouds precomputed by COLMAP~\cite{schoenberger2016mvs,schoenberger2016sfm}.

\subsection{NVS without Pose Input}
\label{subsec:NVS without Pose Input}
Eliminating the dependence of input pose has become a main topic in recent research of NVS, for both NeRF and 3DGS methods. I-NeRF~\cite{9636708} introduced inverse rendering to estimate camera poses through keypoint alignment using pre-trained NeRF. BARF~\cite{lin2021barf} proposed a coarse-to-fine coordinate encoding strategy, with further improvement in GARF~\cite{chng2022gaussian,ramasinghe2022beyond}. Nope-NeRF~\cite{bian2022nopenerf} trained NeRF by incorporating undistorted depth priors. 
For 3DGS-based methods, CFGS~\cite{Fu_2024_CVPR} is the most closely related to our work. It builds the entire 3D Gaussian in a continuous fashion, "growing" some Gaussian points with each new view added. It optimizes the camera pose by minimizing the photometric loss between the rendered image and the next frame image. While it achieves 3DGS scene representation without relying on COLMAP, its optimization depends on the temporal relationship between adjacent images, and the change of view angles between consecutive frames needs to be small. 
ZeroGS~\cite{chen2024zerogstraining3dgaussian} relies on a pre-trained DUSt3R-based~\cite{dust3r_cvpr24} model called Spann3R~\cite{wang20243d}. 
InstantSplat~\cite{fan2024instantsplat} implements camera-free pose reconstruction in sparse views. 
While GSHT~\cite{ji2024sfmfree3dgaussiansplatting} achieves quality enhancement over CFGS, this improvement is constrained by the inherent reliance of the method on the temporal ordering within image sequences.
In summary, current mainstream methods either require the integration of additional information or pre-trained models~\cite{chen2024zerogstraining3dgaussian,dust3r_cvpr24}, hence limited to working with only a small number of images due to high computational resource~\cite{lin2021barf,bian2022nopenerf,fan2024instantsplat}, or assume minimal camera motion~\cite{bian2022nopenerf,Fu_2024_CVPR,ji2024sfmfree3dgaussiansplatting}. To overcome these limitations, we design a new framework that jointly optimizes 3D Gaussian and camera pose.

\section{Method}
\label{sec:method}
\subsection{Problem Definition}
\label{subsec:Problem definition}

Let $\mathcal{I} = \{I_1, \ldots, I_n\}$ be a set of $n$ images from different viewpoints, $\mathcal{G} = \{g_1, \ldots, g_k\}$ be 3D Gaussian points consisting of $k$ points, and $\mathcal{P} = \{P_1, \ldots, P_n\}$ denote the pose information of the $n$ images. Each $P_i$ is represented by a rotation matrix $R_i$ and a shift vector $\mathbf{s}_i$, which describe the rotation and translation relative to the world coordinate system (with $P_1$ being the reference frame)

The objective of 3D reconstruction is to recover optimal 3D structures $\mathcal{G}$, as well as camera poses $\mathcal{P}$, which minimizes the differences between the training images and the projection of the 3D Gaussian points onto the current image views as: 
\begin{equation}
\mathcal{L} = \min_{\mathcal{G}, \mathcal{P}} f_d(\mathcal{I}, f_r(\mathcal{G}, \mathcal{P})),
\end{equation}
where $f_r(\cdot)$ and $f_d(\cdot)$ are the render function and the distance function respectively.

In traditional 3DGS methods, $\mathcal{P}$ is treated as  known parameters, and the problem is reduced to: 
\begin{equation}
\mathcal{L} = \min_{\mathcal{G}} f_d(\mathcal{I}, f_r(\mathcal{G})).
\end{equation}
Specifically, the distance function $f_d$ is defined as the combination of the $L_1$ loss and D-SSIM terms:
\begin{equation}
\mathcal{L} = (1 - \lambda) \mathcal{L}_1 + \lambda \mathcal{L}_{\rm D-SSIM}.
\end{equation}
Detailed definition of $\mathcal{L}_1$ and $\mathcal{L}_{\rm D-SSIM}$ can be found in~\cite{10.1145/3592433}.

In this paper, we treat both $\mathcal{G}$ and $\mathcal{P}$ as learnable parameters, and optimize them jointly in the training step. To this end, we introduce a 3D optical flow loss to penalize the difference between the projections of two different views. The definition and the optimization method are described in detail in Section~\ref{subsec:camera pose refinement}.

\subsection{Joint Optimization Framework}
\label{subsec:JO_Framework}
\begin{figure*}[t]
\centering

\begin{minipage}[t]{0.48\textwidth}
\footnotesize
\begin{algorithm}[H]
\caption{Joint Optimization Framework %($\mathcal{I}$)
}
\label{alg:main}
\begin{algorithmic}[1]

\Statex \% $\mathcal{I}$: input images  
\Statex \% $\mathcal{G}$: Gaussian points  
\Statex \% $\mathcal{P}$: camera poses  
\Statex \% $T_{G}$: max iterations  
\Statex \% $k$: pose refinement interval  
\Statex \% $m$: last iteration for pose refinement

\State Initialize $(\mathcal{G}, \mathcal{P})$
\For{$t \gets 1$ \textbf{to} $T_{G}$}
    \If{$t \bmod k = 0$ \textbf{and} $t \leq m$}
        \State $\mathcal{P} \leftarrow \text{LK3D}(\mathcal{G}, \mathcal{P})$ 
    \Else
        \State $\mathcal{G} \leftarrow \text{3DGS}(\mathcal{G}, \mathcal{P})$ 
    \EndIf
\EndFor
\State \textbf{return} $\mathcal{G}, \mathcal{P}$
\end{algorithmic}
\end{algorithm}
\end{minipage}%
\hfill
\begin{minipage}[t]{0.48\textwidth}
\footnotesize
\begin{algorithm}[H]
\caption{Pose Optimization via LK3D}
\label{alg:pose_final}

\begin{algorithmic}[1]

\Statex \text{\% $\mathcal{I}$: input images}
\Statex \text{\% $\mathcal{G} = \{g_1, \ldots, g_k\}$: Gaussian points}
\Statex \text{\% $\mathcal{P} = \{P_1, \ldots, P_n\}$: camera poses}
\Statex \text{\% $T_{L}$: max iterations}

\ForAll{camera poses $P \in \mathcal{P}$}
    \For{$t \gets 1$ \textbf{to} $T_{L}$ }
        \ForAll{Gaussians $g \in \mathcal{G}$ }
            \State $l_g \leftarrow c(g) - I(\mathcal{W}(\mathbf{x}(g); P))$ 
            \State $d_g \leftarrow \nabla I \frac{\partial \mathcal{W}}{\partial P}$ 
        \EndFor
        \State $H \leftarrow \sum_g (d_g)^\top d_g$ 
        \State $\Delta P \leftarrow H^{-1} \sum_g (d_g)^\top l_g$ 
        \State  $P \leftarrow P + \Delta P$ 
    \EndFor
\EndFor
\State \textbf{return} optimized poses $\mathcal{P}$
\end{algorithmic}
\end{algorithm}
\end{minipage}
\vspace{-1em}
\end{figure*}

Our joint optimization framework establishes a dual-phase alternating minimization scheme to solve the coupled problem in Section~\ref{subsec:Problem definition}. Let $\mathcal{G}^{(t)}$ and $\mathcal{P}^{(t)} = \{R_i^{(t)}, \textbf{s}_i^{(t)}\}_{i=1}^n$ denote the 3D Gaussian parameters and camera poses at iteration $t$. As shown in Algorithms~\ref{alg:main}, these two parts of parameters are optimized by an alternating direction method, which contains two phases as follows:  

\textbf{Phase 1: Gaussian Parameter Update.} 
With fixed camera poses $\mathcal{P}^{(t)}$, we optimize $\mathcal{G}^{(t)}$ using the standard 3DGS pipeline. This involves minimizing the photometric reprojection error between rendered views and observed images:  
\begin{equation}  
\mathcal{G}^{(t+1)} = \arg\min_{\mathcal{G}} f_d( \mathcal{I}, f_r(\mathcal{G}, \mathcal{P}^{(t)}) ),
\end{equation}  
where $f_r(\cdot)$ denotes the differentiable rendering function of 3DGS. The optimization employs adaptive density control, spherical harmonic coefficients and opacity modulation as the original 3DGS formulation. 

\textbf{Phase 2: Camera Pose Update.}
With frozen Gaussian parameters $\mathcal{G}^{(t+1)}$, we refine camera poses by solving:  
\begin{equation}  
\mathcal{P}^{(t+1)} = \arg\min_{\mathcal{P}} f_d( \mathcal{I}, f_r(\mathcal{G}^{(t+1)}, \mathcal{P}) ).
\end{equation}  
Specifically, the incremental pose adjustment is computed using the LK3D algorithm described in Algorithms~\ref{alg:pose_final}, of which the detailed explanation is described in Section~\ref{subsec:camera pose refinement}. 

The two phases alternate at a fixed number of iterations. The differentiable nature of 3DGS rendering enables gradient flow through both phases. This alternating scheme progressively reduces the joint loss to convergence, with each phase benefiting from increasingly accurate estimates of the other.

\subsection{Initialize Camera Poses and Gaussian Points}  
Our initialization pipeline adopts a Structure-from-Motion (SfM) strategy similar in spirit to standard frameworks~\cite{schoenberger2016sfm}, but is independently implemented in a lightweight and modular manner tailored for downstream joint optimization. We extract \textit{Scale-Invariant Feature Transform} (SIFT)~\cite{2004Distinctive} descriptors and perform multi-view matching using \textit{Random Sample Consensus} (RANSAC)~\cite{10.1145/358669.358692} to estimate fundamental matrices. An initial camera pair with the highest number of correspondences is selected, and its relative pose is recovered via essential matrix decomposition. The 3D structure is then progressively expanded using \textit{Perspective-n-Point} (PnP)~\cite{article} pose estimation and multi-view triangulation. All camera poses and 3D points are jointly refined through global \textit{Bundle Adjustment} (BA)~\cite{10.5555/646271.685629} with robust cost functions to minimize reprojection errors.

Unlike the standard 3DGS, our method reconstructs both the initial sparse Gaussian points and camera poses entirely from scratch, without relying on external tools or pose priors. This design ensures compatibility with our joint optimization pipeline and provides greater control over reconstruction quality, sparsity, and initialization behavior.

\subsection{Camera Pose Refinement}
\label{subsec:camera pose refinement}
We propose a method of optimizing camera pose based on 3D Gaussian points and image reprojection error.
As illustrated in Algorithms~\ref{alg:main}, our method interleaves camera pose refinement with 3DGS training during the initial $m$ iterations. Specifically, we freeze the Gaussian parameters when performing pose optimization at regular intervals, while updating the 3DGS parameters in the remaining iterations. This alternating strategy ensures stable gradient propagation for subsequent scene reconstruction.
During the training process, the camera pose was optimized with the 3DGS model, and the camera rotation matrix and shift vector were calculated using  the projection of a 3D Gaussian points from multiple viewpoints.

The objective of this method is to reduce the photometric discrepancy between the source image and the reprojected appearance of 3D Gaussian points by adjusting the camera pose.  Specifically, given an initial estimate of the camera pose $P = [R\,|\, \mathbf{s}]$, where $R(\boldsymbol{\theta})$ is a rotation matrix parameterized by Euler angles $\boldsymbol{\theta} = (\theta_x, \theta_y, \theta_z)$, and $\mathbf{s}$ is a shift vector, the goal of the optimization is to refine $\boldsymbol{\theta}$ and $\mathbf{s}$ such that the projection $\mathcal{W}(\mathbf{x}(g);  P)$ of each Gaussian point $\mathbf{x}(g)$ onto the image plane better aligns with its corresponding appearance in the source image.  This alignment is achieved by minimizing the pixel-wise color difference
, thereby enabling accurate and robust camera pose estimation.

\vspace{1mm}\noindent\textbf{Lucas-Kanade 3D Optical Flow Algorithm.}
Let $\mathbf{p} = [R|\mathbf{s}]\in\mathcal{P}$ denote the camera pose of a certain target image.
Given a Gaussian point $g\in\mathcal{G}$, we denote $\mathbf{c}(g)$ the color value of $g$ and $\mathbf{x}(g)=(x_g, y_g, z_g)^\top$ the 3D position coordinates of $g$ in the world coordinate system. 
We then define a transformation function $\mathcal{W}(\mathbf{x}(g); \mathbf{p})$ that maps the 3D Gaussian coordinates $\mathbf{x}(g)$ from the world coordinate system to the target image plane following the standard projective geometry.

The goal of optimization is to minimize the discrepancy between the transformed image and the target image, which is defined as follows:
\begin{equation}
\mathbf{p}^* = \arg\min_{\mathbf{p}} \sum_{g\in\mathcal{G}} \big( \mathbf{c}(g) - I(\mathcal{W}(\mathbf{x}(g); \mathbf{p})) \big)^2.
\end{equation}
By minimizing the differences in pixels between the source and transformed images, the optimal pose parameters $\mathbf{p}^*$ can be determined.

It is difficult to directly compute the optimal camera pose $P$, since no close-form solution is available. 
Our method uses a gradient-based update approach by extending the standard LK algorithm~\cite{10.5555/1623264.1623280} to 3-dimensional space, which iteratively revises the transformation matrix $\mathbf{p}$ with an increment $\Delta \mathbf{p}$ as:
\begin{equation}
\Delta \mathbf{p}^* = \arg\min_{\Delta \mathbf{p}} \sum_{g\in\mathcal{G}} \big( I(\mathcal{W}(\mathbf{x}(g); \mathbf{p} + \Delta \mathbf{p})) - \mathbf{c}(g) \big)^2.
\end{equation}
For computation efficiency, we further approximate this using a first-order Taylor expansion:
\begin{equation}
\Delta \mathbf{p}^* \approx \arg \min_{\Delta P} \sum_{g\in\mathcal{G}} \big( I(\mathcal{W}(\mathbf{x}(g); \mathbf{p})) + \nabla I \frac{\partial \mathcal{W}}{\partial \mathbf{p}} \Delta \mathbf{p} - \mathbf{c}(g) \big)^2,
\end{equation}
where $\nabla I$ represents the image gradient and $\frac{\partial \mathcal{W}}{\partial \mathbf{p}}$ is the Jacobian matrix of the transformation function $\mathcal{W}$ with respect to the transformation parameters $\mathbf{p}$. 
According to the principle that the derivative at the extreme value is zero, the pose increment $\Delta \mathbf{p}$  is computed via Gauss-Newton approximation:
\begin{equation}
\Delta \mathbf{p} 
\approx H^{-1} \sum_{g\in\mathcal{G}} 
\Big(
\nabla I \frac{\partial \mathcal{W}}{\partial \mathbf{p}} 
\Big)^\top
\big( \mathbf{c}(g) - I(\mathcal{W}(\mathbf{x}(g); \mathbf{p}))\big),
\end{equation}
where the Hessian $H$ is computed as:
\begin{equation}
H = \sum_{g\in\mathcal{G}} \left( \nabla I \frac{\partial \mathcal{W}}{\partial \mathbf{p}} \right)^\top \left( \nabla I \frac{\partial \mathcal{W}}{\partial \mathbf{p}} \right).
\end{equation}

This method significantly reduces the error between the source and target photos, resulting in accurate camera poses.
Note that the refinement of each camera pose $P\in\mathcal{P}$ can be performed independently following the same pipeline, which facilitates the parallel implementation of the LK3D algorithm. 

\begin{table*}[htbp]
\centering
\small
\renewcommand{\arraystretch}{1.0}
\setlength{\tabcolsep}{2pt}
\captionsetup{font=small}
\caption{\textbf{Quantitative comparison on Tanks and Temples.} 
The best results are highlighted in \textbf{bold}, and the second in \underline{underline}, and the same styles are adopted in the subsequent tables. 
}
\resizebox{0.78\textwidth}{!}{
\begin{tabular}{c|cccc | cccc | cccc}
\toprule[1.2pt]
\multirow{2}{*}{\textbf{Scene}} & 
\multicolumn{4}{c|}{\textbf{PSNR} $\uparrow$} & 
\multicolumn{4}{c|}{\textbf{SSIM} $\uparrow$} & 
\multicolumn{4}{c}{\textbf{LPIPS} $\downarrow$} \\
\cmidrule(lr){2-5} \cmidrule(lr){6-9} \cmidrule(l){10-13}
& 3DGS & CFGS~\cite{Fu_2024_CVPR} & GSHT~\cite{ji2024sfmfree3dgaussiansplatting} & Ours &
3DGS & CFGS & GSHT & Ours &
3DGS & CFGS & GSHT & Ours \\
\midrule[0.8pt]
Ballroom    & \textbf{32.13} & 16.83 & 16.56 & \underline{30.96} & \underline{0.86} & 0.45 & 0.45 & \textbf{0.94} & \underline{0.12} & 0.40 & 0.25 & \textbf{0.05} \\
Barn        & \textbf{28.30} & 17.28 & 21.16 & \underline{26.88} & \textbf{0.92} & 0.51 & 0.63 & \underline{0.88} & \textbf{0.09} & 0.42 & 0.27 & \underline{0.12} \\
Church      & \textbf{29.01} & 20.51 & 20.55 & \underline{25.88} & \textbf{0.92} & 0.64 & 0.75 & \underline{0.86} & \textbf{0.09} & 0.33 & \underline{0.15} & 0.16 \\
Family      & \underline{25.67} & 14.37 & \textbf{29.02} & 25.44 & \underline{0.90} & 0.45 & \textbf{0.91} & 0.88 & \underline{0.12} & 0.47 & \textbf{0.09} & 0.15 \\
Francis     & 24.97 & 20.45 & \textbf{28.89} & \underline{27.92} & 0.83 & 0.62 & \underline{0.84} & \textbf{0.87} & 0.23 & 0.37 & \underline{0.20} & \textbf{0.19} \\
Horse       & 20.22 & 17.49 & \textbf{27.94} & \underline{26.53} & 0.80 & 0.61 & \textbf{0.90} & \textbf{0.90} & 0.22 & 0.35 & \textbf{0.09} & \underline{0.11} \\
Ignatius    & \textbf{26.87} & 17.16 & 20.95 & \underline{25.13} & \textbf{0.85} & 0.37 & 0.61 & \underline{0.81} & \textbf{0.12} & 0.41 & 0.21 & \underline{0.15} \\ 
Museum      & \textbf{27.25} & 16.36 & 12.44 & \underline{26.54} & \textbf{0.88} & 0.52 & 0.30 & \underline{0.87} & \textbf{0.09} & 0.47 & 0.59 & \underline{0.10} \\

\midrule[0.8pt]
Mean        & \underline{26.80} & 17.55 & 22.57 & \textbf{26.91} & \underline{0.87} & 0.52 & 0.67 & \textbf{0.88} & \textbf{0.13} & 0.40 & 0.23 & \textbf{0.13} \\
\bottomrule[1.2pt]
\end{tabular}
}

\label{table:nvs_tank_sub}
\vspace{-1em}
\end{table*}

\begin{table*}[htbp]
\centering
\small
\renewcommand{\arraystretch}{1.0}
\setlength{\tabcolsep}{3pt}
\captionsetup{font=small}
\caption{\textbf{Quantitative comparison on LLFF-NeRF}. 
For the Fortress and Leaves scenes (marked with *), we directly cite the results of CFGS and GSHT from zeroGS~\cite{chen2024zerogstraining3dgaussian}, because our experimental environment could not meet the running requirements of their codes. 
}
\resizebox{0.78\textwidth}{!}{
\begin{tabular}{c|cccc|cccc|cccc}
\toprule[1.2pt]
\multirow{2}{*}{\textbf{Scene}} & 
\multicolumn{4}{c|}{\textbf{PSNR} $\uparrow$} & 
\multicolumn{4}{c|}{\textbf{SSIM} $\uparrow$} & 
\multicolumn{4}{c}{\textbf{LPIPS} $\downarrow$} \\
\cmidrule(lr){2-5} \cmidrule(lr){6-9} \cmidrule(l){10-13}
& 3DGS & CFGS & GSHT & Ours &
3DGS & CFGS & GSHT & Ours &
3DGS & CFGS & GSHT & Ours \\
\midrule[0.8pt]
Fern       & \textbf{23.55} & 16.65 & 18.09 & \underline{22.93} & \textbf{0.80} & 0.50 & 0.56 & \underline{0.77} & \underline{0.23} & 0.46 & 0.44 & \textbf{0.22} \\
Flower     & \underline{25.56} & 21.16 & 19.20 & \textbf{27.50} & \underline{0.82} & 0.67 & 0.67 & \textbf{0.85} & \underline{0.24} & 0.41 & 0.46 & \textbf{0.20} \\
Fortress$*$& \textbf{29.50} & 14.73 & 16.26 & \underline{29.13} & \textbf{0.87} & 0.40 & 0.48 & \underline{0.86} & \textbf{0.18} & 0.46 & 0.46 & \underline{0.19} \\
Horns      & \textbf{26.98} & 16.13 & 17.62 & \underline{26.71} & \textbf{0.88} & 0.49 & 0.56 & \underline{0.87} & \textbf{0.19} & 0.52 & 0.54 & \underline{0.20} \\
Leaves$*$  & \underline{17.91} & 15.38 & 15.69 & \textbf{18.25} & \underline{0.59} & 0.42 & 0.42 & \textbf{0.60} & \textbf{0.21} & 0.40 & 0.33 & \underline{0.27} \\
Orchids    & \textbf{19.45} & 13.65 & 13.73 & \underline{19.07} & \textbf{0.65} & 0.29 & 0.29 & \underline{0.64} & \textbf{0.25} & 0.55 & 0.56 & \textbf{0.25} \\
Room       & \underline{31.85} & 19.25 & 19.76 & \textbf{32.14} & \textbf{0.95} & 0.77 & 0.80 & \textbf{0.95} & \textbf{0.13} & 0.36 & 0.35 & \textbf{0.13} \\
Trex       & \underline{26.00} & 18.16 & 18.30 & \textbf{27.41} & \underline{0.90} & 0.61 & 0.64 & \textbf{0.91} & \underline{0.20} & 0.44 & 0.47 & \textbf{0.18} \\
\midrule[0.8pt]
Mean       & \underline{25.10} & 16.89 & 17.33 & \textbf{25.39} & \textbf{0.81} & 0.52 & 0.55 & \underline{0.80} & \textbf{0.20} & 0.45 & 0.45 & \underline{0.21} \\
\bottomrule[1.2pt]
\end{tabular}
}
\label{table:nvs_llff}
\vspace{-1em}
\end{table*}

\subsection{Optimization Strategy of Rotation Matrix.}
In the pose optimization based on reprojection error, the rotation part of the camera transformation matrices must strictly remain as valid rotation matrices with orthonormal columns and determinant equal to one. 
Direct gradient updates on rotation matrices may violate their orthogonality, leading to numerical instability. Here we adopt an Euler angle parameterization strategy that decomposes rotation matrices into independent Euler angles (pitch $\alpha$, yaw $\beta$, roll $\gamma$) around $x$, $y$ and $z$-axes, and utilizes their analytic derivatives for stable iteration while preserving orthogonality.

Specifically, the rotation matrix $R$ is parameterized as:
\begin{equation}
R = R_z(\gamma)R_y(\beta)R_x(\alpha),
\end{equation}
where the axial rotation matrices $R_x(\alpha)$, $R_y(\beta)$, and $R_z(\gamma)$ inherently satisfy orthogonality. During optimization, the Jacobians of the rotation matrix with respect to Euler angles are computed via chain rule:
\begin{align}
\frac{\partial R}{\partial \alpha} &= R_z(\gamma) R_y(\beta)\frac{\partial R_x(\alpha)}{\partial \alpha}, \\
\frac{\partial R}{\partial \beta} &= R_z(\gamma)\frac{\partial R_y(\beta)}{\partial \beta} R_x(\alpha), \\
\frac{\partial R}{\partial \gamma} &= \frac{\partial R_z(\gamma)}{\partial \gamma}R_y(\beta)R_x(\alpha).
\end{align}

This parameterization decouples the rotation matrix degrees of freedom into unconstrained Euler angle increments $\Delta\alpha, \Delta\beta, \Delta\gamma$. Using gradient descent with learning rate $\eta$, the angles are updated as:
\begin{equation}
\alpha \leftarrow \alpha + \eta\Delta\alpha,\quad \beta \leftarrow \beta + \eta\Delta\beta,\quad \gamma \leftarrow \gamma + \eta\Delta\gamma.
\end{equation}

The strategy offers two main advantages for minimizing the reprojection error. First, the local linearization of Euler angle updates preserves the orthogonality of $R\in \text{SO}(3)$, preventing manifold deviations that can arise from direct matrix optimization. Second, compared to global matrix parameterization, the angle-based decomposition significantly reduces the complexity of Jacobian computations, enhancing both optimization efficiency and numerical stability.

\section{Experiments}
\label{sec:experiments}
\subsection{Datasets}
\label{Datasets}
 We conducted extensive experiments on various datasets, including LLFF-NeRF~\cite{mildenhall2019llff}, Tanks and Temples~\cite{Knapitsch2017} and
Shiny~\cite{Wizadwongsa2021NeX}.
\textbf{LLFF-NeRF:} This dataset contains real-world multi-view images captured by various devices, comprising eight scenes. The number of images varies across scenes, with the scene \textit{fern} having the fewest (twenty images) and \textit{horns} the most (sixty-two images). 
\textbf{Tanks and Temples:} This dataset comprises eight scenes, encompassing both indoor and outdoor environments. In line with the configurations of  CFGS~\cite{Fu_2024_CVPR} and GSHT~\cite{ji2024sfmfree3dgaussiansplatting}, we further enhanced the complexity of the dataset. Given the limited variation in camera poses in the original dataset, we uniformly sampled one-fifth of the images from each scene to amplify the pose variation between consecutive frames. \textbf{Shiny:} The dataset consists of a number of challenging scenes with significant reflected or refracted lighting changes.
Since the data volume per scene varies across datasets, we adopted proportional data splitting rather than using fixed quantities. For each scene, seven-eighths of the data were allocated for training, with the remaining one-eighth reserved for testing.
\begin{figure*}[!t]
   \vspace{-1em}
    \vskip 0.2in
    \begin{center}
    \centering
    \small
    \captionsetup{font=small}
    \resizebox{0.9\linewidth}{!}{
        \begin{tabular}{@{\hspace{0mm}}c@{\hspace{0mm}}c@{\hspace{.5mm}}c@{\hspace{.5mm}}c@{\hspace{.5mm}}c@{\hspace{.5mm}}c@{\hspace{0mm}}}
            % 第一行：Fern场景
    \begin{minipage}[t]{0.02\textwidth} 
        \rotatebox{90}{\centering {\footnotesize LLFF-Fern}} 
    \end{minipage} &
            \includegraphics[width=0.199\linewidth]{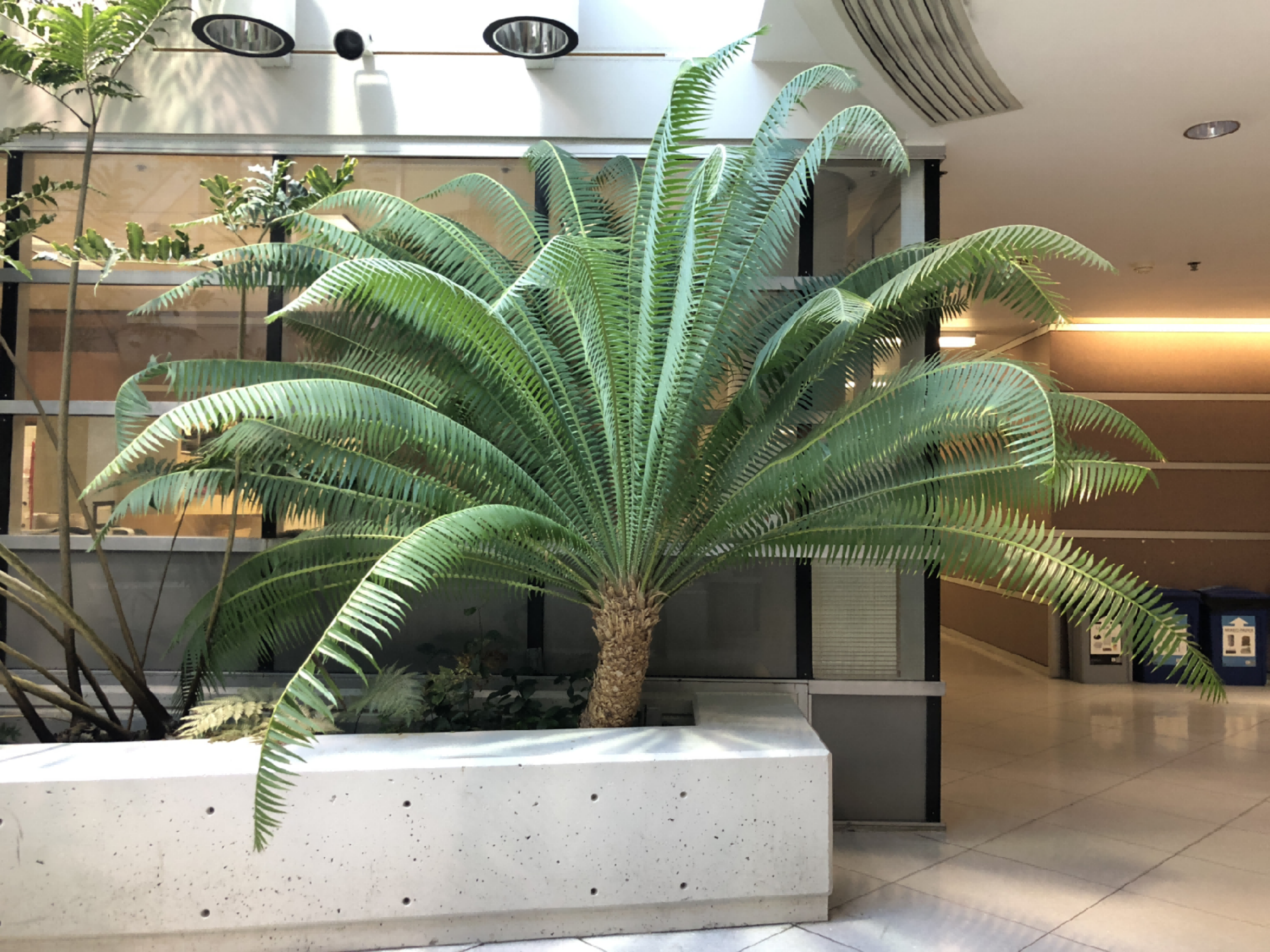} &
            \includegraphics[width=0.199\linewidth]{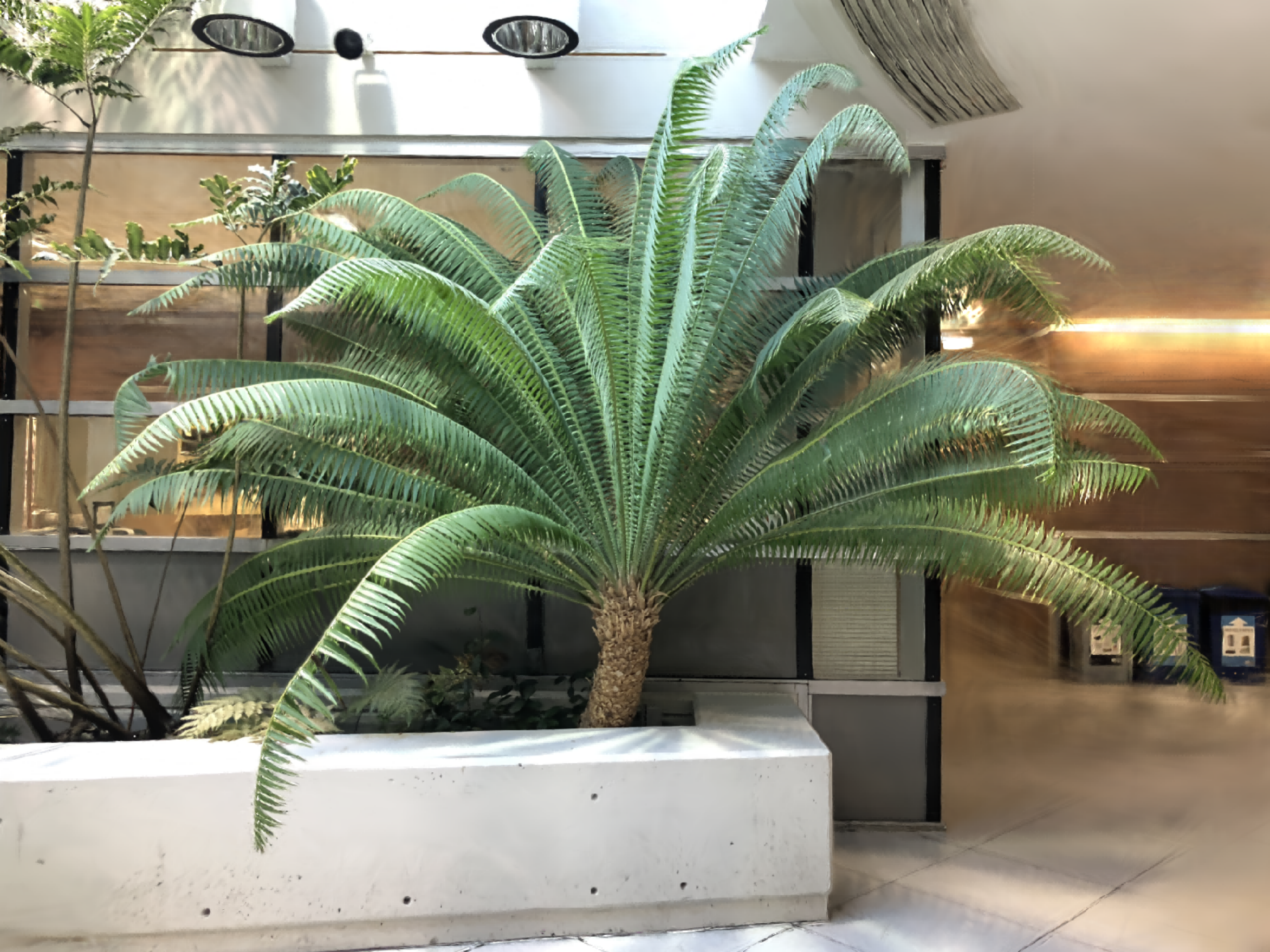} &
            \includegraphics[width=0.199\linewidth]{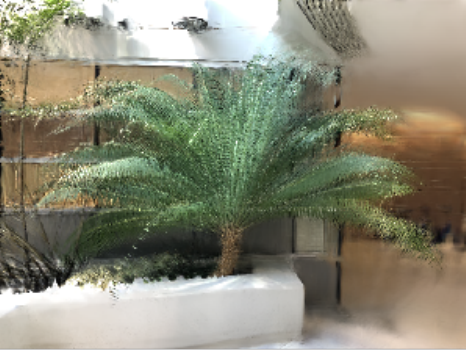} &
            \includegraphics[width=0.199\linewidth]{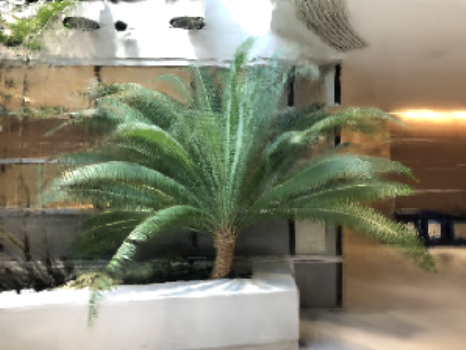} &
            \includegraphics[width=0.199\linewidth]{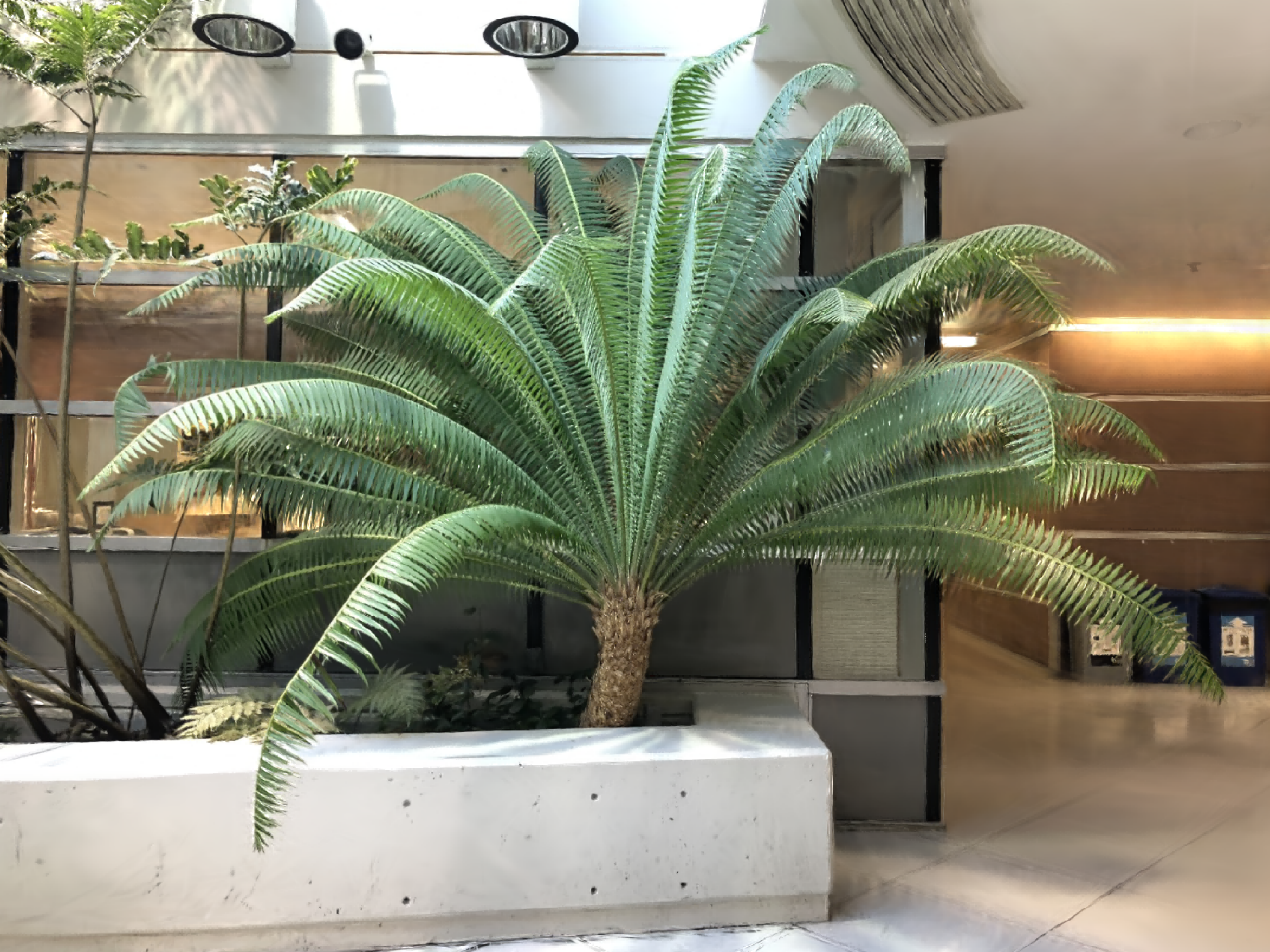} \\
            
            % 第二行：Flower场景
    \begin{minipage}[t]{0.02\textwidth} 
        \rotatebox{90}{\centering {\footnotesize LLFF-Flower}}
    \end{minipage} &
            \includegraphics[width=0.199\linewidth]{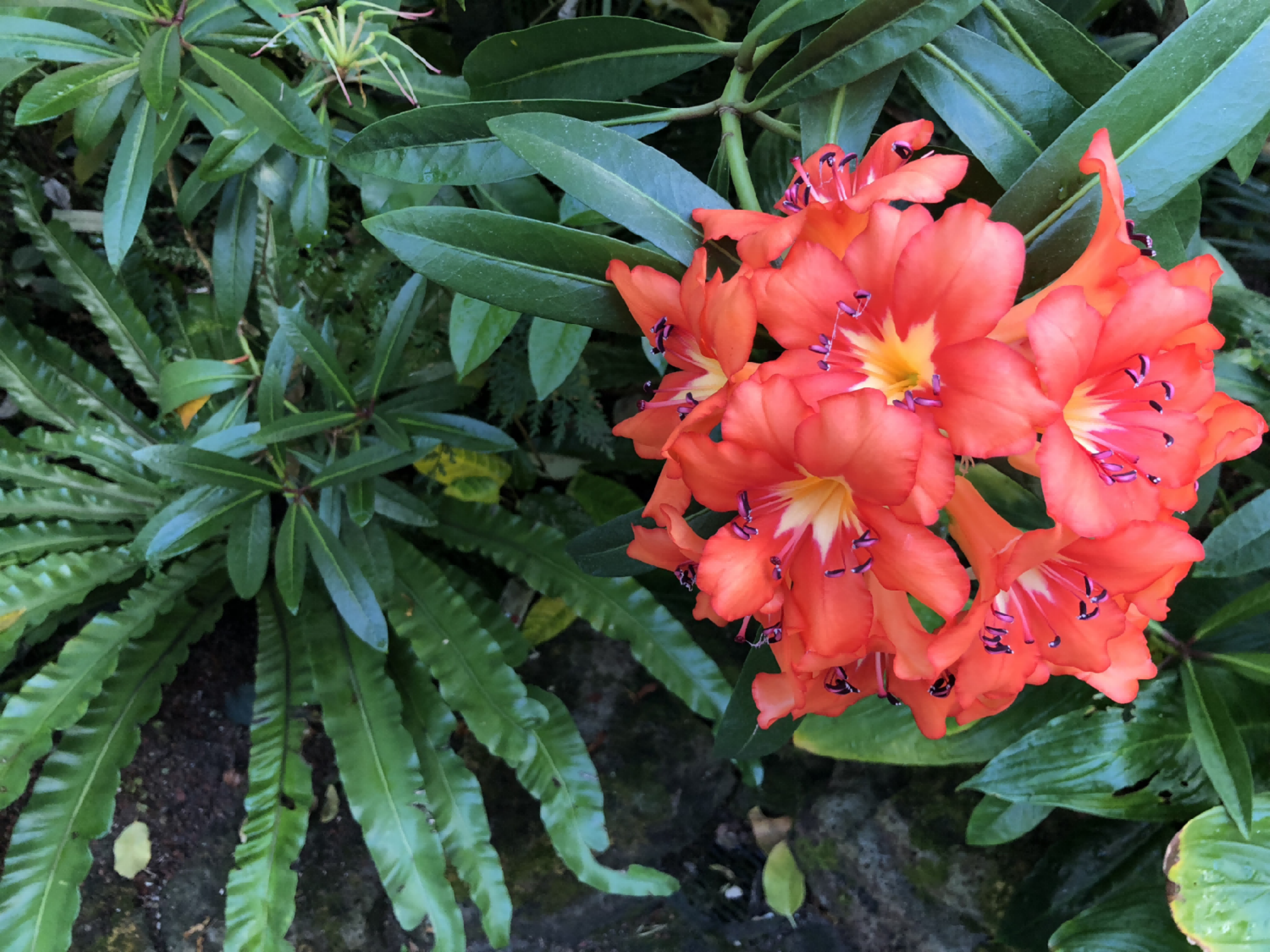} &
            \includegraphics[width=0.199\linewidth]{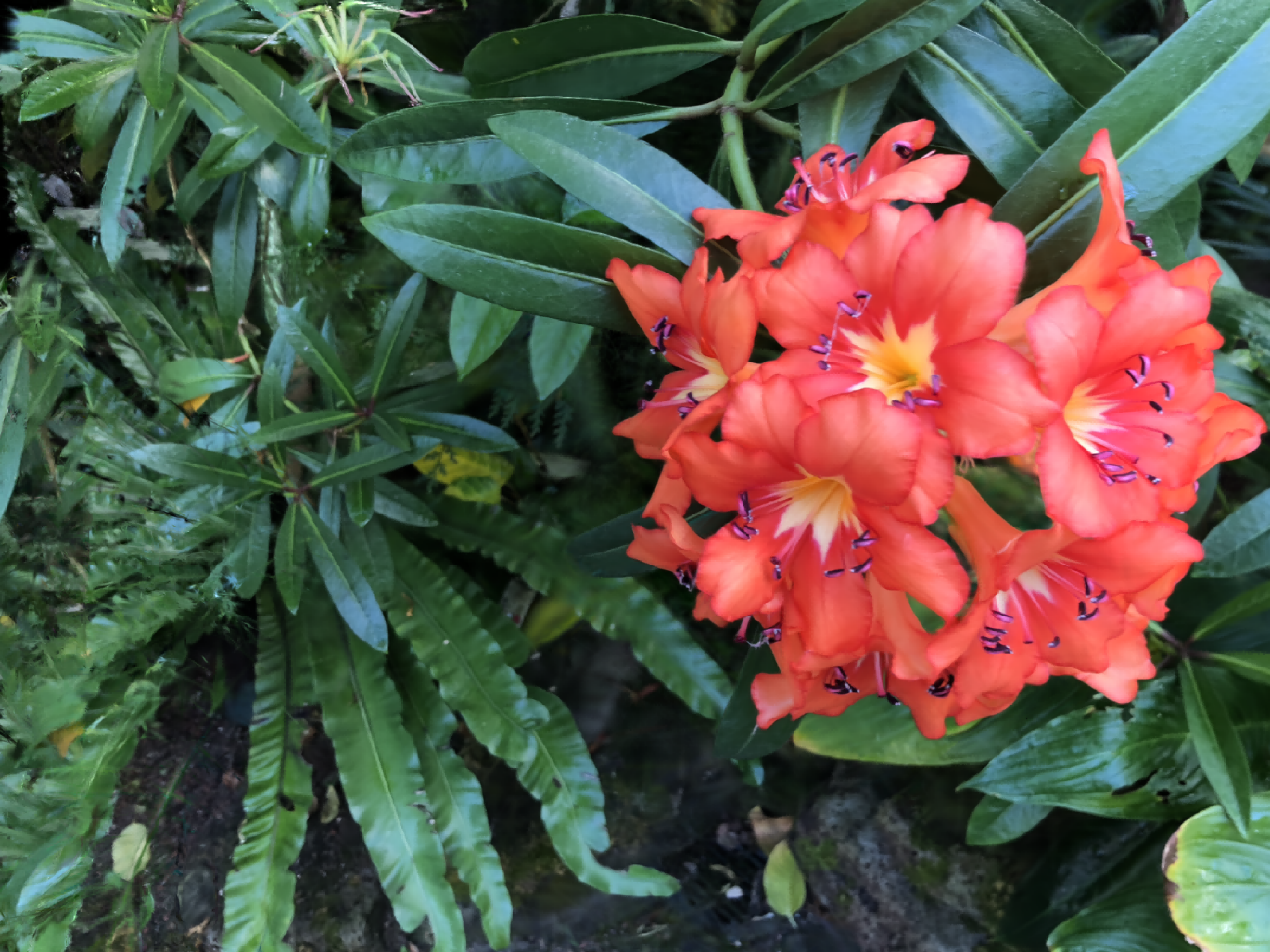} &
            \includegraphics[width=0.199\linewidth]{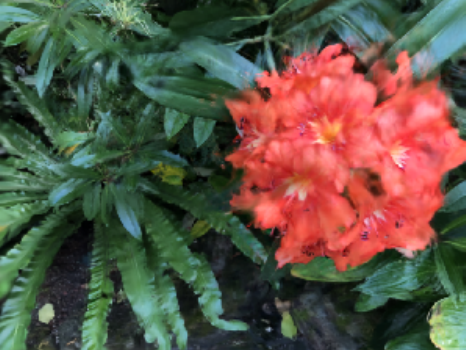} &
            \includegraphics[width=0.199\linewidth]{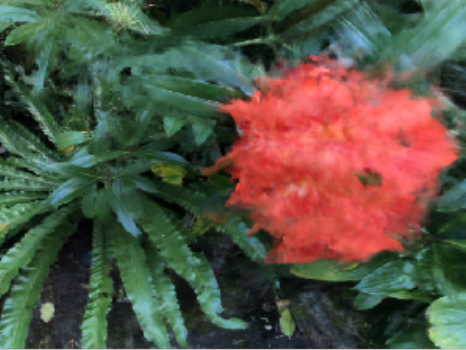} &
            \includegraphics[width=0.199\linewidth]{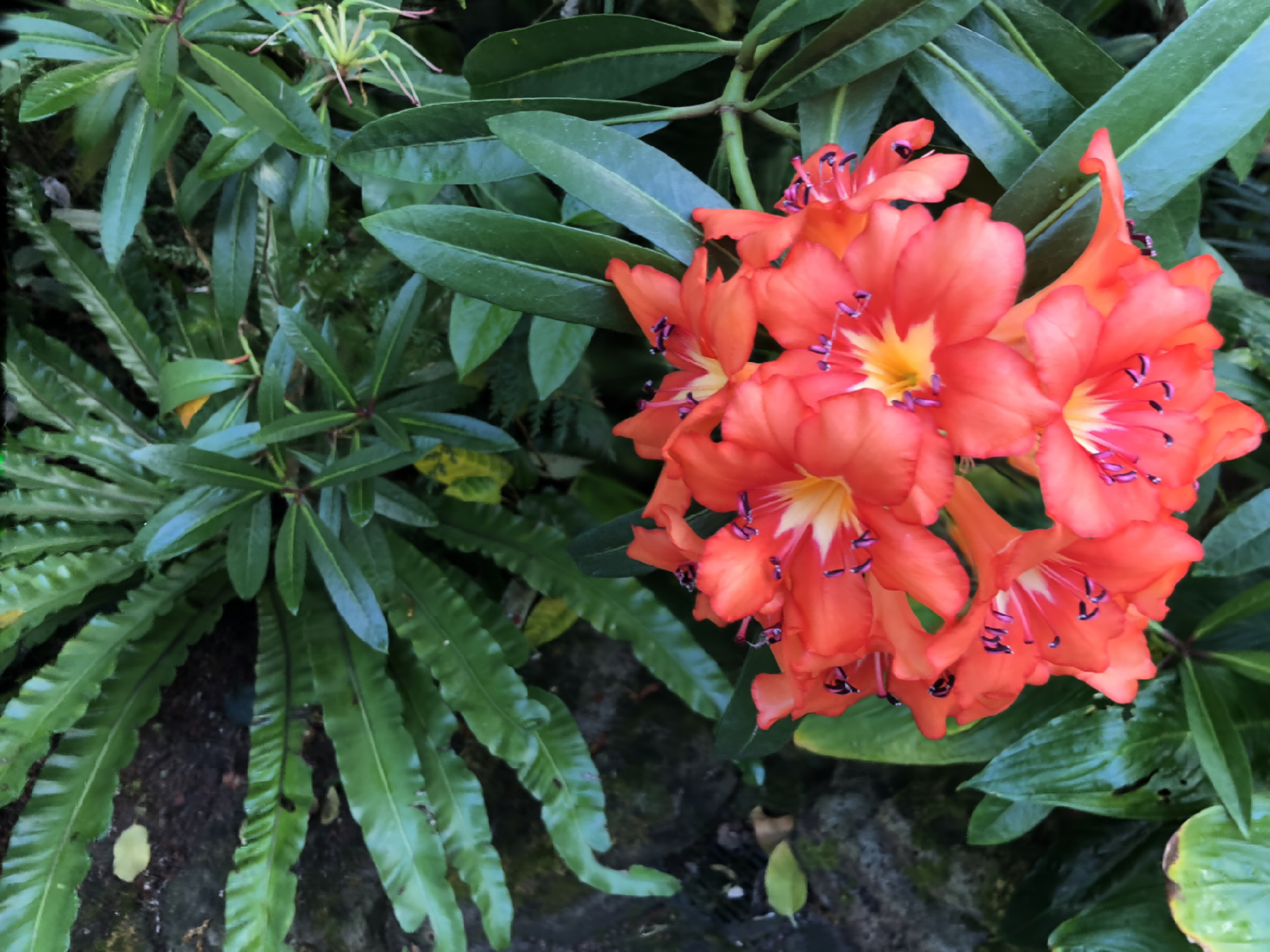} \\
            
            % 第三行：Trex场景
   \begin{minipage}[t]{0.02\textwidth} 
        \rotatebox{90}{\centering {\footnotesize LLFF-Trex}} 
    \end{minipage} &
            \includegraphics[width=0.199\linewidth]{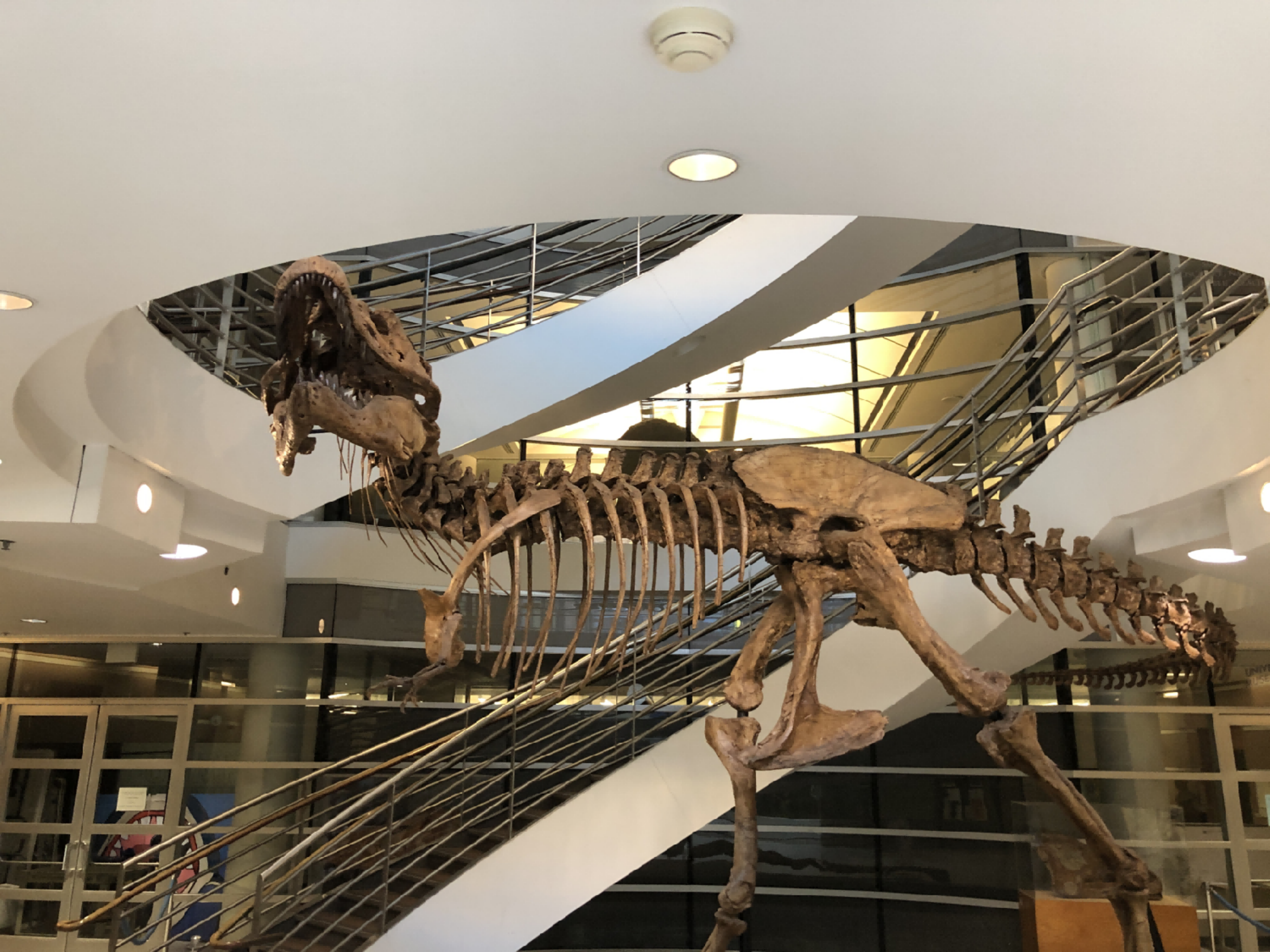} &
            \includegraphics[width=0.199\linewidth]{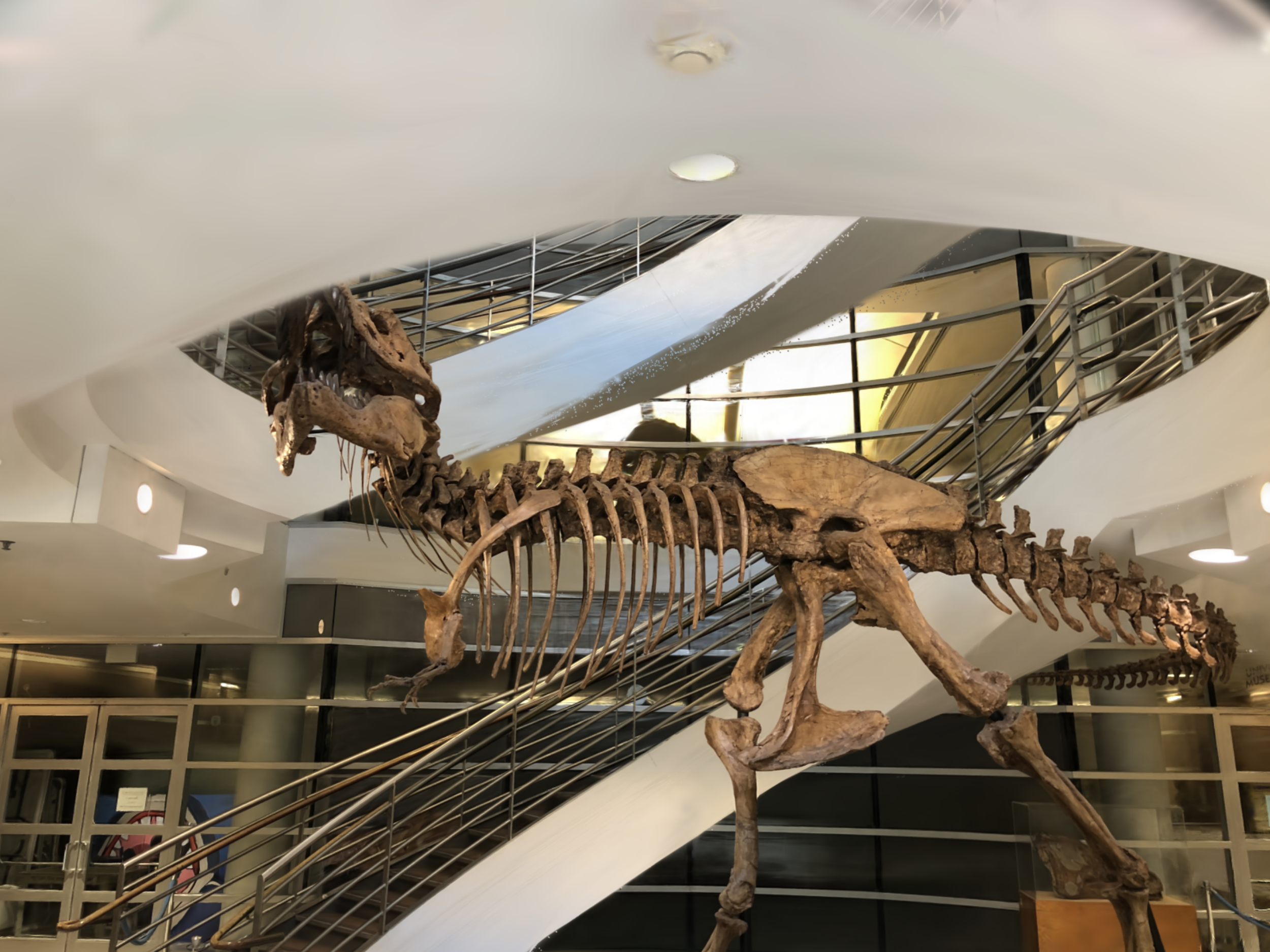} &
            \includegraphics[width=0.199\linewidth]{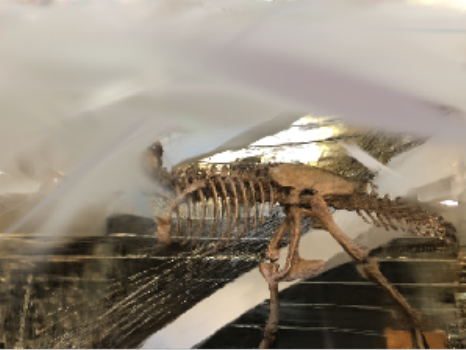} &
            \includegraphics[width=0.199\linewidth]{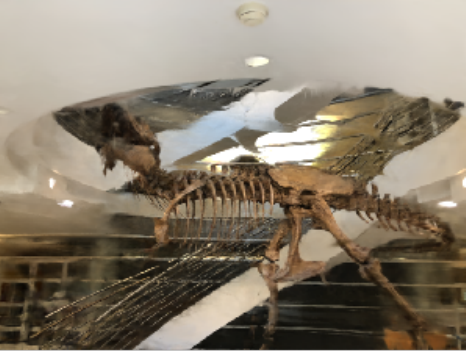} &
            \includegraphics[width=0.199\linewidth]{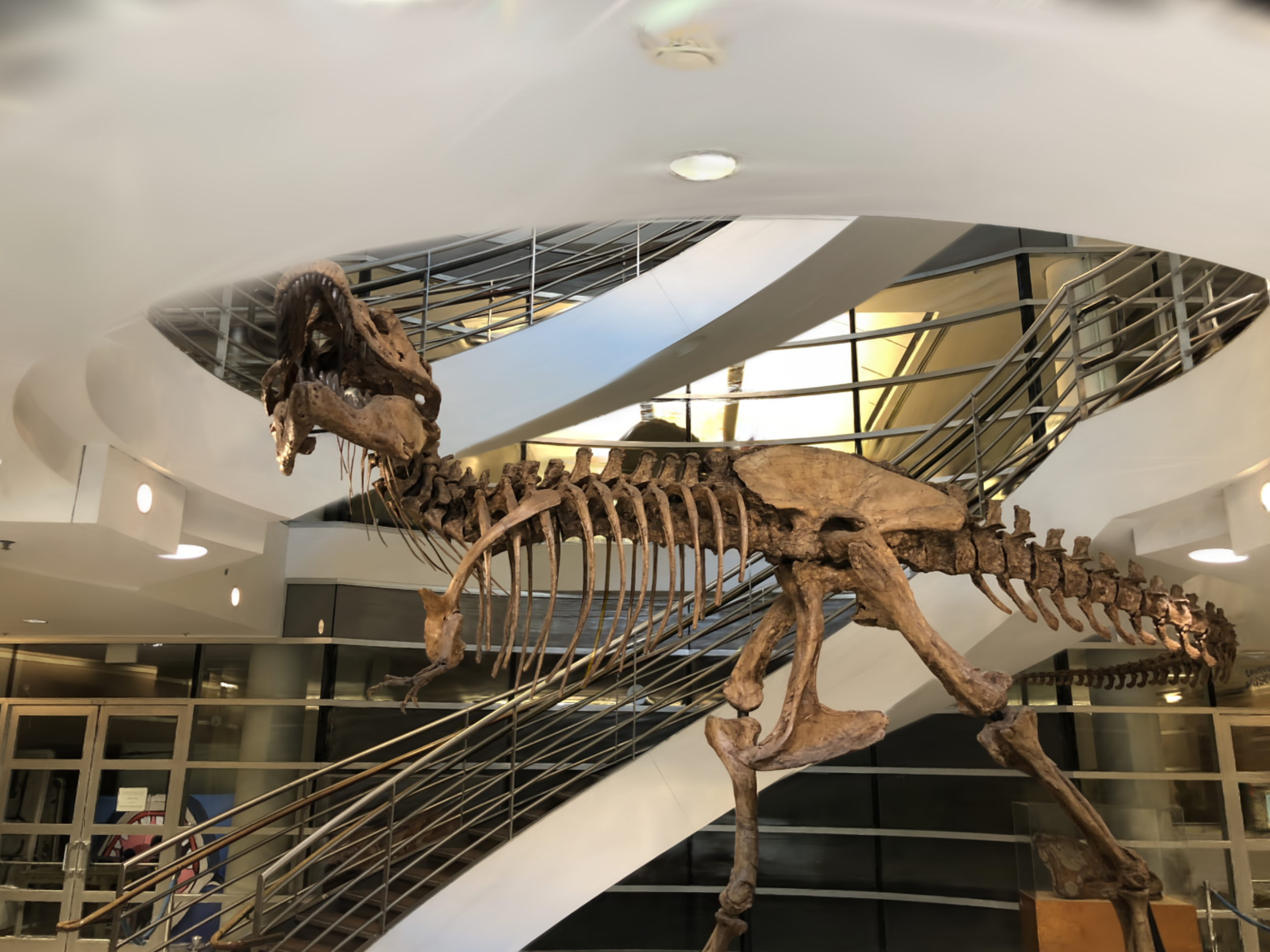} \\
            
            % 第四行：Lab场景
   \begin{minipage}[t]{0.02\textwidth} 
              % \vspace{-1cm} 
        \rotatebox{90}{\centering {\footnotesize Shiny-Lab}} 
    \end{minipage} &
            \includegraphics[width=0.199\linewidth]{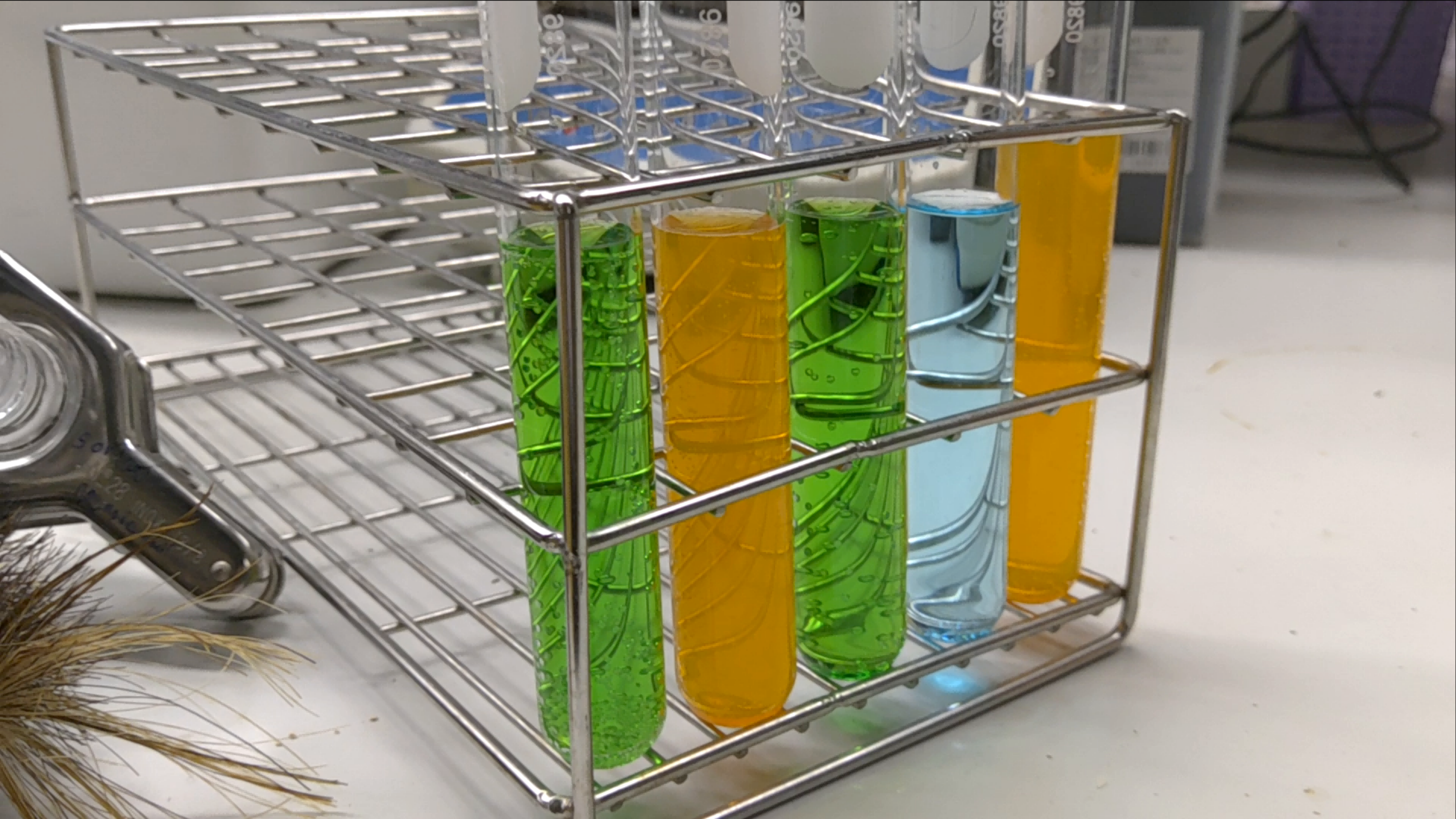} &
            \includegraphics[width=0.199\linewidth]{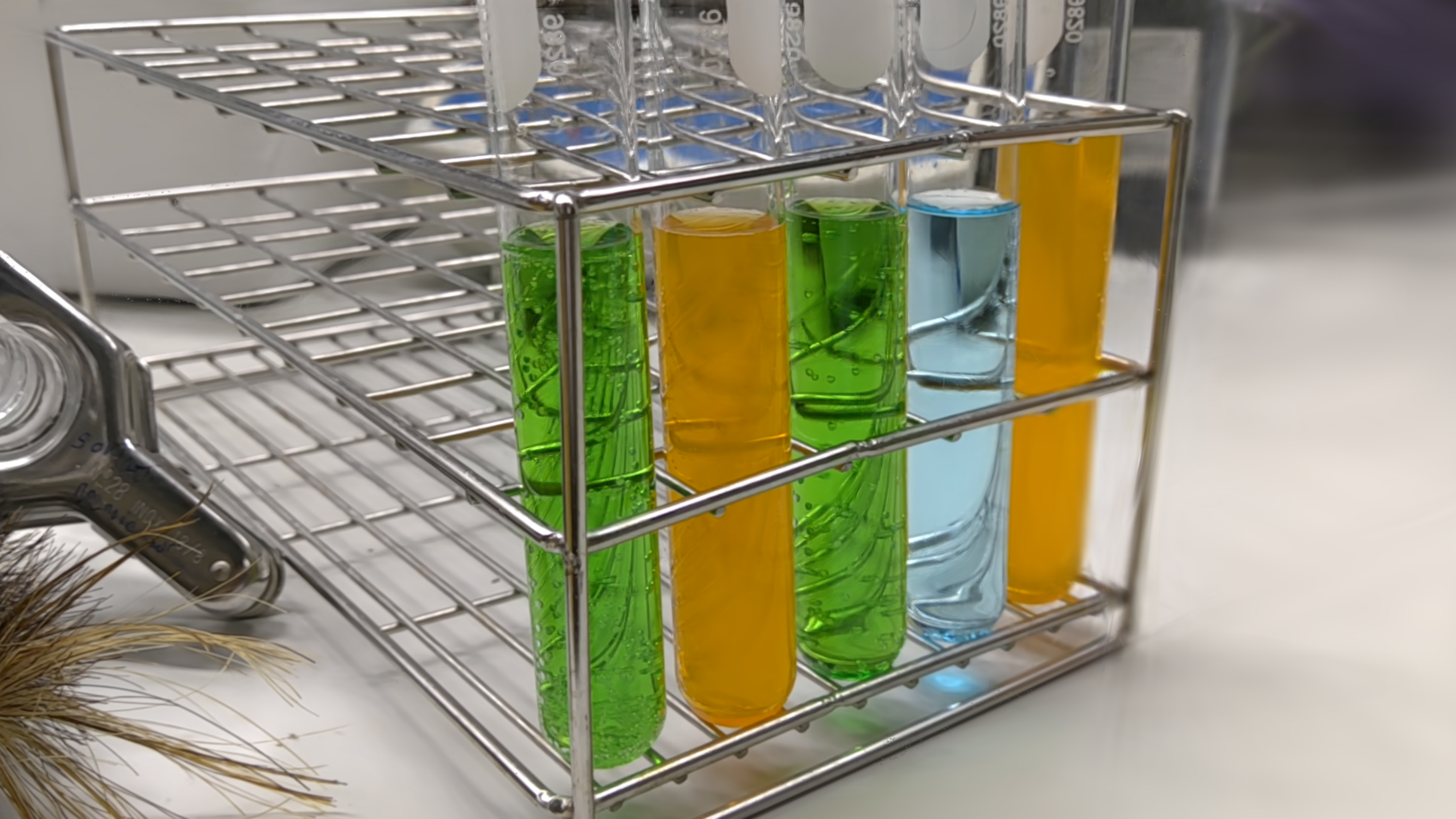} &
            \includegraphics[width=0.199\linewidth]{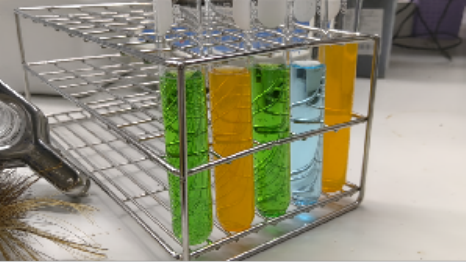} &
            \includegraphics[width=0.199\linewidth]{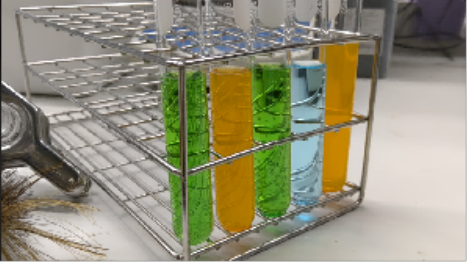} &
            \includegraphics[width=0.199\linewidth]{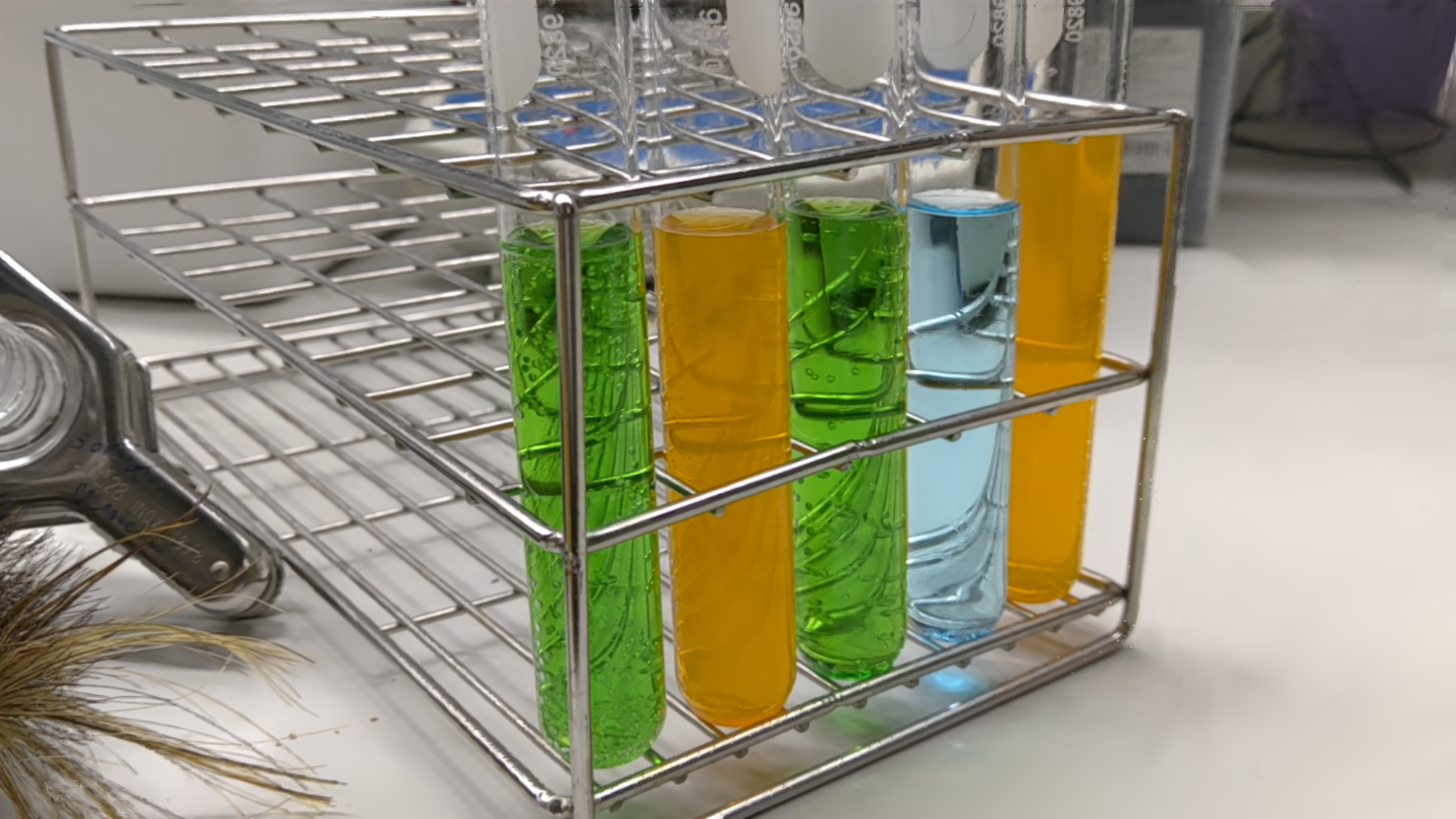} \\
            
            % 第五行：Horse场景
     \begin{minipage}[t]{0.02\textwidth} 
        \rotatebox{90}{\centering {\footnotesize Tanks-Horse}} 
    \end{minipage} &
            \includegraphics[width=0.199\linewidth]{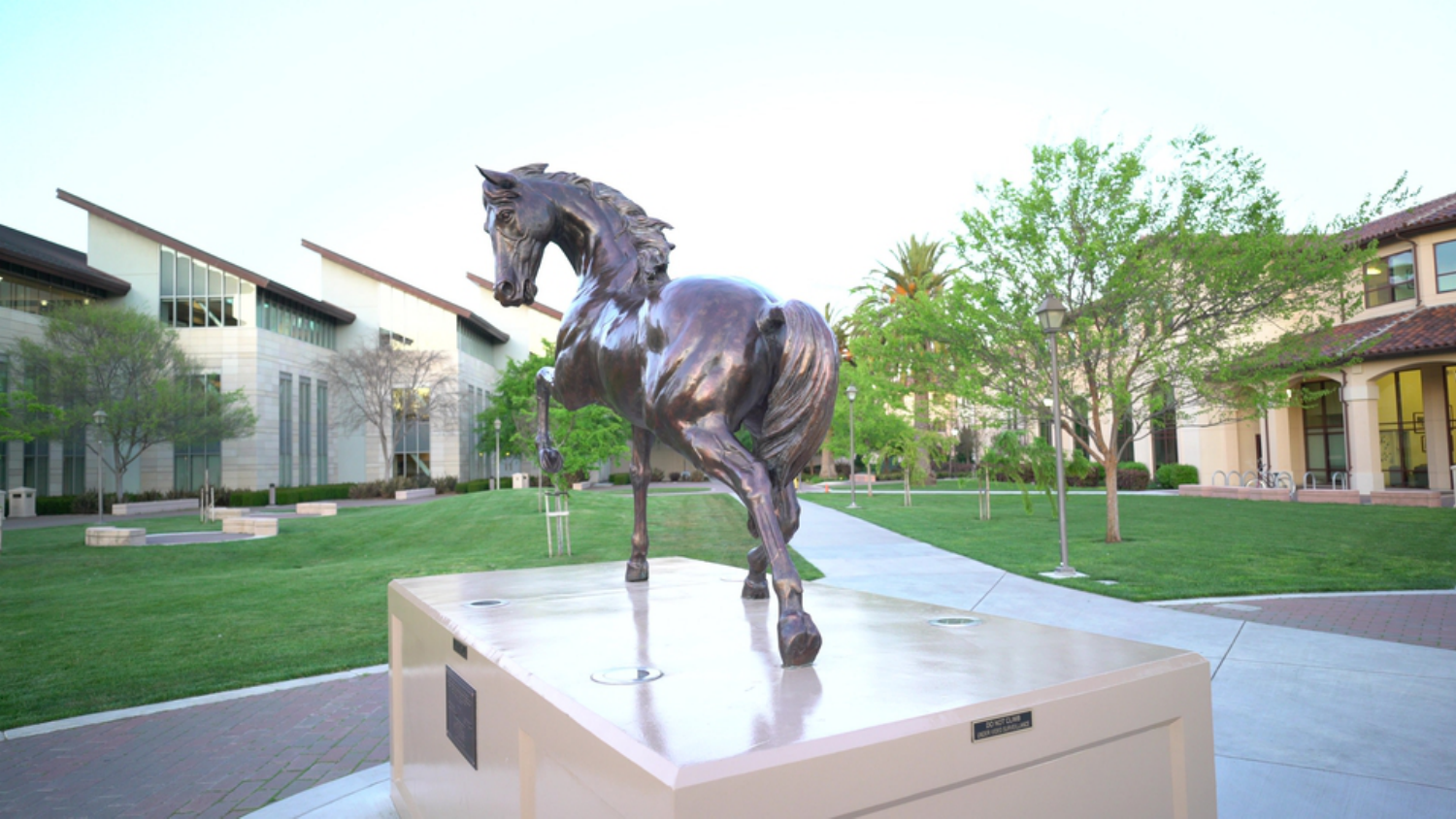} &
            \includegraphics[width=0.199\linewidth]{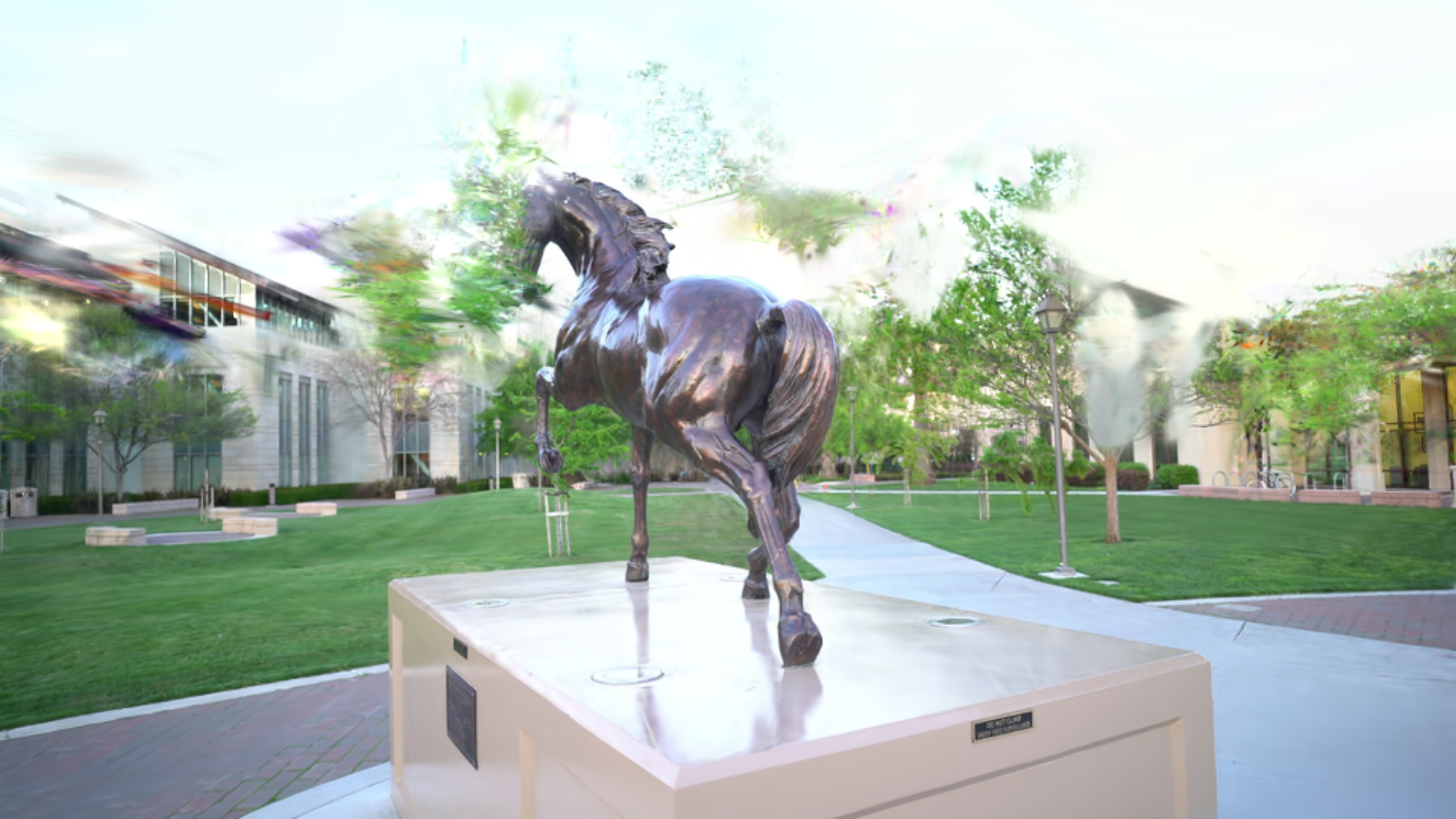} &
            \includegraphics[width=0.199\linewidth]{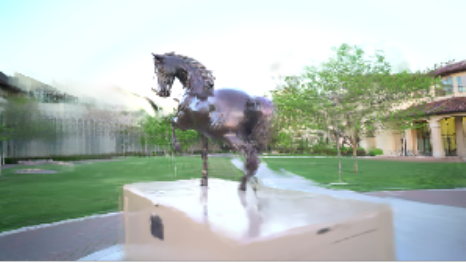} &
            \includegraphics[width=0.199\linewidth]{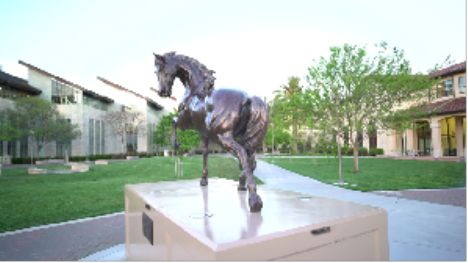} &
            \includegraphics[width=0.199\linewidth]{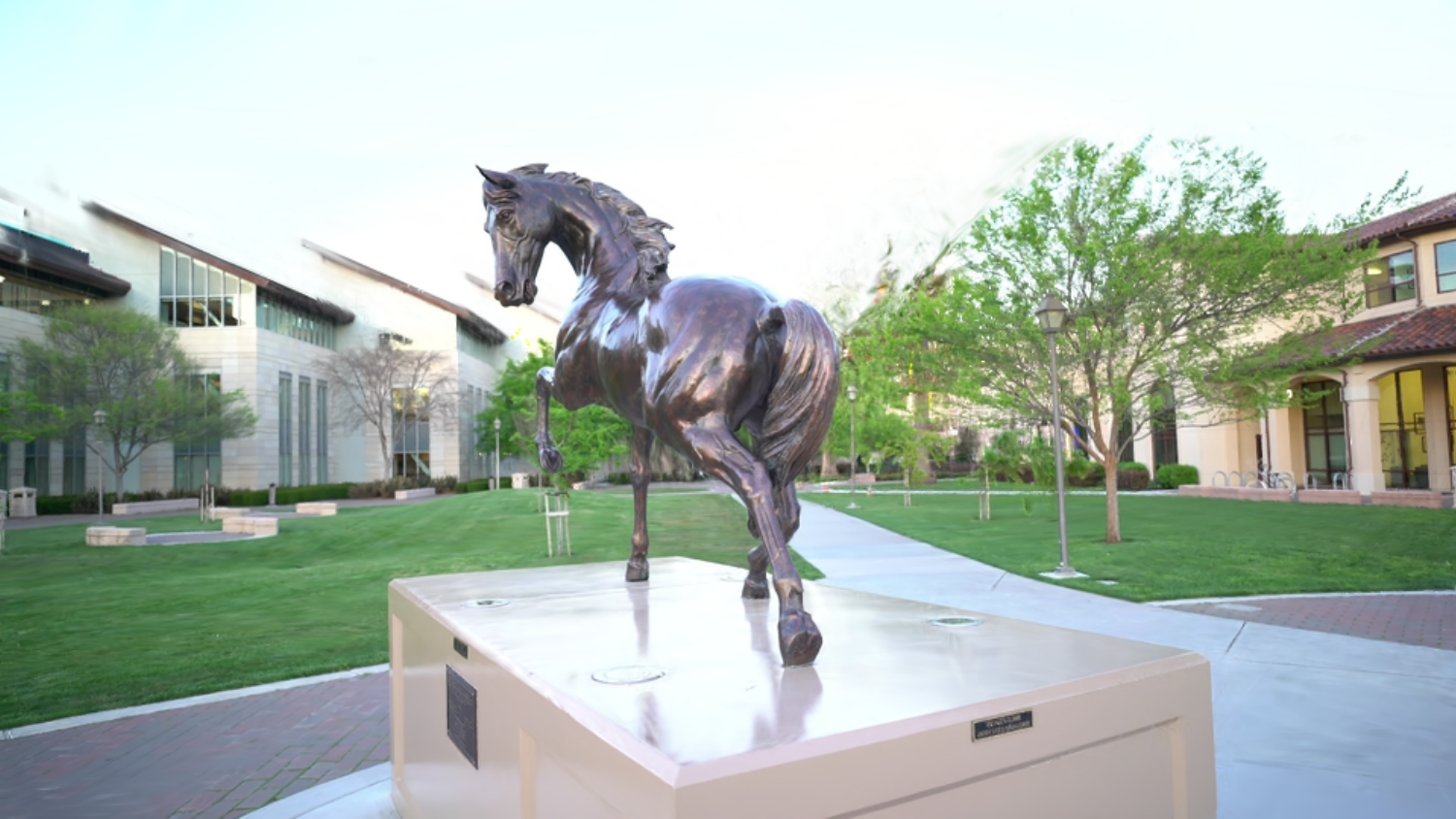} \\
            
            % 标注行
            \multicolumn{1}{@{}c@{}}{} & (a) GT &  (b) 3DGS & (c) CFGS & (d) GSHT & (e) Ours \\
        \end{tabular}}
    \vspace{-1mm}
    \caption{\textbf{Qualitative results of several representative samples from LLFF-NeRF, Tanks and Temples, Shiny.} Our method achieves consistently high rendering quality across all scenes.}
    \label{fig:nvs_all_2}
\vspace{-3mm}
\end{center}
\vspace{-3mm}

\end{figure*}
\begin{table*}[htbp]
\centering
\small
\renewcommand{\arraystretch}{1.0}
\setlength{\tabcolsep}{3pt}
\captionsetup{font=small}
\caption{\textbf{Quantitative comparison on Shiny}. The original dataset contains eight scenes, but both CFGS and GSHT suffer from running errors and fail to report the final results in some scenes. Thus we only report the experimental results for four scenes. 
}
\resizebox{0.77\textwidth}{!}{
\begin{tabular}{c|cccc|cccc|cccc}
\toprule[1.2pt]
\multirow{2}{*}{\textbf{Scene}} & 
\multicolumn{4}{c|}{\textbf{PSNR} $\uparrow$} & 
\multicolumn{4}{c|}{\textbf{SSIM} $\uparrow$} & 
\multicolumn{4}{c}{\textbf{LPIPS} $\downarrow$} \\
\cmidrule(lr){2-5} \cmidrule(lr){6-9} \cmidrule(l){10-13}
& 3DGS & CFGS & GSHT & Ours &
3DGS & CFGS & GSHT & Ours &
3DGS & CFGS & GSHT & Ours \\ 
\midrule[0.8pt]
Cd       & \textbf{28.29} & 26.60 & 26.44 & \underline{28.18} & \textbf{0.94} & 0.87 & 0.90 & \textbf{0.94} & \textbf{0.12} & 0.17 & 0.16 & \textbf{0.12} \\
Giants   & \textbf{21.69} & 14.37 & 16.01 & \underline{20.52} & \textbf{0.72} & 0.45 & 0.36 & \underline{0.68} & \underline{0.28} & 0.47 & 0.62 & \textbf{0.25} \\
Lab      & \underline{28.54} & 26.26 & 27.99 & \textbf{29.28} & \underline{0.93} & 0.82 & 0.91 & \textbf{0.94} & \textbf{0.15} & 0.18 & \textbf{0.15} & \textbf{0.15} \\
Tools    & \textbf{24.59} & 12.44 & 11.67 & \underline{24.33} & \textbf{0.84} & 0.50 & 0.47 & \underline{0.83} & \underline{0.35} & 0.59 & 0.60 & \textbf{0.32} \\
\midrule[0.8pt]
Mean     & \textbf{25.77} & 18.51 & 19.05 & \underline{25.58} & \textbf{0.86} & 0.58 & 0.59 & \underline{0.85} & \underline{0.23} & 0.39 & 0.43 & \textbf{0.21} \\
\bottomrule[1.2pt]
\end{tabular}
}

\label{table:nvs_shiny}
\vspace{-1.5em}
\end{table*}

\newcommand{\stackgap}{3pt}

\newcommand{\twostack}[2]{%
  \begin{minipage}[t]{\linewidth}
    \centering
    \includegraphics[width=\linewidth]{#1}\par\vspace{\stackgap}%
    \includegraphics[width=\linewidth]{#2}%
  \end{minipage}%
}
\begin{figure*}[t]
\centering

% ---------- Row 1 ----------
\begin{minipage}[b]{0.30\textwidth}
  \centering
  \raisebox{-20mm}{% 负号=向下移动，自己调 6mm/10mm/12mm/15mm
    \subfloat[GT]{%
      \includegraphics[width=\linewidth]{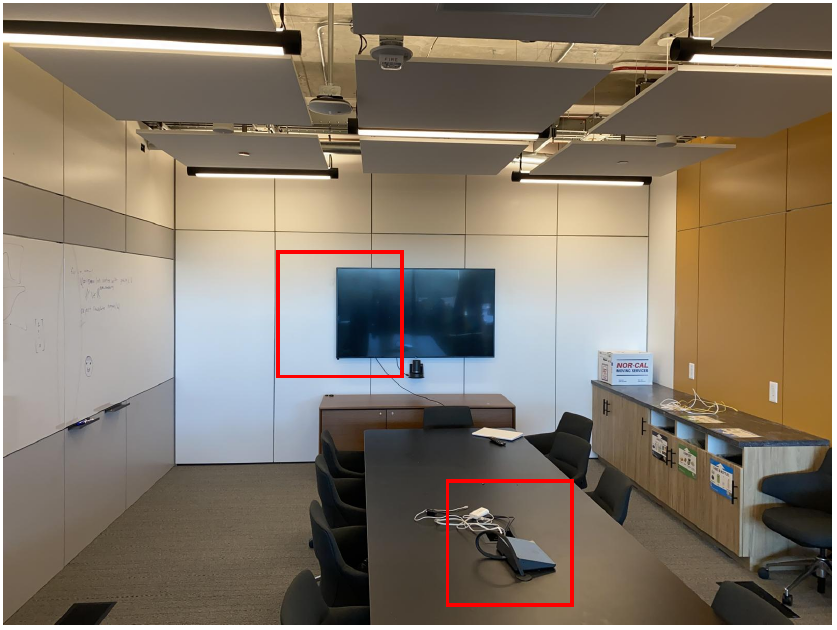}%
    }%
  }
\end{minipage}\hfill
\begin{minipage}[b]{0.69\textwidth}
  \vspace{4mm} % 右侧整体下移（3~6mm 之间试）
  \centering
  \begin{minipage}[b]{0.235\linewidth}
    \centering
    \subfloat[ 3DGS]{\twostack
      {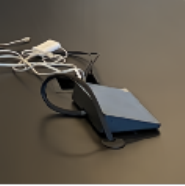} % 上：桌面
      {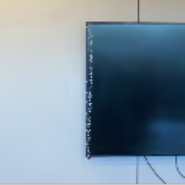}}% 下：电视
  \end{minipage}\hfill
  \begin{minipage}[b]{0.235\linewidth}
    \centering
    \subfloat[ CFGS]{\twostack
      {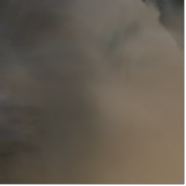}
      {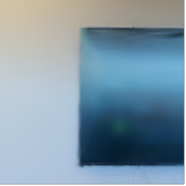}}
  \end{minipage}\hfill
  \begin{minipage}[b]{0.235\linewidth}
    \centering
    \subfloat[ GSHT]{\twostack
      {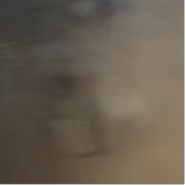}
      {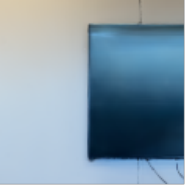}}
  \end{minipage}\hfill
  \begin{minipage}[b]{0.235\linewidth}
    \centering
    \subfloat[ Ours]{\twostack
      {figs/detail_nvs/all_figs/room\_ours\_2.pdf}
      {figs/detail_nvs/all_figs/room\_ours\_1.pdf}}
  \end{minipage}
\end{minipage}

\vspace{6pt}

% ---------- Row 2 ----------
\begin{minipage}[b]{0.30\textwidth}
  \centering
  \raisebox{-15mm}{% 负号=向下移动，自己调 6mm/10mm/12mm/15mm
    \subfloat[GT ]{%
      \includegraphics[width=\linewidth]{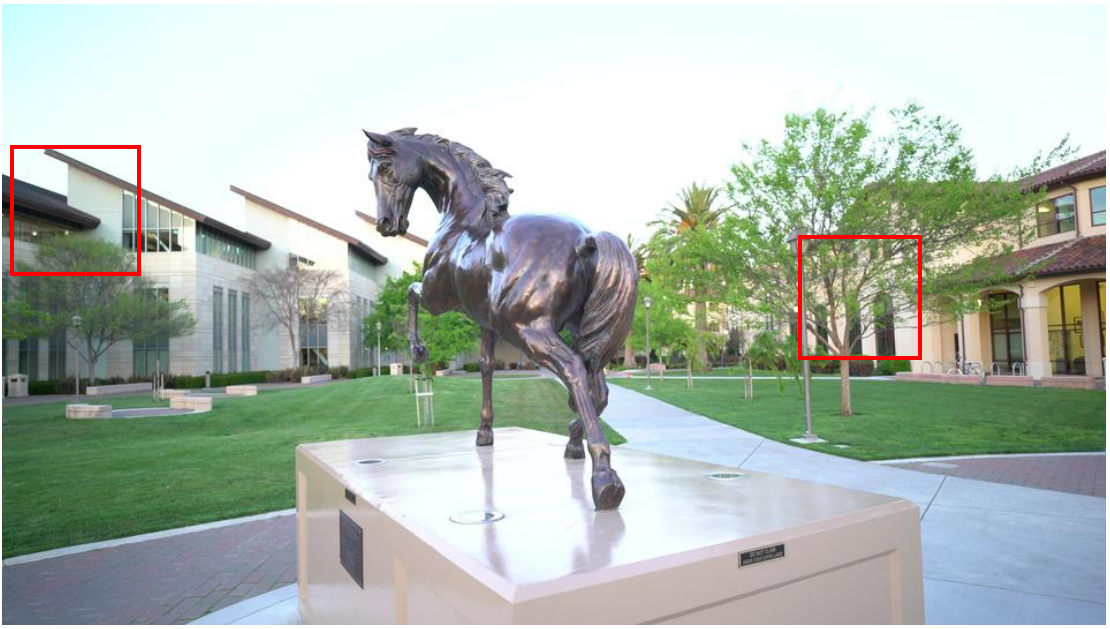}%
    }%
  }

\end{minipage}\hfill
\begin{minipage}[b]{0.69\textwidth}
  \vspace{4mm}
  \centering
  \begin{minipage}[b]{0.235\linewidth}
    \centering
    \subfloat[ 3DGS]{\twostack
      {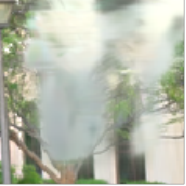}
      {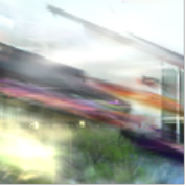}}
  \end{minipage}\hfill
  \begin{minipage}[b]{0.235\linewidth}
    \centering
    \subfloat[ CFGS]{\twostack
      {figs/detail_nvs/all_figs/horse\_cf\_2.pdf}
      {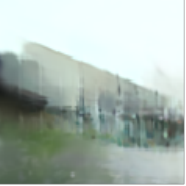}}
  \end{minipage}\hfill
  \begin{minipage}[b]{0.235\linewidth}
    \centering
    \subfloat[ GSHT]{\twostack
      {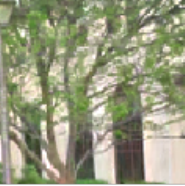}
      {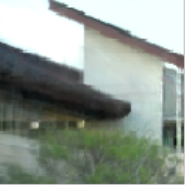}}
  \end{minipage}\hfill
  \begin{minipage}[b]{0.235\linewidth}
    \centering
    \subfloat[ Ours]{\twostack
      {figs/detail_nvs/all_figs/horse\_ours\_2.pdf}
      {figs/detail_nvs/all_figs/horse\_ours\_1.pdf}}
  \end{minipage}
\end{minipage}

\caption{\textbf{Comparison of fine-grained scene details.}
Our method preserves sharper textures in novel-view synthesis, benefiting from camera pose optimization during training.}
\label{fig:detail_compare}
\vspace{-4mm}
\end{figure*}

\subsection{Evaluation Metrics}
\label{Evaluation Metrics}
 We employed the same evaluation metrics as   CFGS~\cite{Fu_2024_CVPR} and GSHT~\cite{ji2024sfmfree3dgaussiansplatting}. For novel view synthesis, we used standard metrics: \textit{peak signal-to-noise ratio} (PSNR), \textit{structural similarity index measure} (SSIM)~\cite{Wang2004ImageQA} and \textit{learned perceptual image patch similarity} (LPIPS)~\cite{zhang2018unreasonableeffectivenessdeepfeatures}. 
For pose evaluation, we treated COLMAP-estimated poses as ground truth and measured  \textit{absolute trajectory error} (ATE), which includes \textit{relative rotation error} (RPEr) and \textit{relative translation error} (RPEt), along with \textit{relative pose error} (RPE). ATE quantifies the discrepancy between estimated camera positions and ground truth, while RPE measures relative pose errors between image pairs.

\begin{figure*}[htbp]
\centering
\captionsetup{font=small, skip=2pt}
\setlength{\tabcolsep}{3pt}
\begin{tabular}{@{}r@{\hspace{5pt}}cccccc@{}}
\multirow{3}{*}[1.3cm]{\rotatebox{90}{\small CFGS}} &
\includegraphics[width=0.155\linewidth]{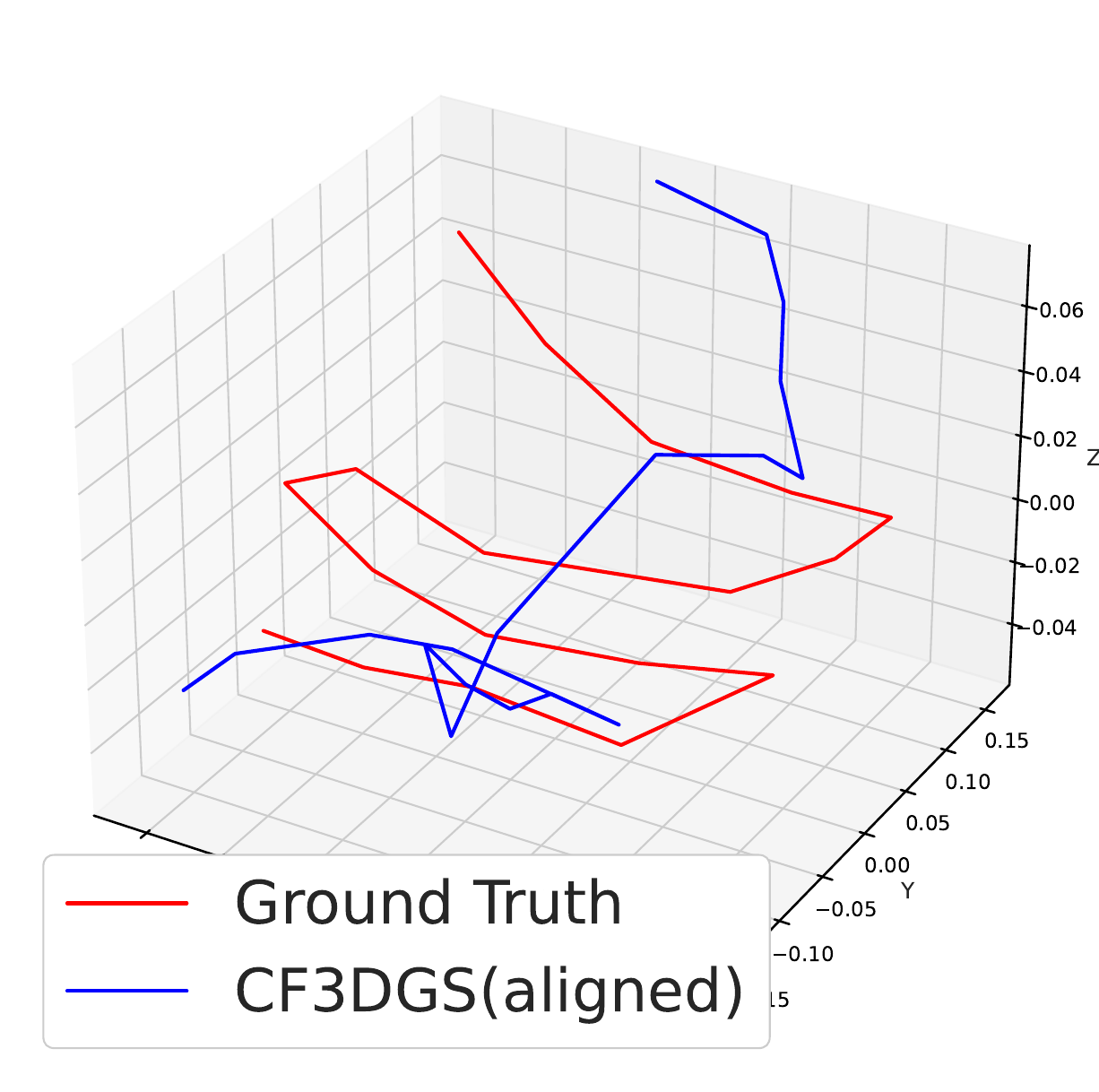} &
\includegraphics[width=0.155\linewidth]{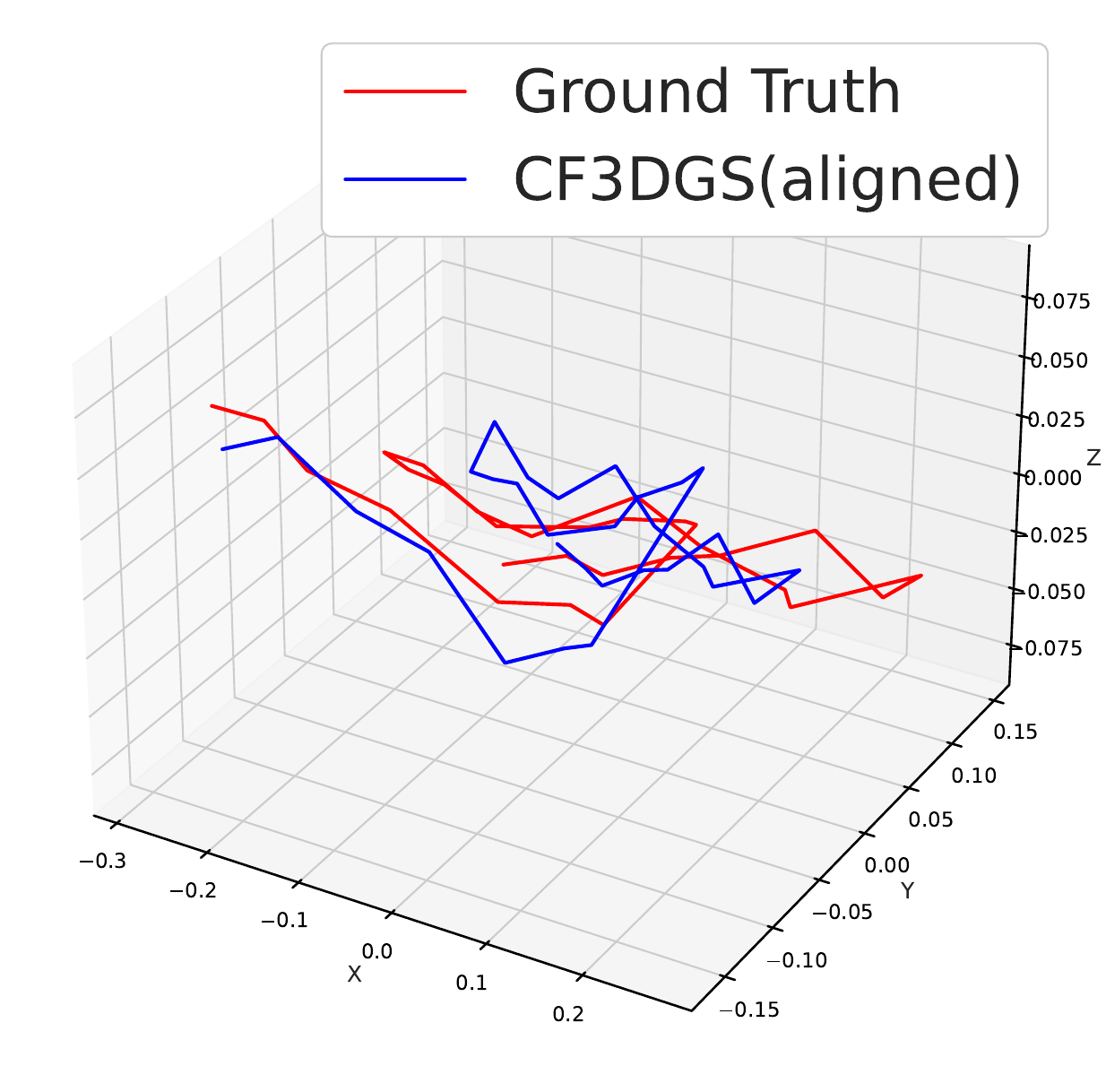} &
\includegraphics[width=0.155\linewidth]{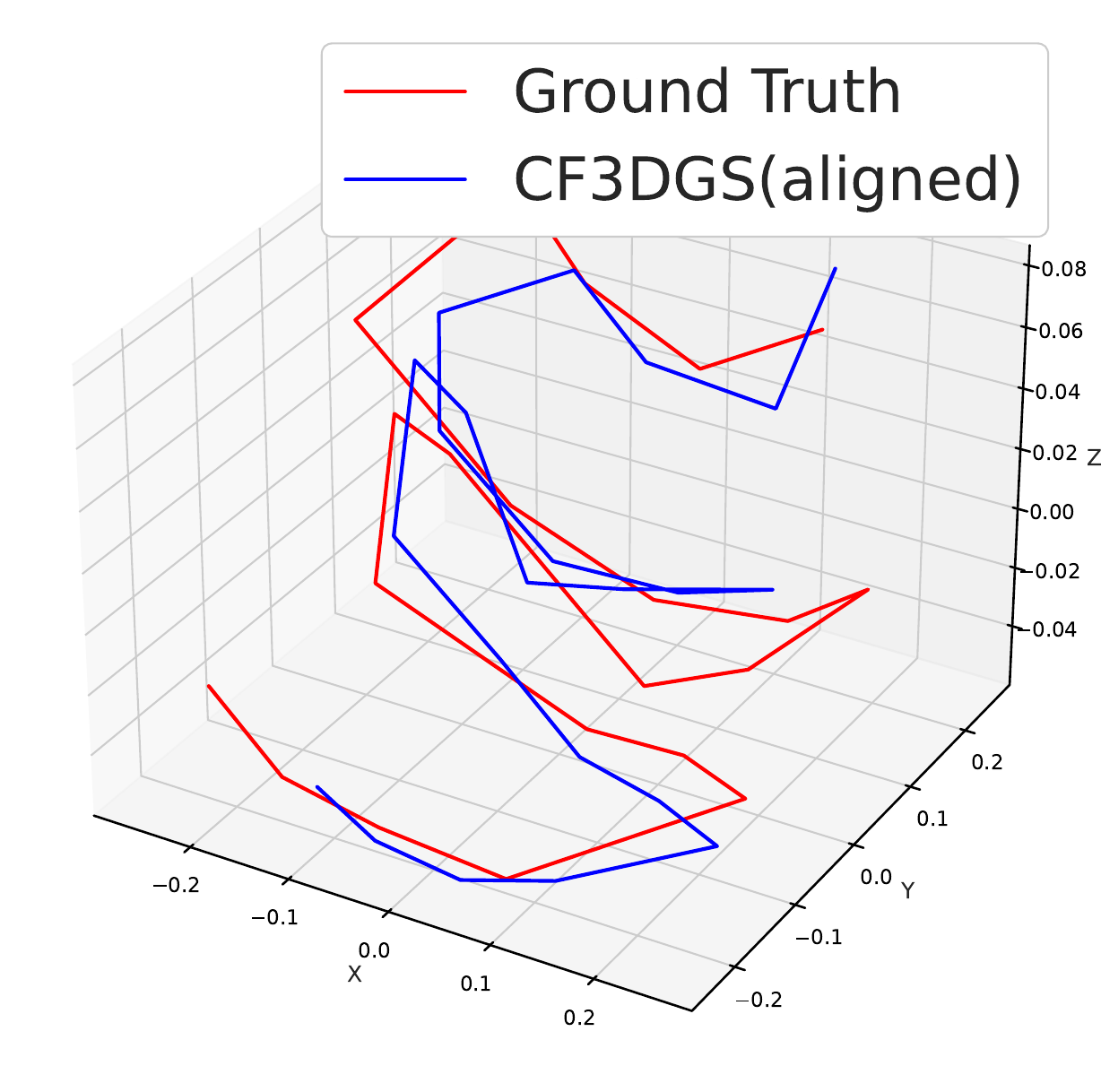} &
\includegraphics[width=0.155\linewidth]{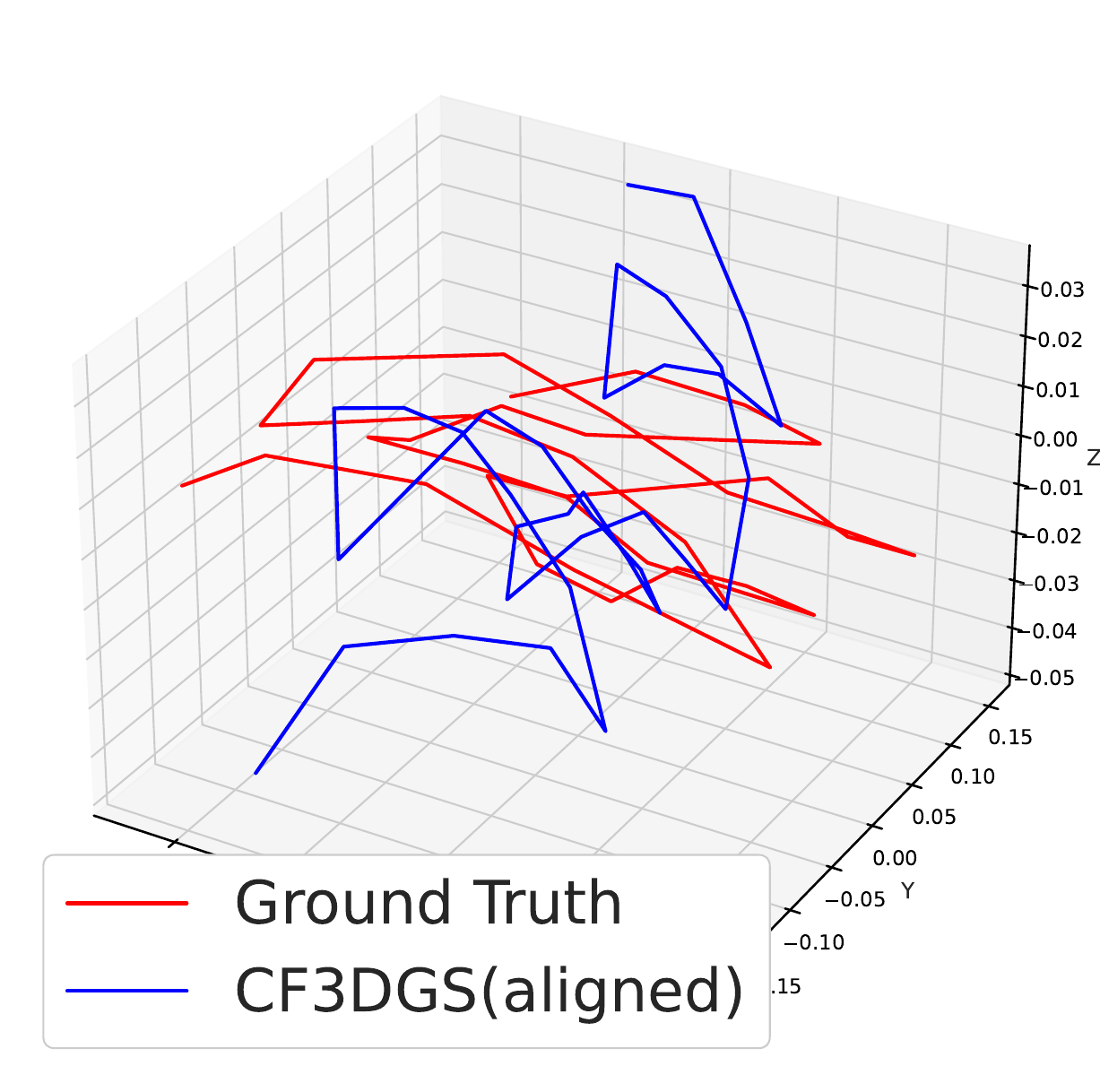} &
\includegraphics[width=0.155\linewidth]{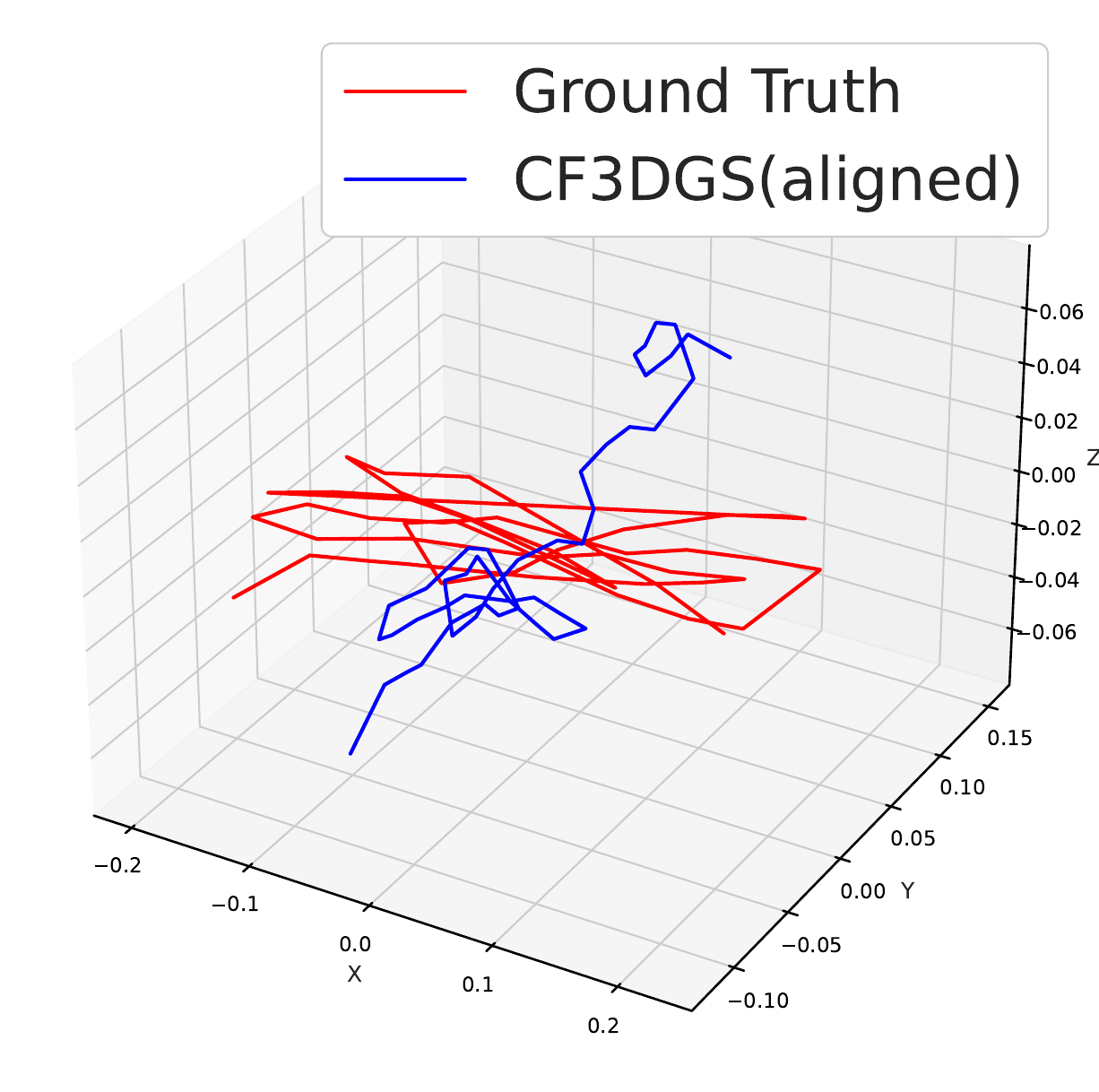} &
\includegraphics[width=0.155\linewidth]{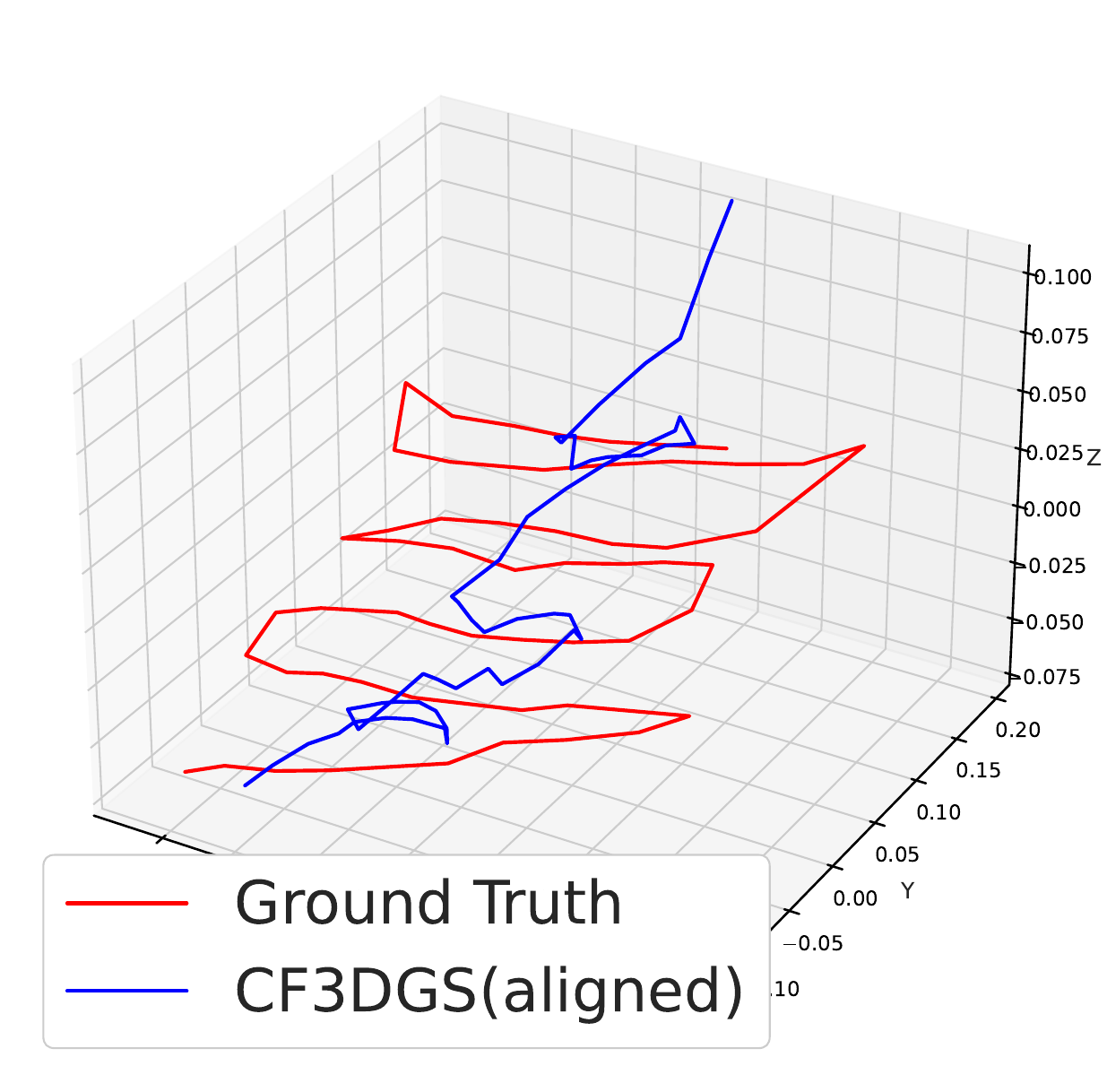} \\
\multirow{3}{*}[1.3cm]{\rotatebox{90}{\small GSHT}} &
\includegraphics[width=0.155\linewidth]{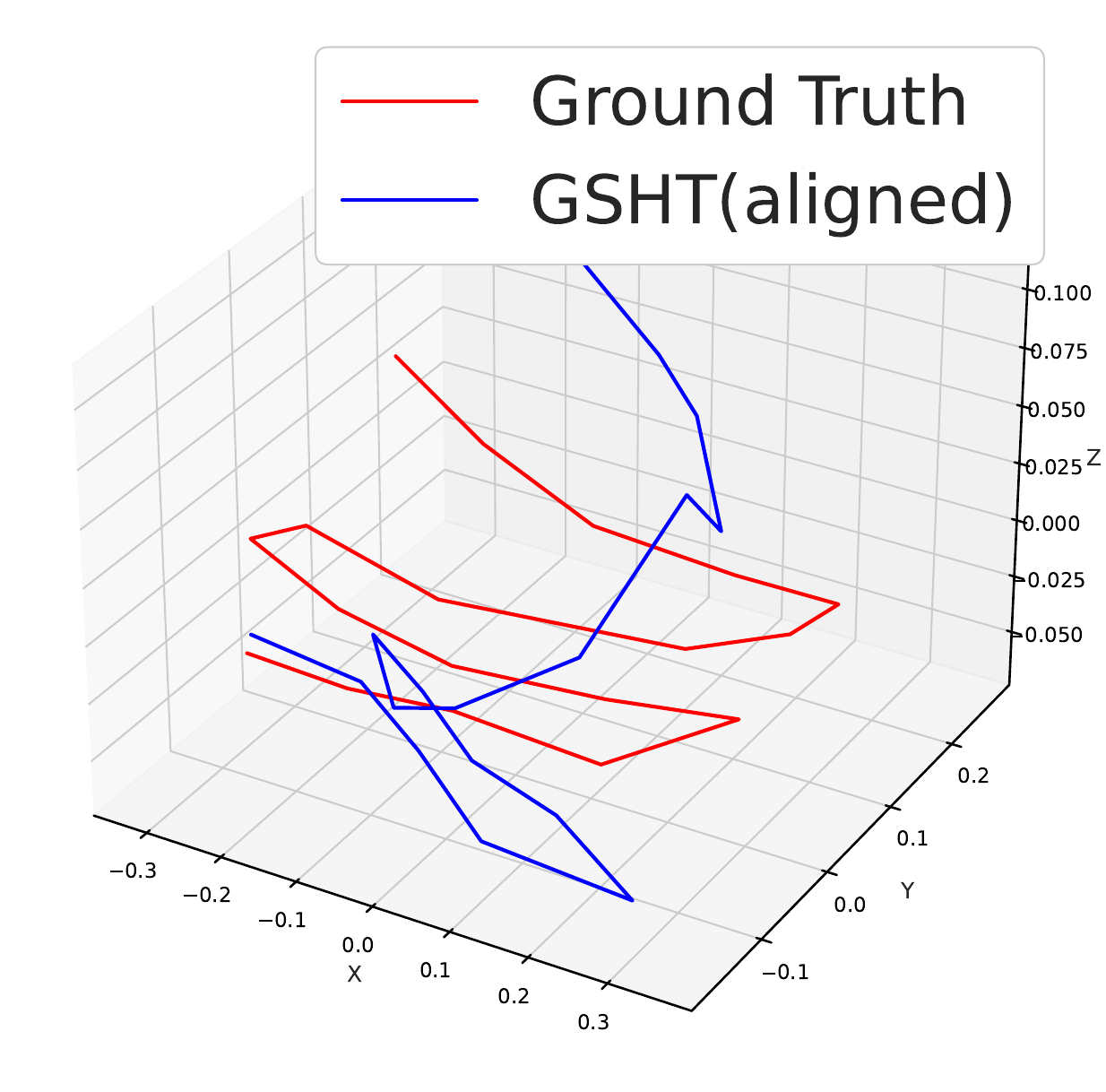} &
\includegraphics[width=0.155\linewidth]{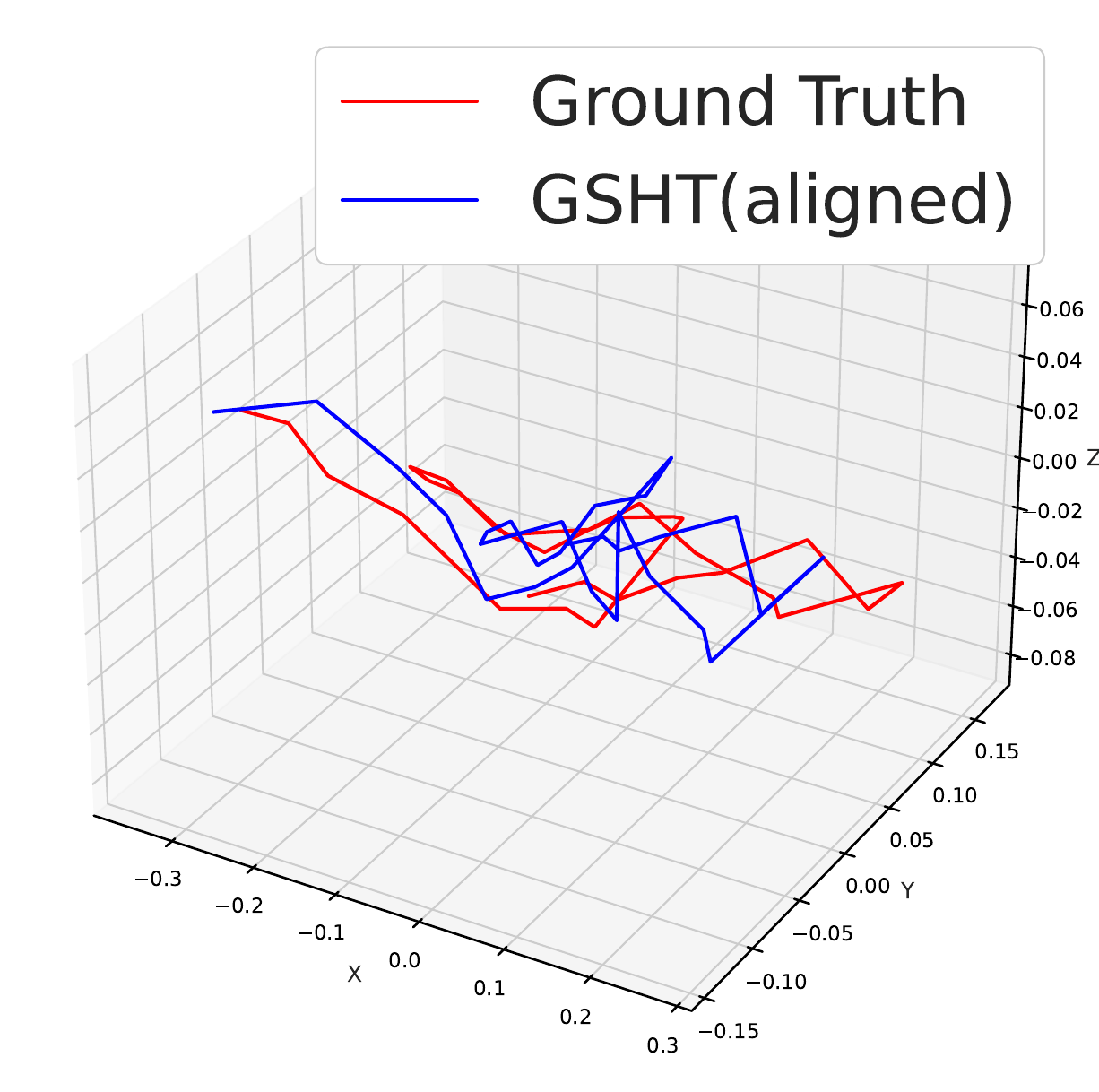} &
\includegraphics[width=0.155\linewidth]{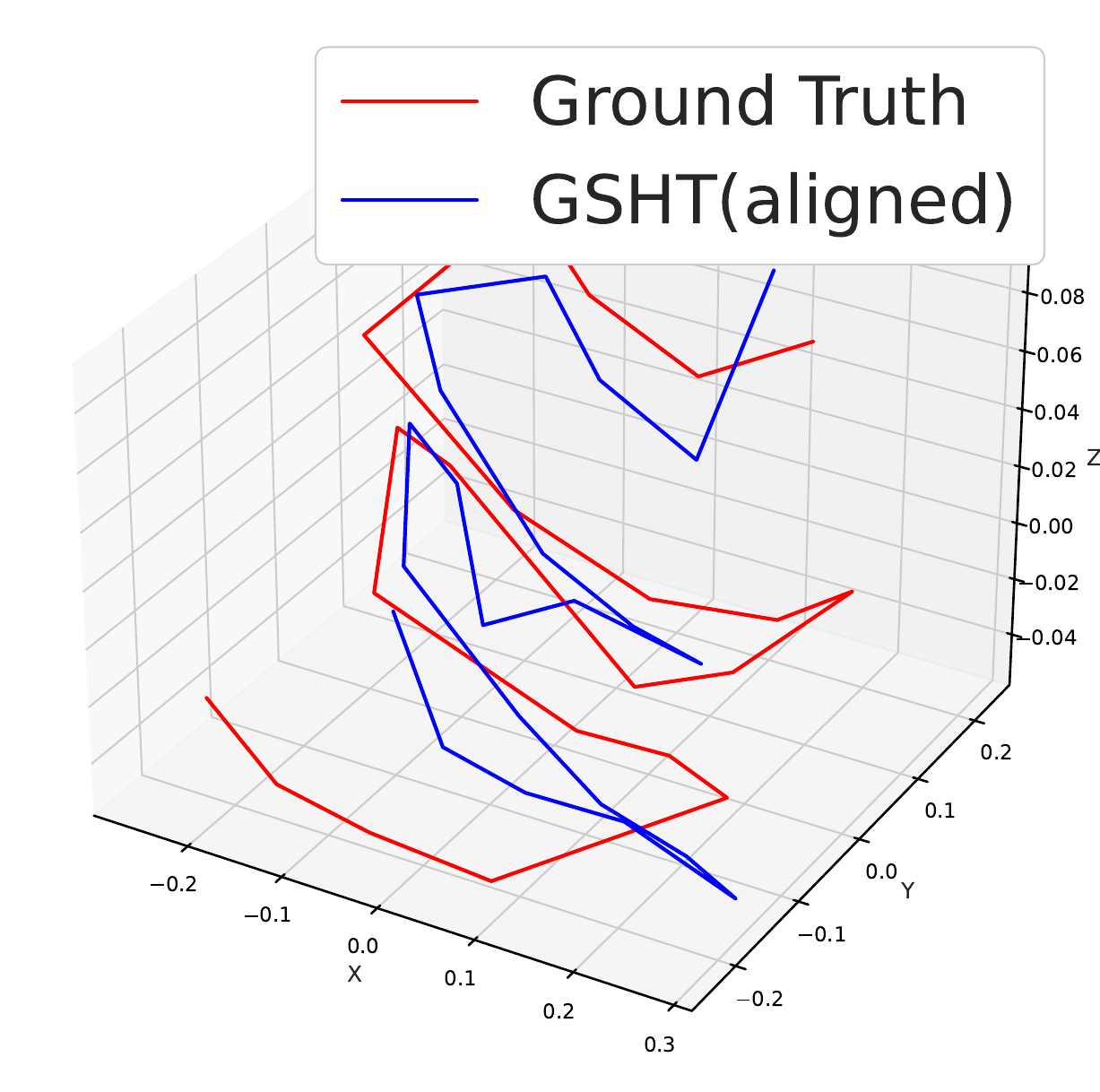} &
\includegraphics[width=0.155\linewidth]{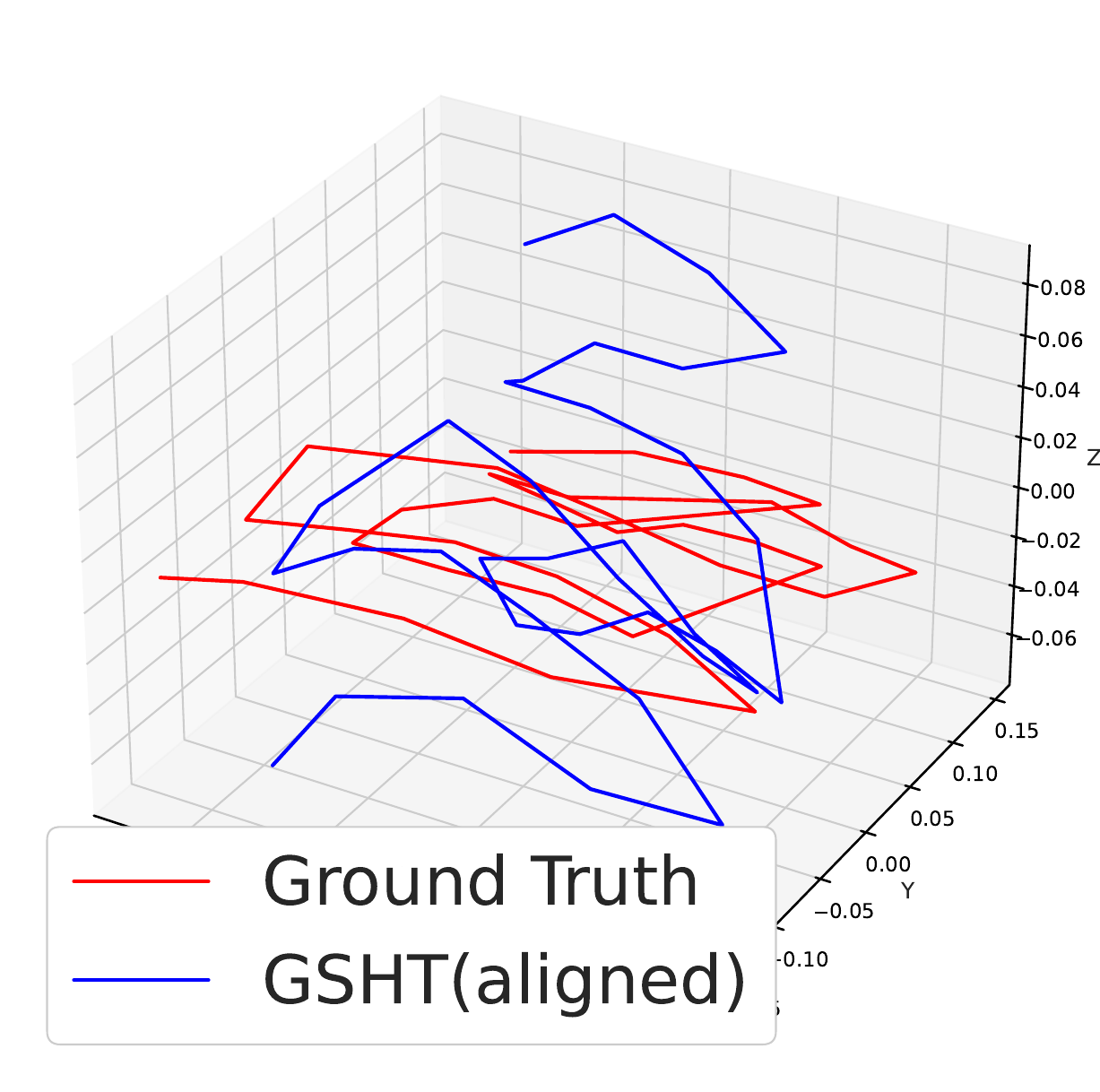} &
\includegraphics[width=0.155\linewidth]{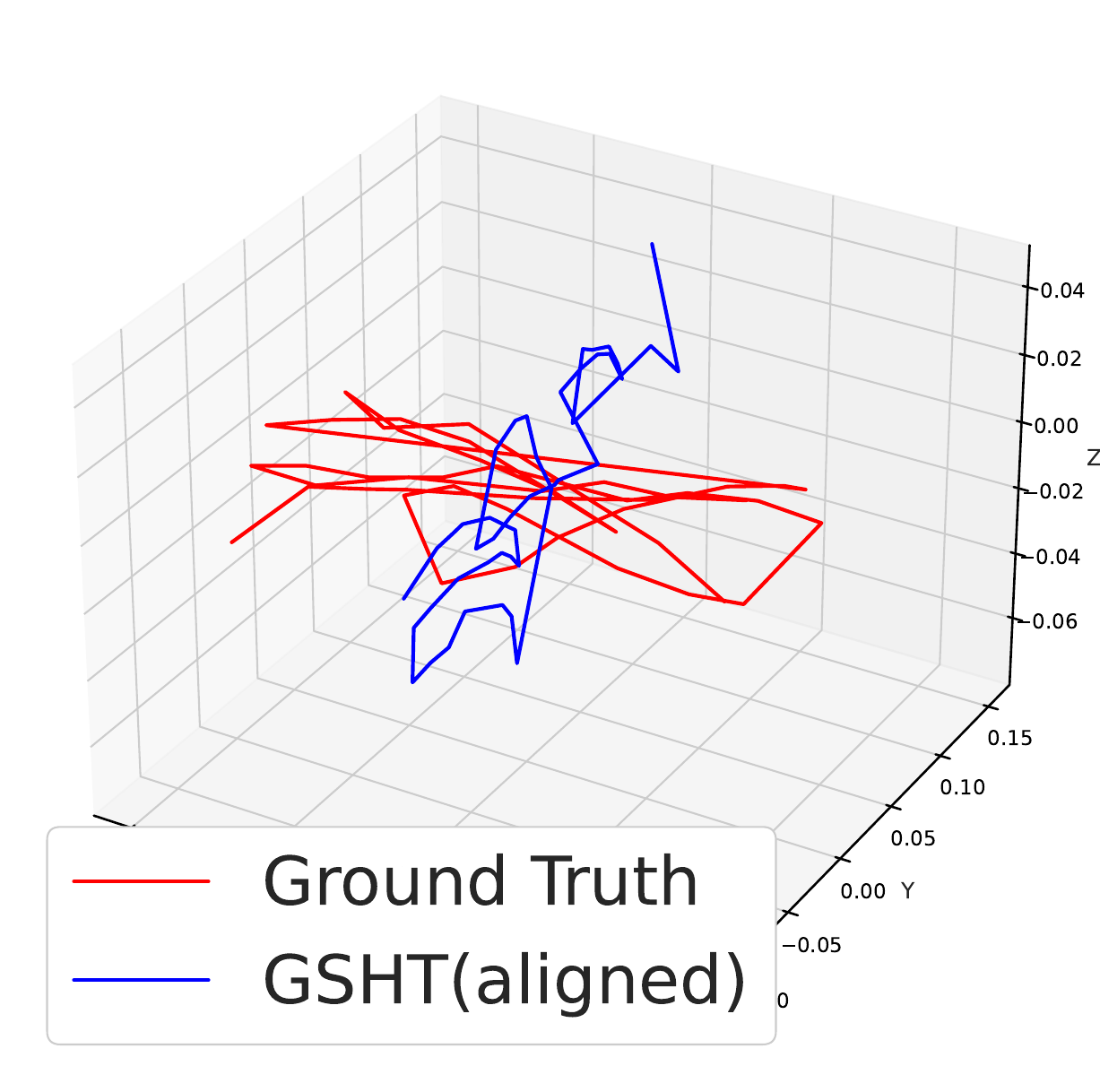} &
\includegraphics[width=0.155\linewidth]{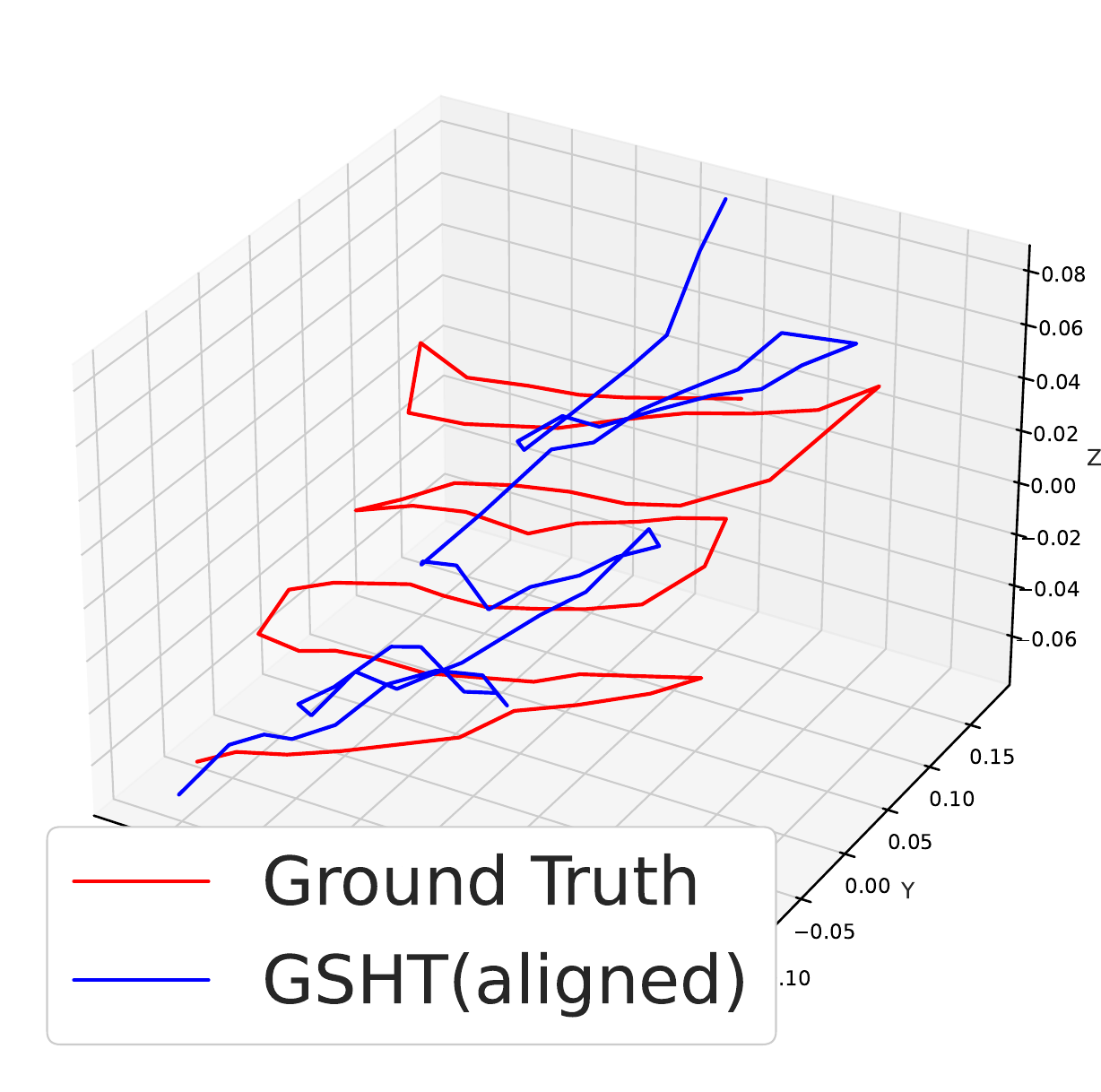} \\
\multirow{3}{*}[1.3cm]{\rotatebox{90}{\small Ours}} &
\includegraphics[width=0.155\linewidth]{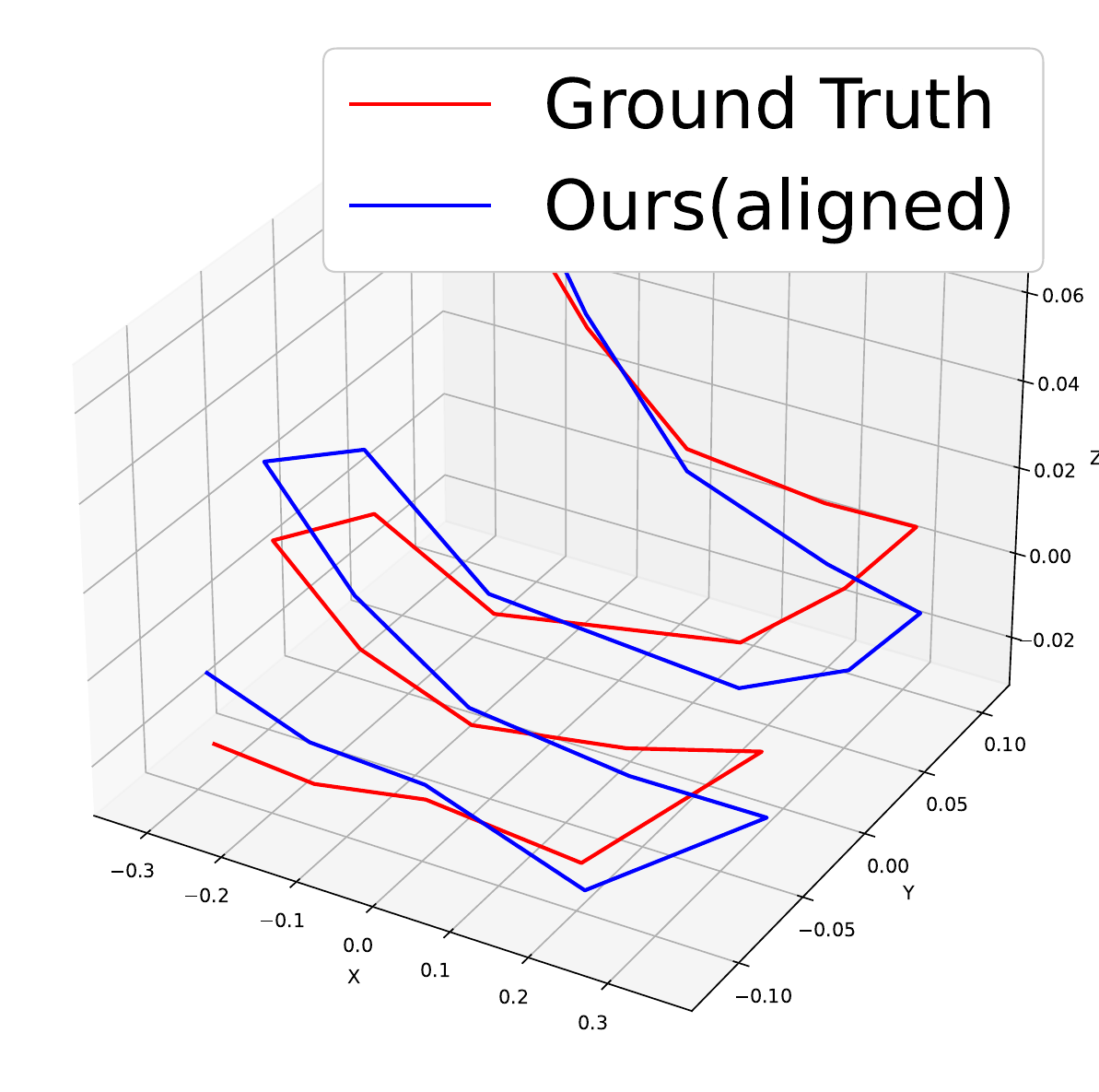} &
\includegraphics[width=0.155\linewidth]{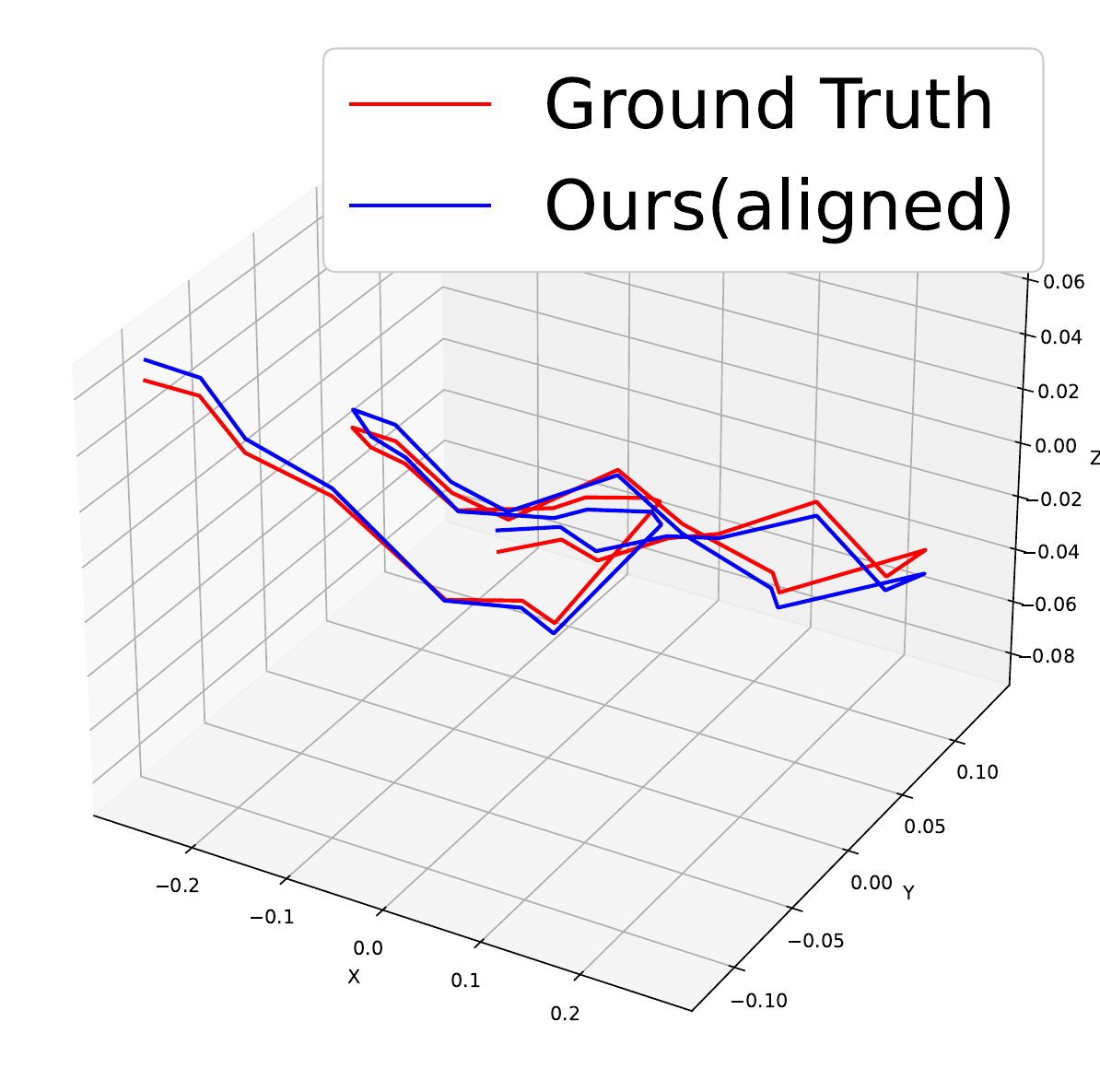} &
\includegraphics[width=0.155\linewidth]{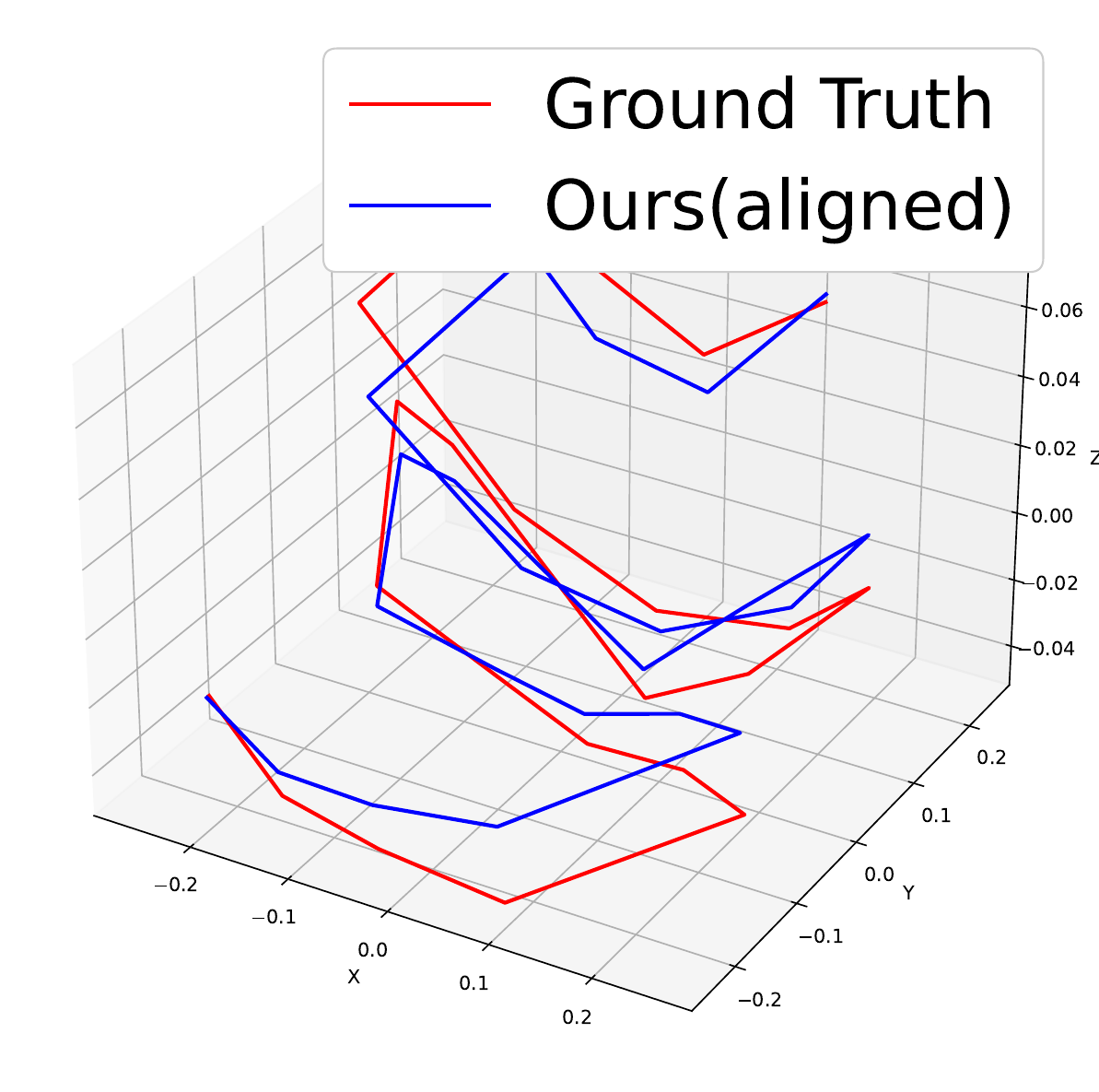} &
\includegraphics[width=0.155\linewidth]{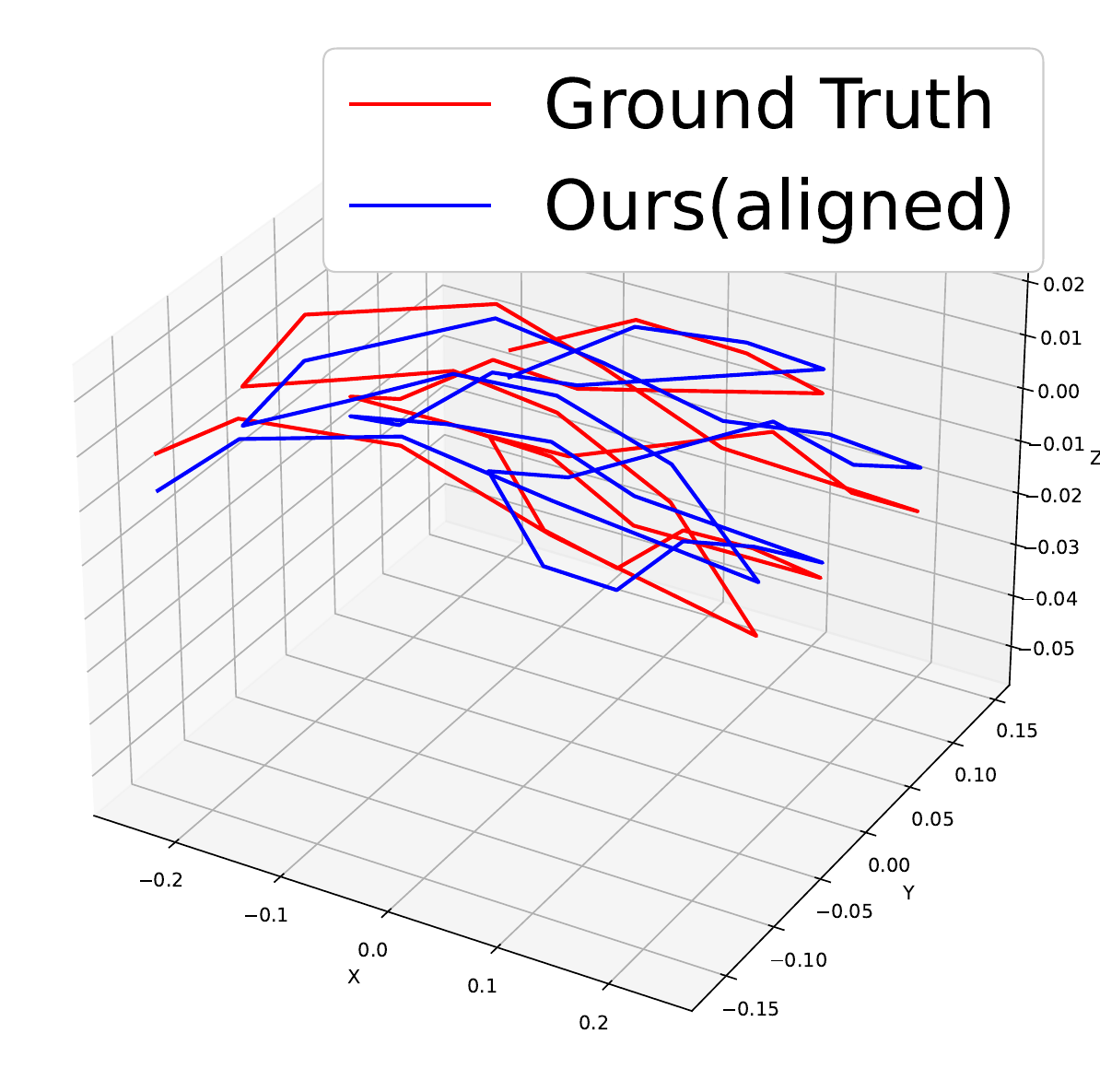} &
\includegraphics[width=0.155\linewidth]{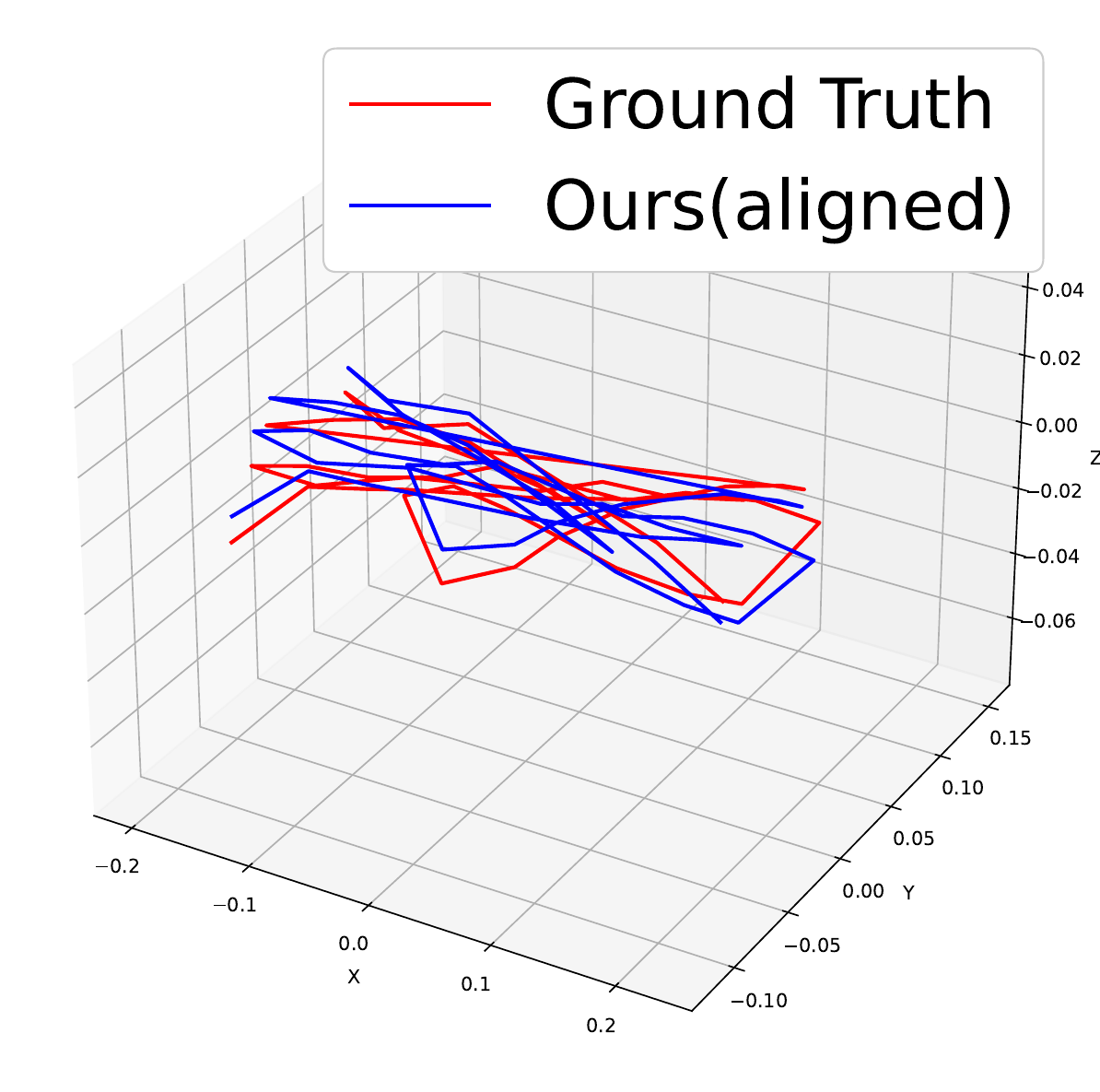} &
\includegraphics[width=0.155\linewidth]{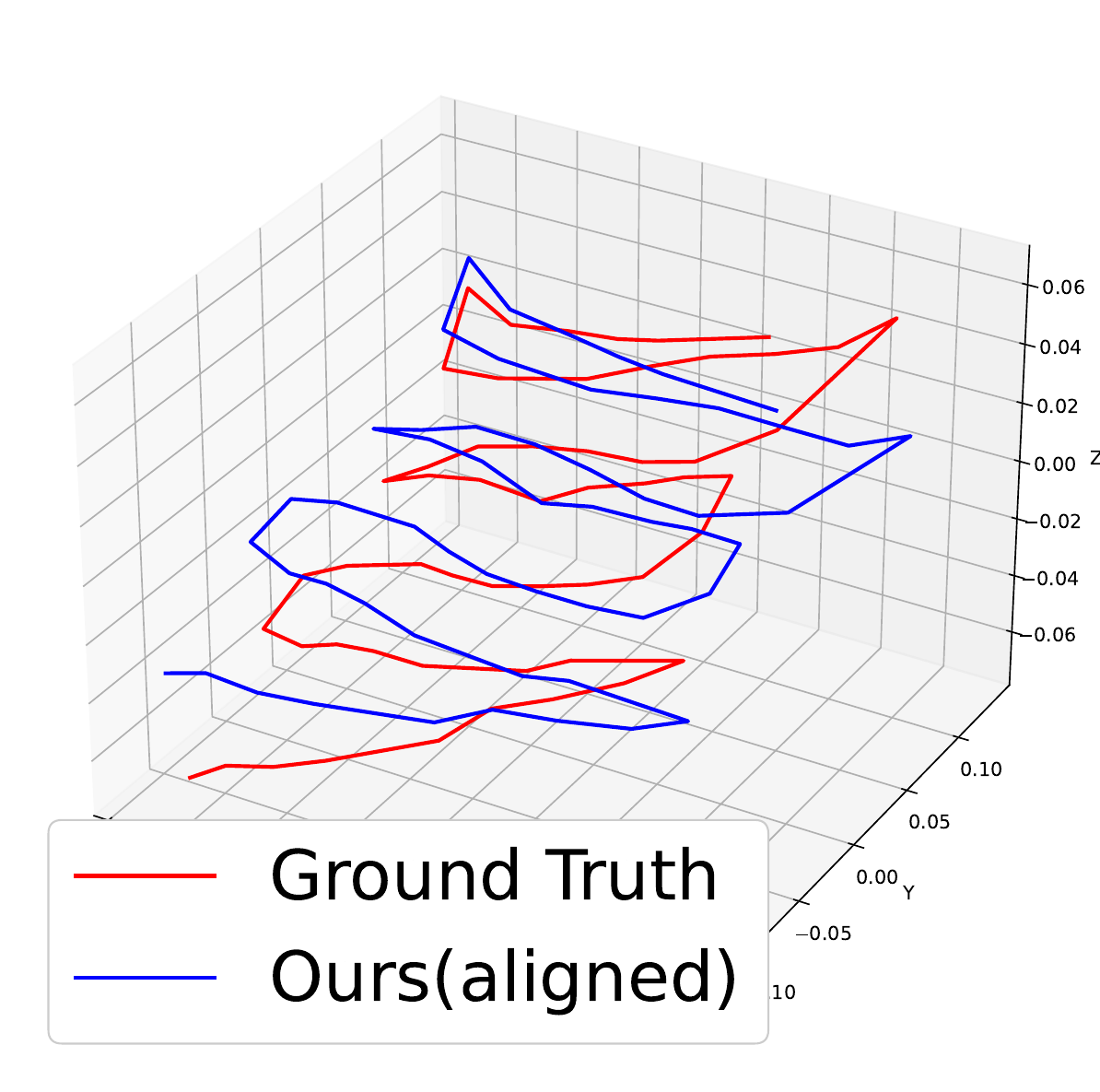} \\
& \small Fern & \small Flower & \small Orchids & \small Room & \small Trex & \small Horns \\
\end{tabular}
\caption{Trajectory comparison on the LLFF dataset. Each row represents a method (CFGS, GSHT and Ours), and each column represents a scene.
}
\label{fig:pose_comparison}
\vspace{-2mm}
\end{figure*}
\begin{table*}[htbp]
\centering
\small
\renewcommand{\arraystretch}{1.0}
\setlength{\tabcolsep}{6pt}
\captionsetup{font=small, skip=2pt}
\caption{\textbf{Pose estimation performance comparison} on LLFF dataset. }
\label{tab:pose_comparison}
\begin{tabular}{c|ccc|ccc|ccc}
\toprule[1.2pt]
\multirow{2}{*}{\textbf{Scene}} &
\multicolumn{3}{c|}{\textbf{RPE\textsubscript{trans} ↓}} &
\multicolumn{3}{c|}{\textbf{RPE\textsubscript{rot} ↓}} &  
\multicolumn{3}{c}{\textbf{ATE ↓}} \\
\cmidrule(lr){2-4} \cmidrule(lr){5-7} \cmidrule(l){8-10}
& CFGS & GSHT & Ours & CFGS & GSHT & Ours & CFGS & GSHT & Ours \\
\midrule[0.8pt]
Fern     & 8.908 & \underline{6.656} & \textbf{0.146} & 2.830 & \underline{2.349} & \textbf{0.039} & 0.161 & \underline{0.129} & \textbf{0.014} \\
Flowers  & \underline{2.615} & 3.534 & \textbf{0.100} & \underline{0.148} & 0.229 & \textbf{0.052} & \underline{0.064} & 0.073 & \textbf{0.005} \\
Horns    & 3.395 & \underline{2.428} & \textbf{0.051} & 1.573 & \underline{1.310} & \textbf{0.027} & 0.088 & \underline{0.072} & \textbf{0.019} \\
Orchids  & \underline{3.586} & 4.170 & \textbf{0.135} & \underline{1.992} & 2.059 & \textbf{0.117} & \underline{0.074} & 0.098 & \textbf{0.018} \\
Room     & 5.290 & \underline{2.898} & \textbf{0.039} & 1.792 & \underline{1.675} & \textbf{0.030} & 0.117 & \underline{0.082} & \textbf{0.005} \\
Trex     & \underline{5.065} & 5.849 & \textbf{0.084} & \underline{1.901} & 2.112 & \textbf{0.026} & \underline{0.120} & 0.127 & \textbf{0.006} \\
\midrule[0.8pt]
Mean     & 4.810 & \underline{4.256} & \textbf{0.093} & 1.706 & \underline{1.622} & \textbf{0.049} & 0.104 & \underline{0.097} & \textbf{0.011} \\
\bottomrule[1.2pt]
\end{tabular}
\vspace{-3mm}
\end{table*}

\subsection{Implementation Details}
\label{Implementation Details}
 We initialized camera poses and sparse Gaussian points using only scene images and camera intrinsics. During 3D Gaussian reconstruction, we alternately optimized 3D Gaussian points and camera poses. Global pose optimization was performed every 100 iterations, limited to the first 15,000 iterations. This restricted optimization strategy prevents error accumulation, as pose estimation errors could degrade reconstruction quality, which might further corrupt pose estimation accuracy. Thus, optimizing poses only during the initial quarter of training iterations is empirically justified. 
All experiments were conducted on a single RTX 3090 GPU. Unless otherwise stated, our experiments follow the same 3DGS parameter settings. 
\begin{figure*}[t]
\centering
\captionsetup{font=small, skip=2pt}
\setlength{\tabcolsep}{2pt} % 紧凑列间距
\renewcommand{\arraystretch}{0.5}

\begin{tabular}{@{}r@{\hspace{5pt}}cccccccc@{}}
% 第一行：CFGS
\multirow{3}{*}[1.1cm]{\rotatebox{90}{\textbf{\small CFGS}}} &
\includegraphics[width=0.155\linewidth]{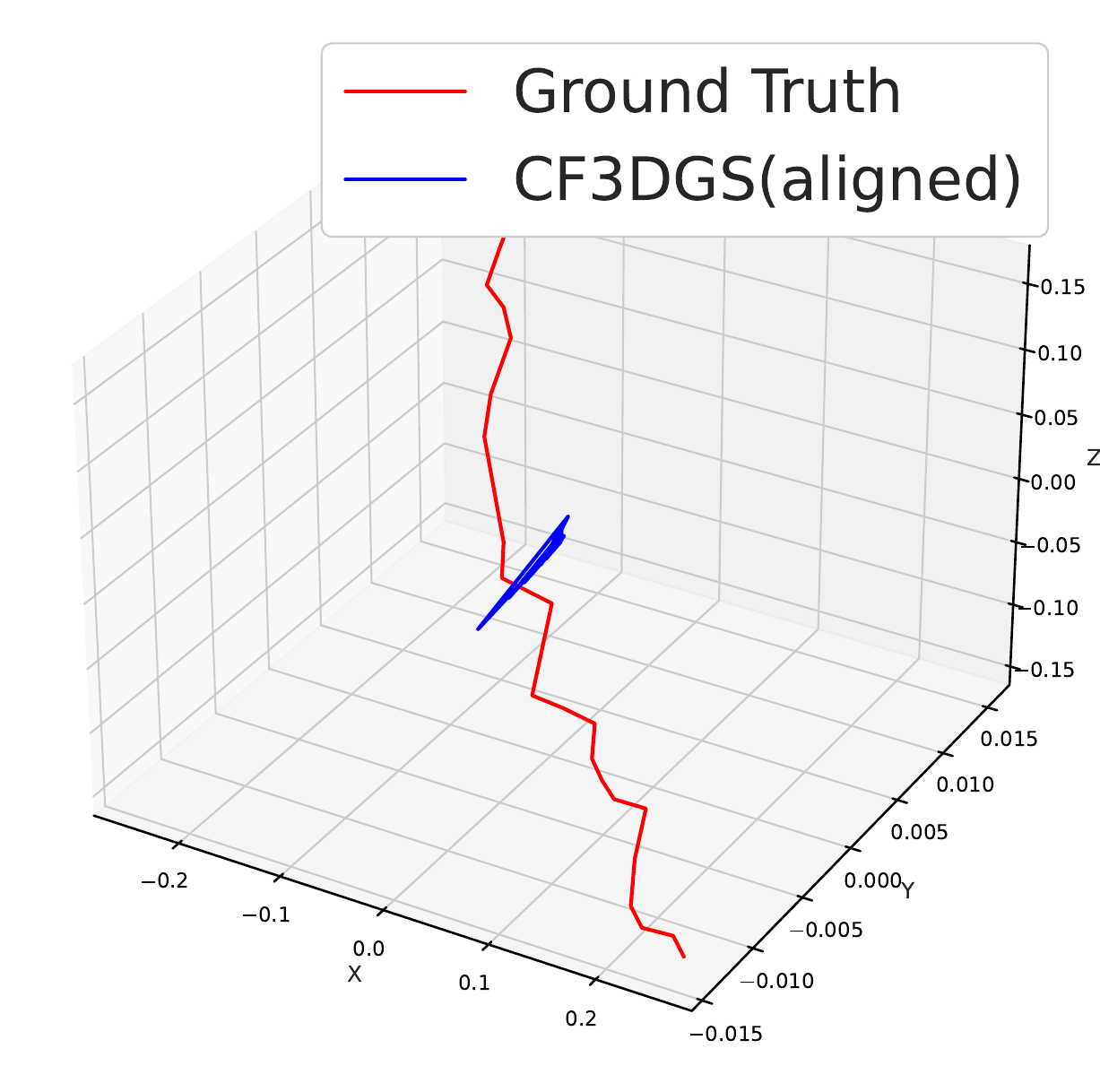} &
\includegraphics[width=0.155\linewidth]{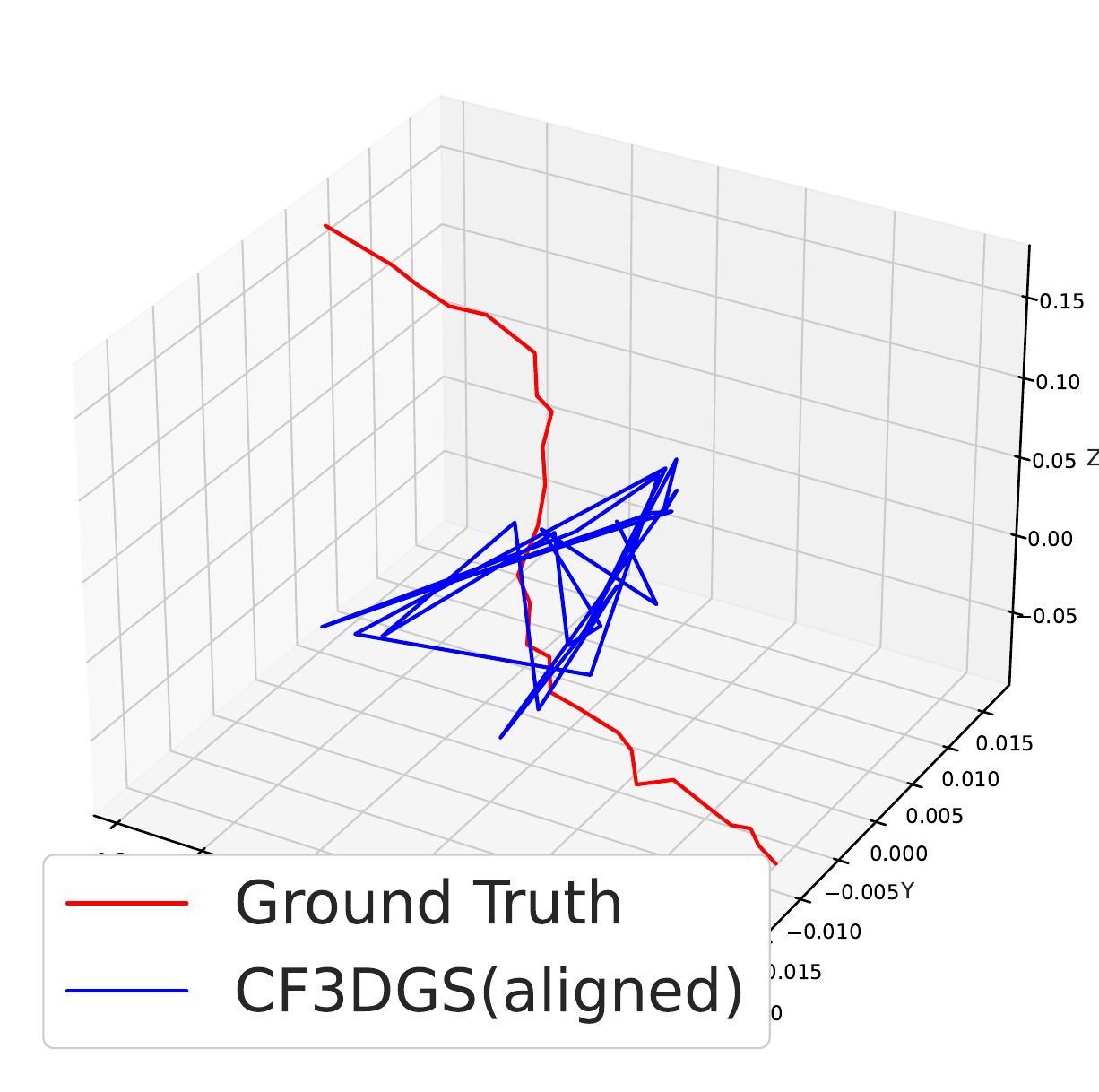} &
\includegraphics[width=0.155\linewidth]{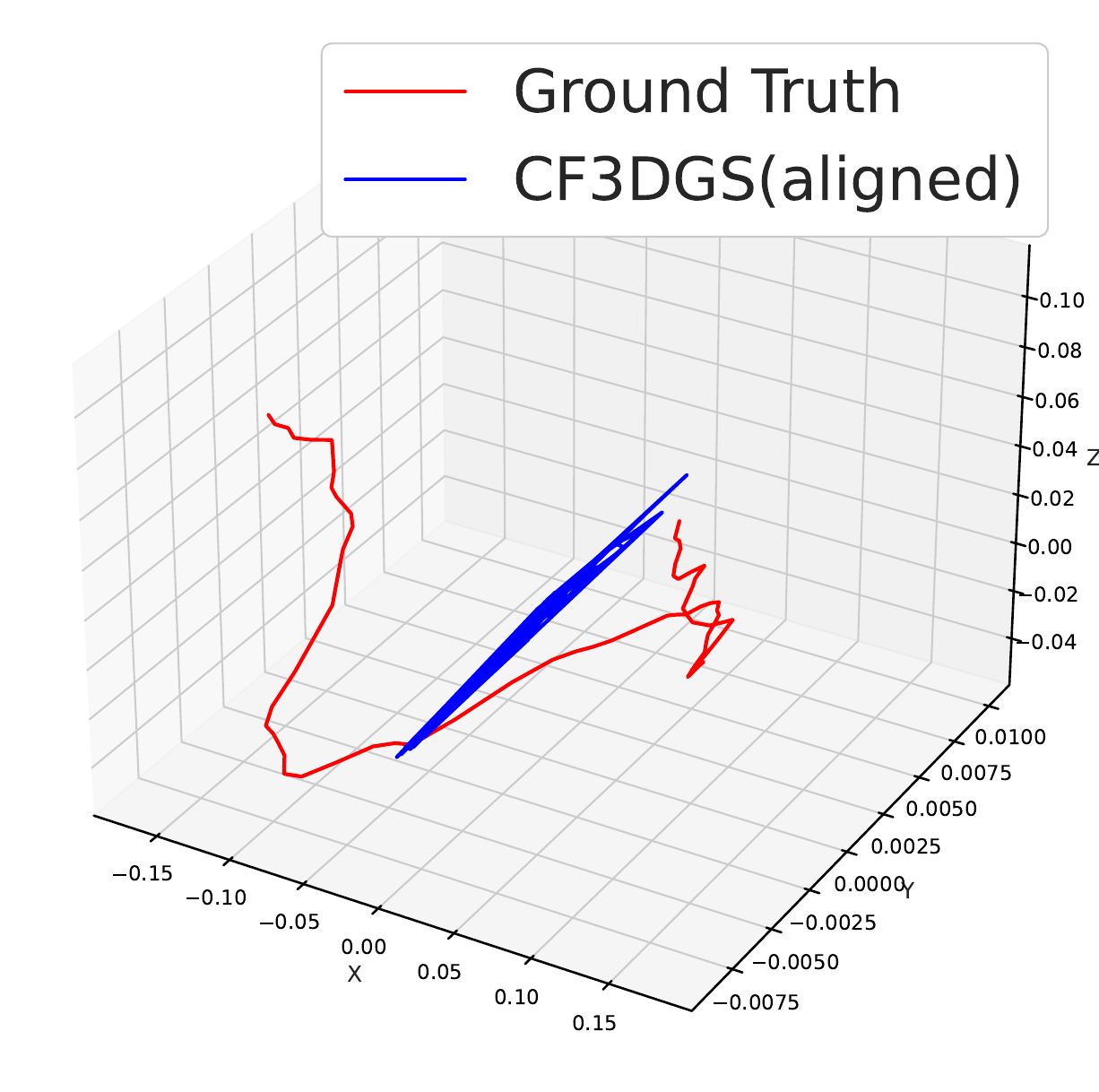} &
\includegraphics[width=0.155\linewidth]{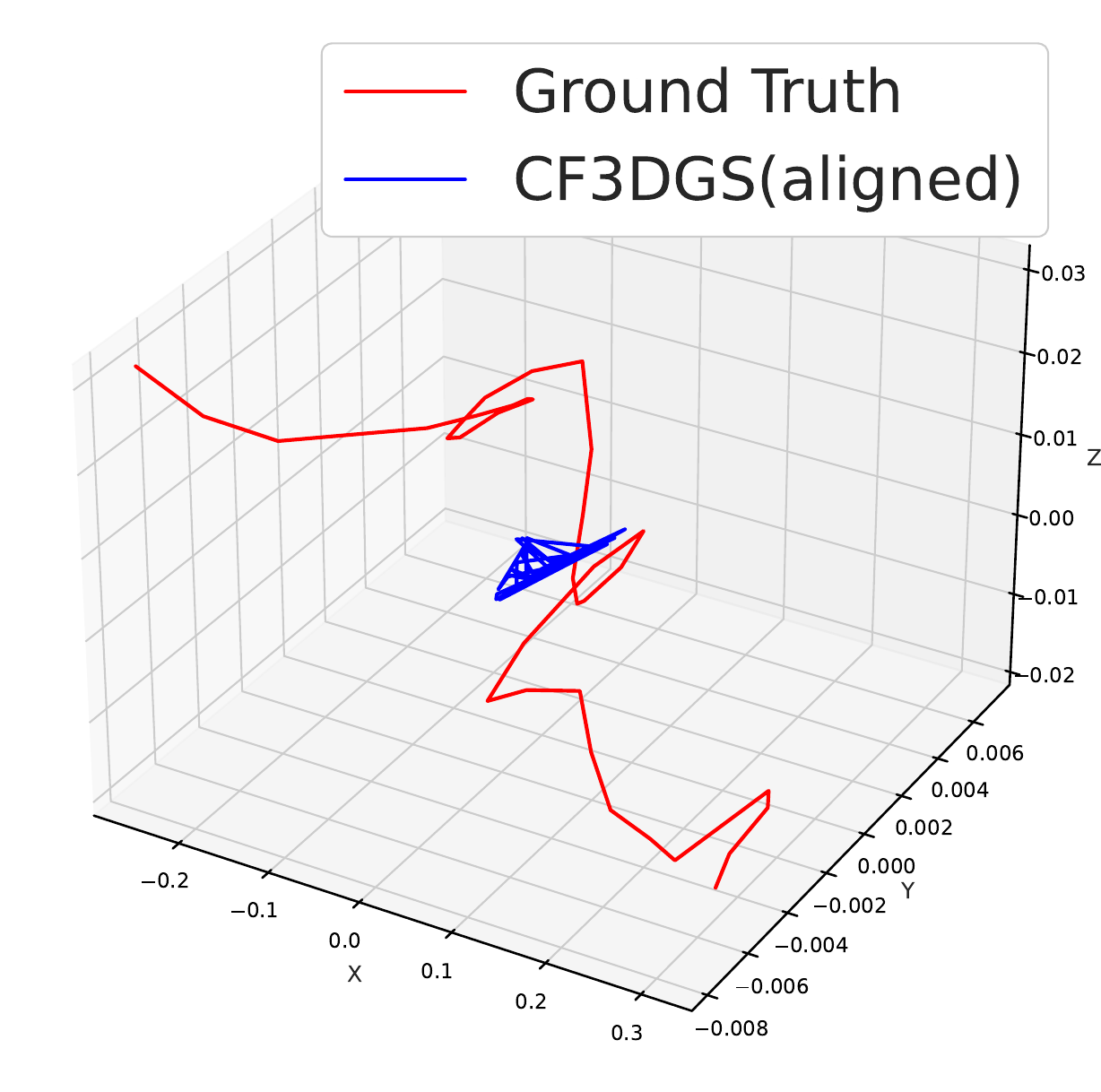} &
\includegraphics[width=0.155\linewidth]{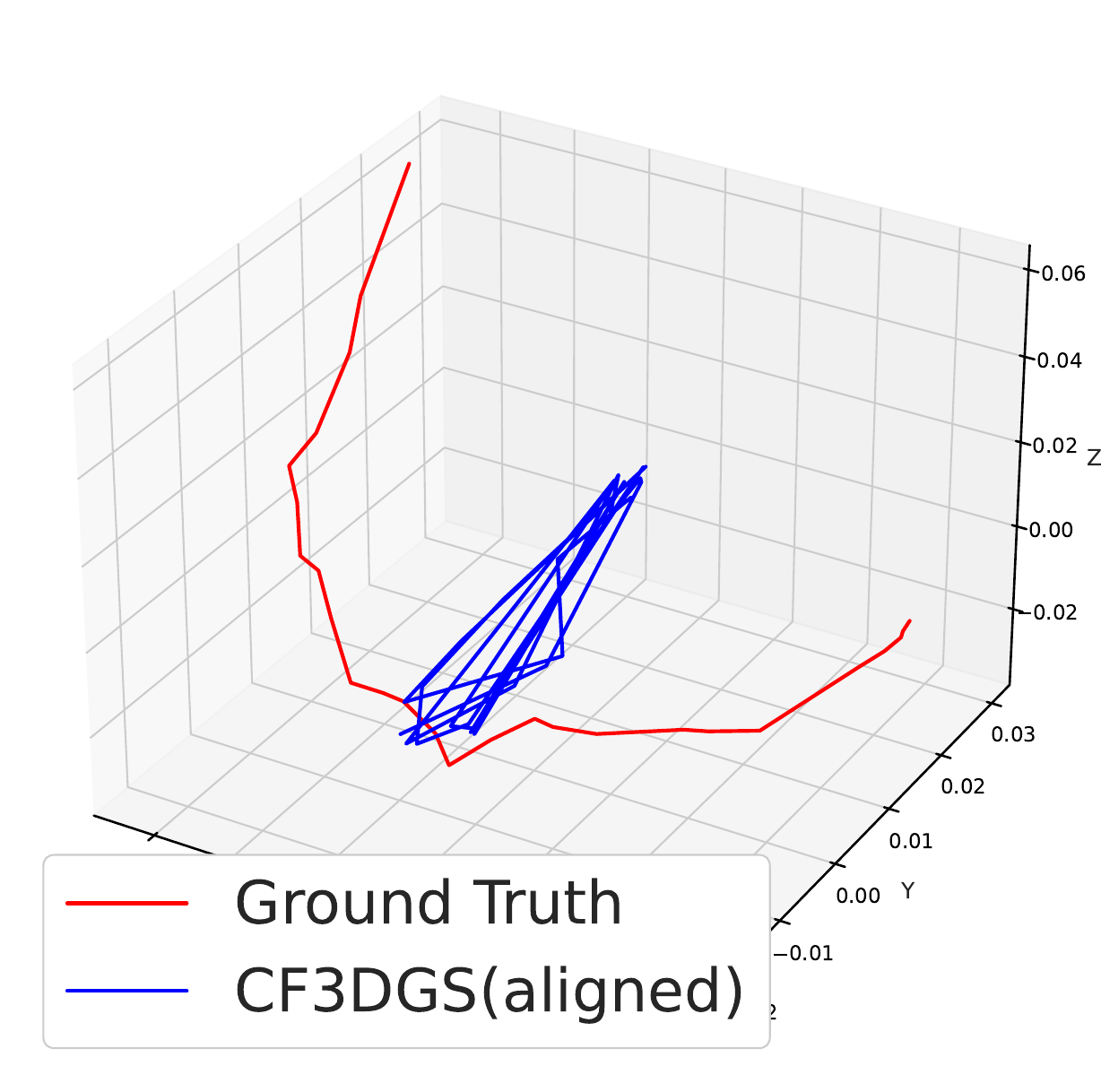} &
\includegraphics[width=0.155\linewidth]{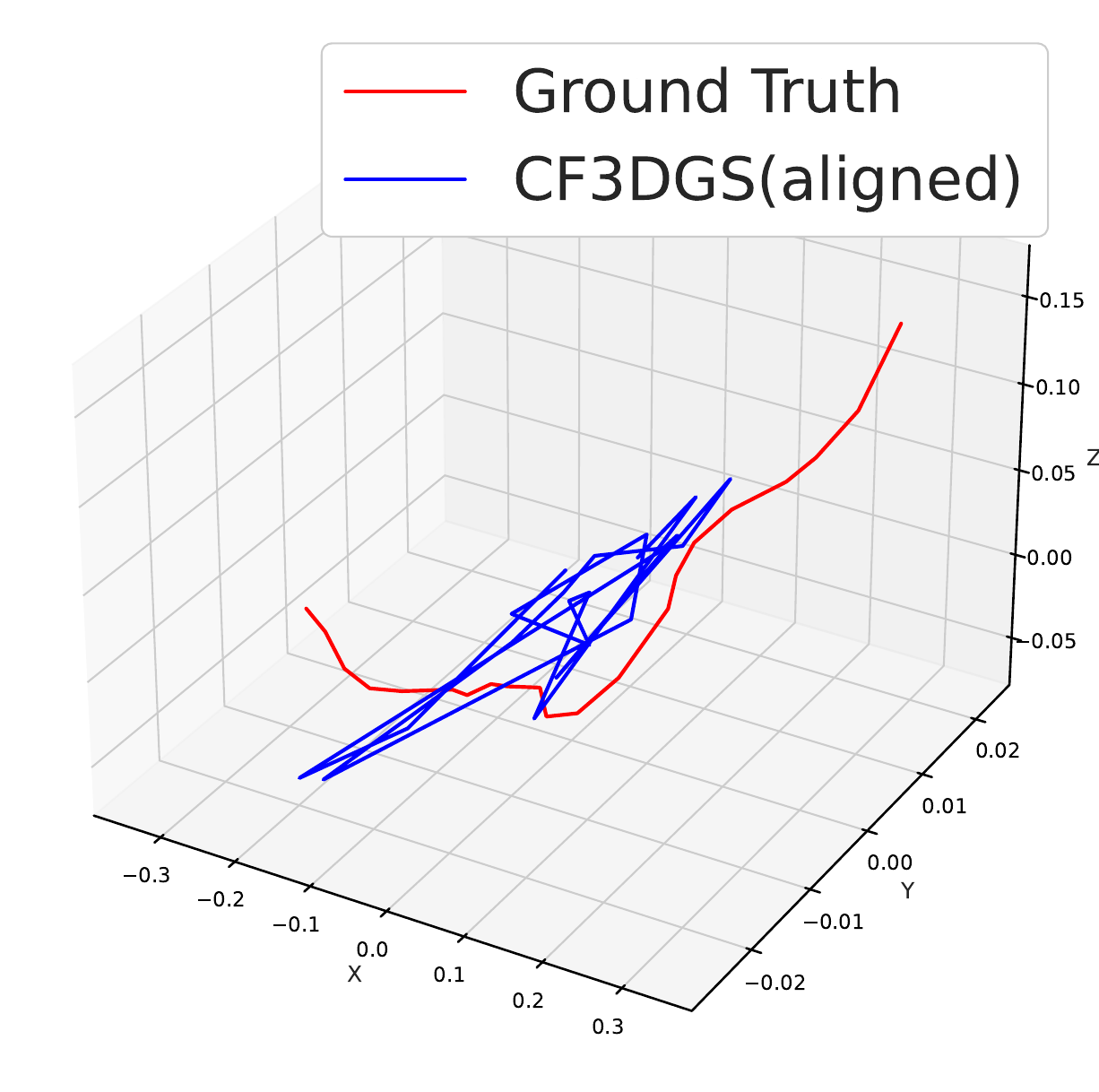} \\

% 第二行：GSHT
\multirow{3}{*}[1.1cm]{\rotatebox{90}{\textbf{\small GSHT}}} &
\includegraphics[width=0.155\linewidth]{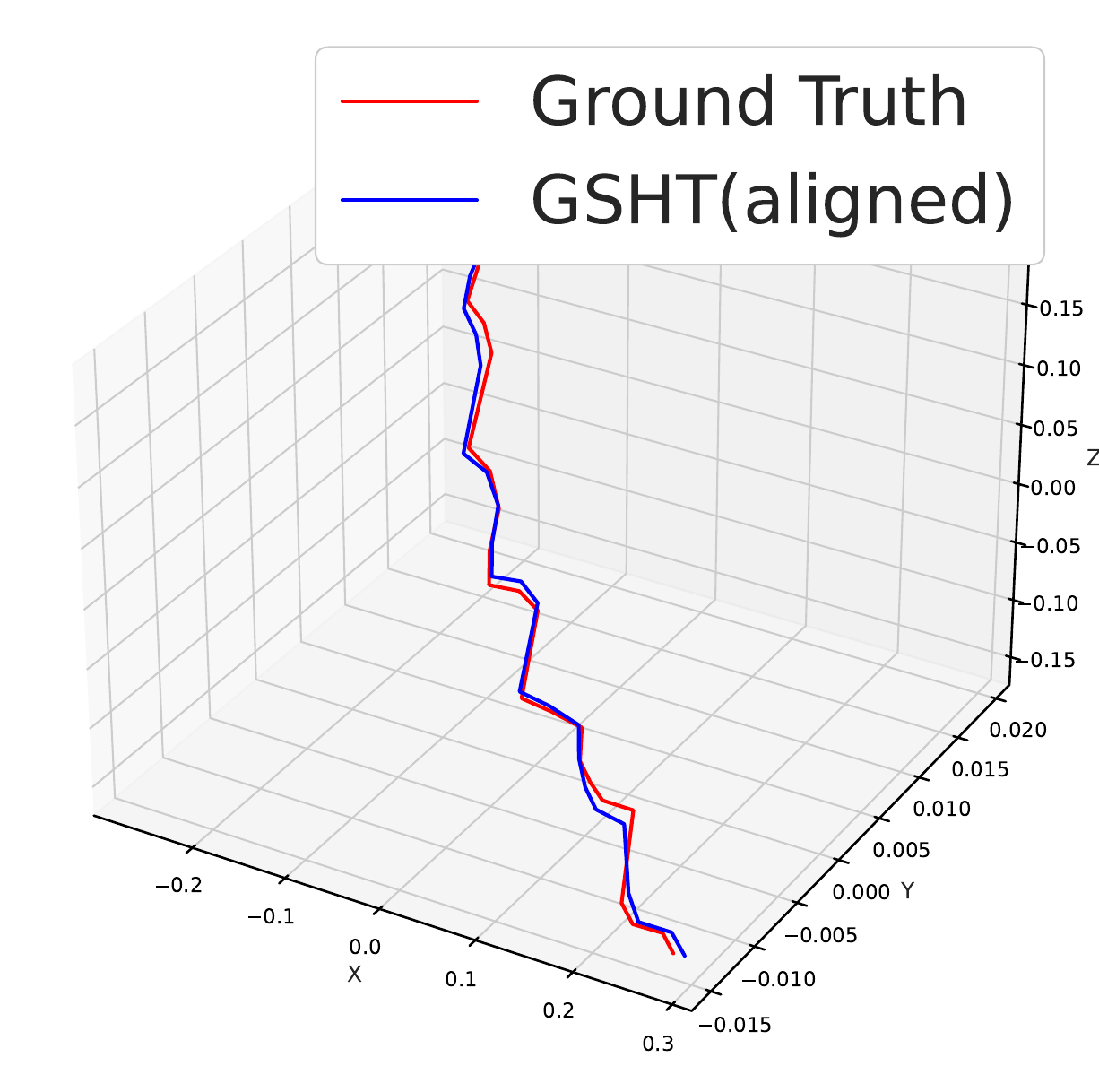} &
\includegraphics[width=0.155\linewidth]{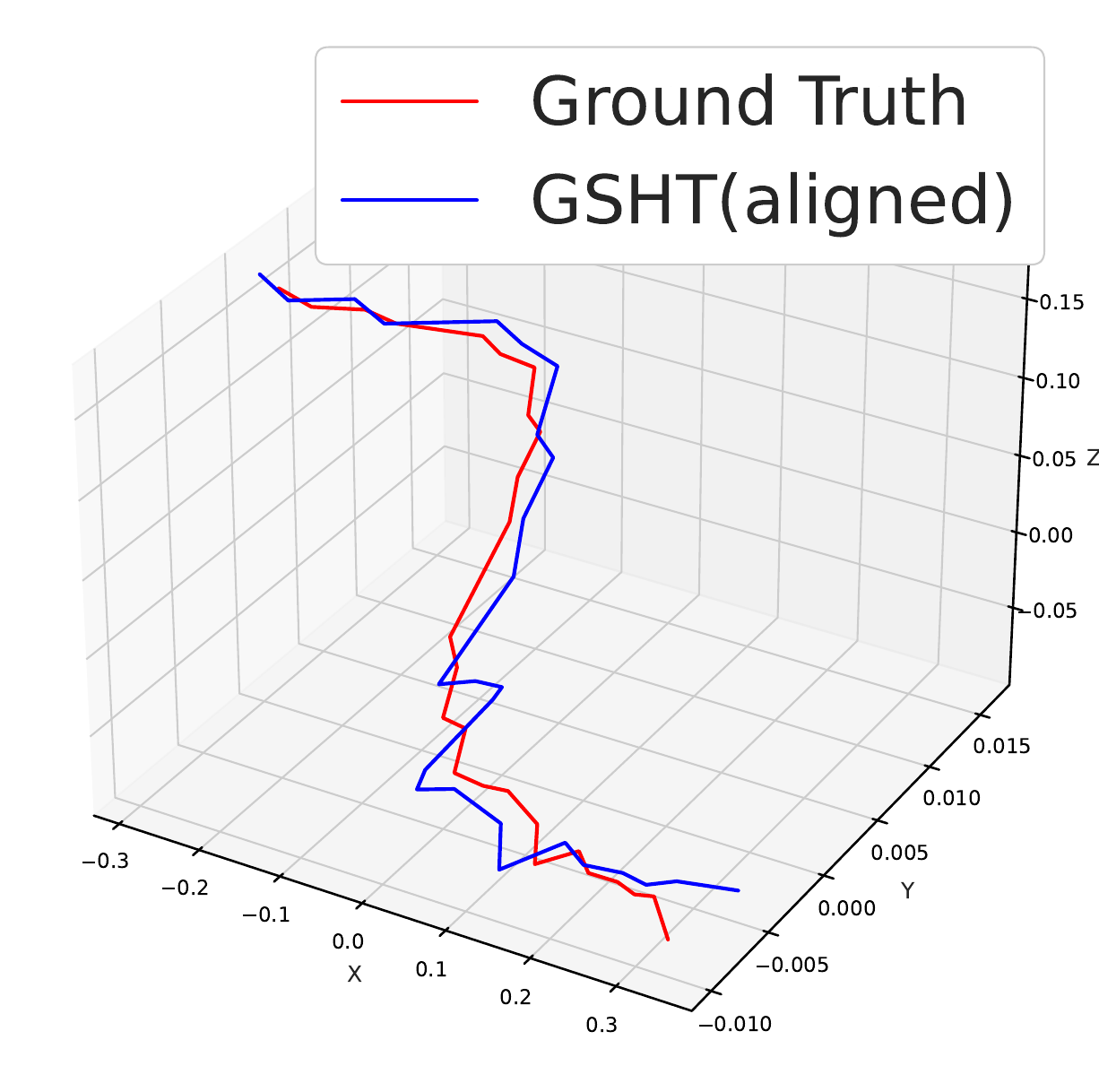} &
\includegraphics[width=0.155\linewidth]{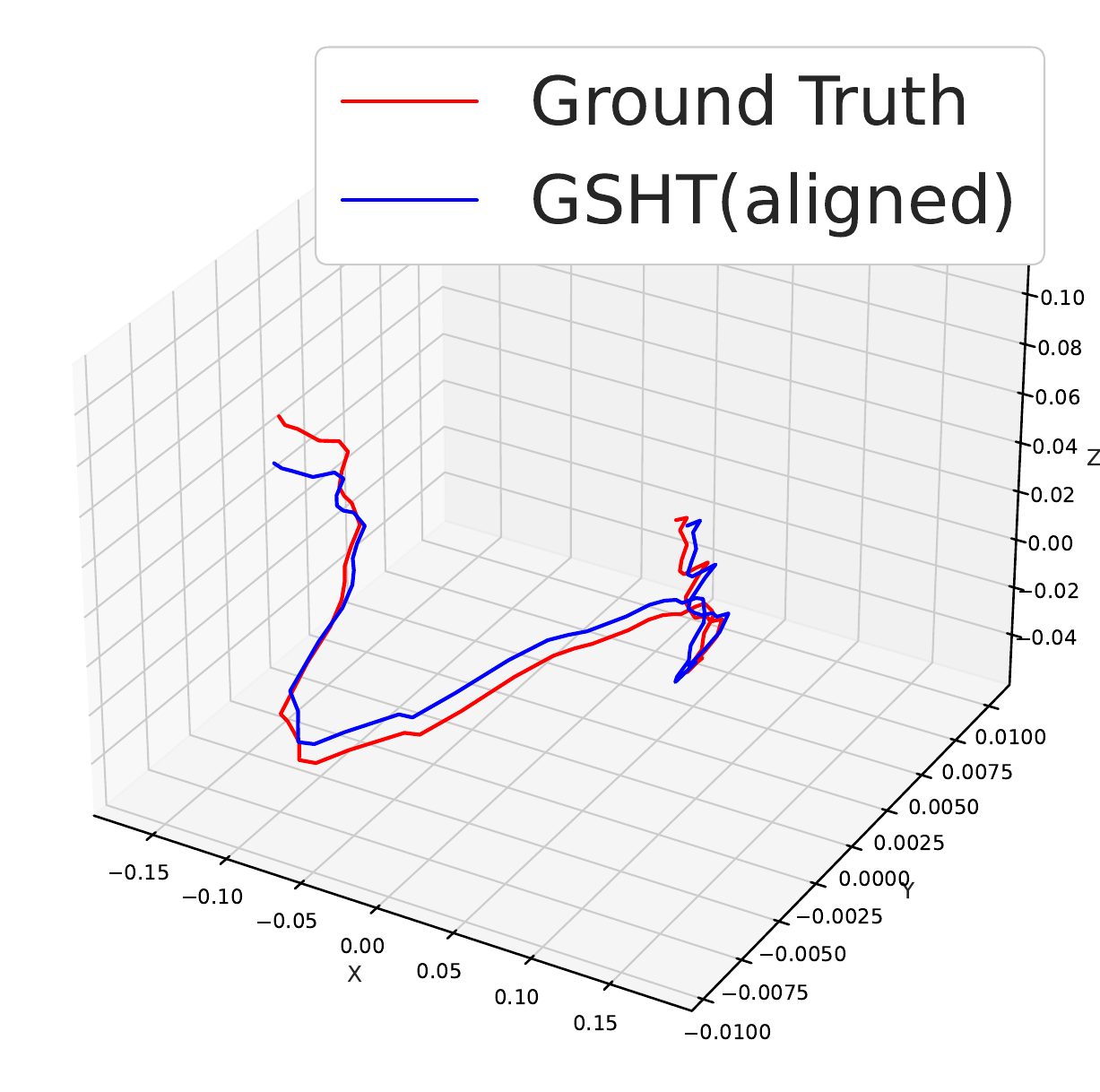} &
\includegraphics[width=0.155\linewidth]{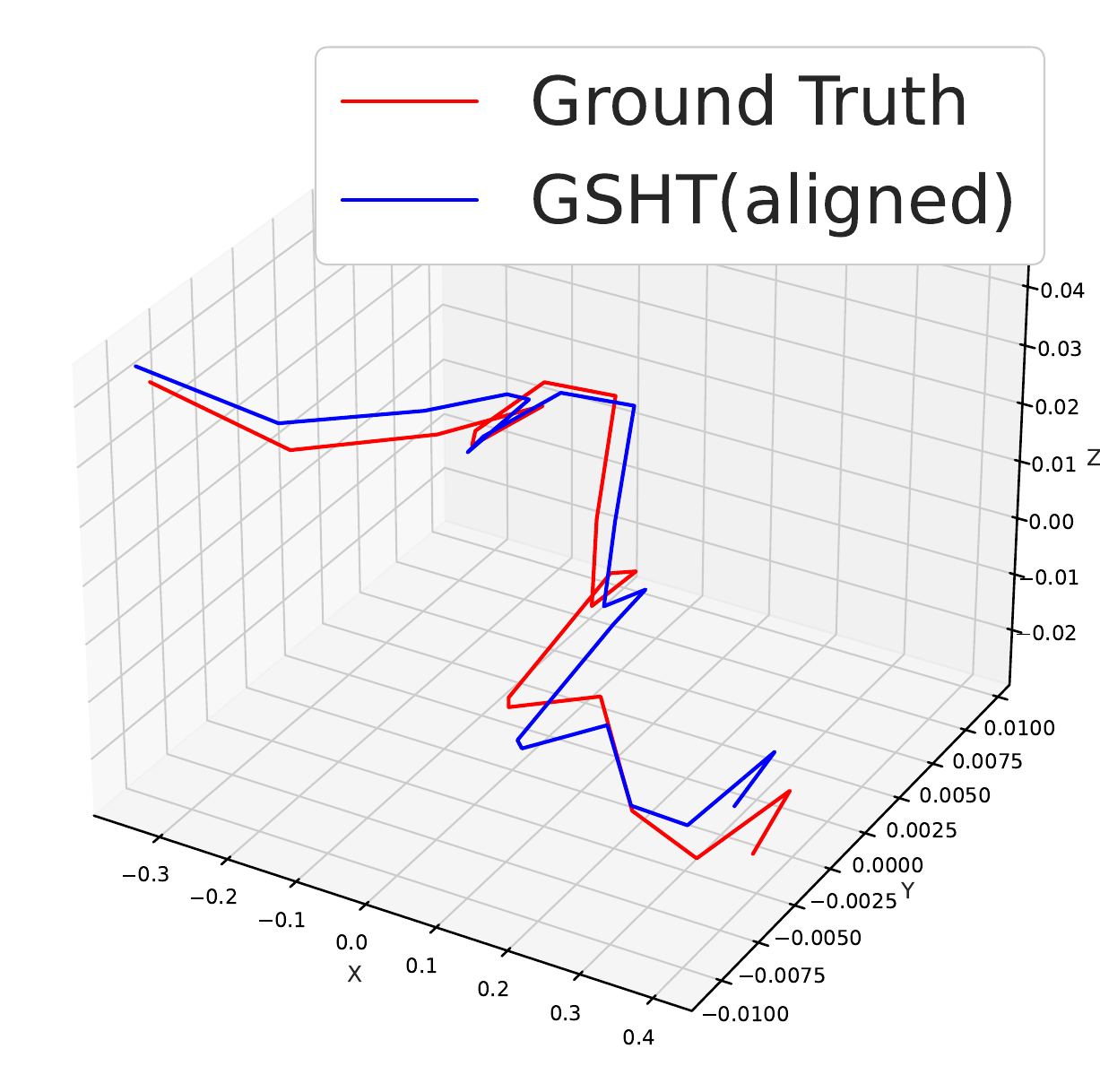} &
\includegraphics[width=0.155\linewidth]{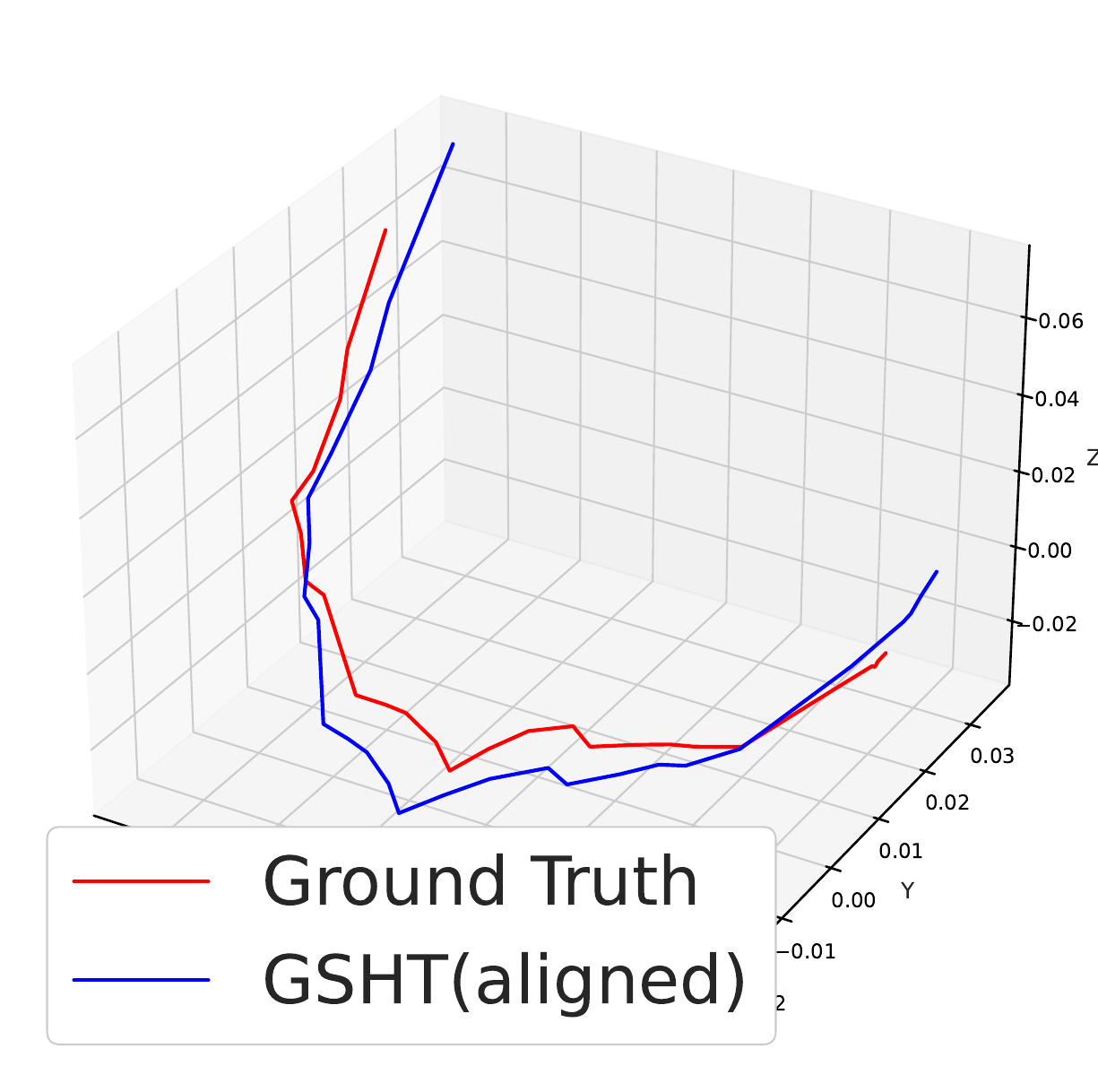} &
\includegraphics[width=0.155\linewidth]{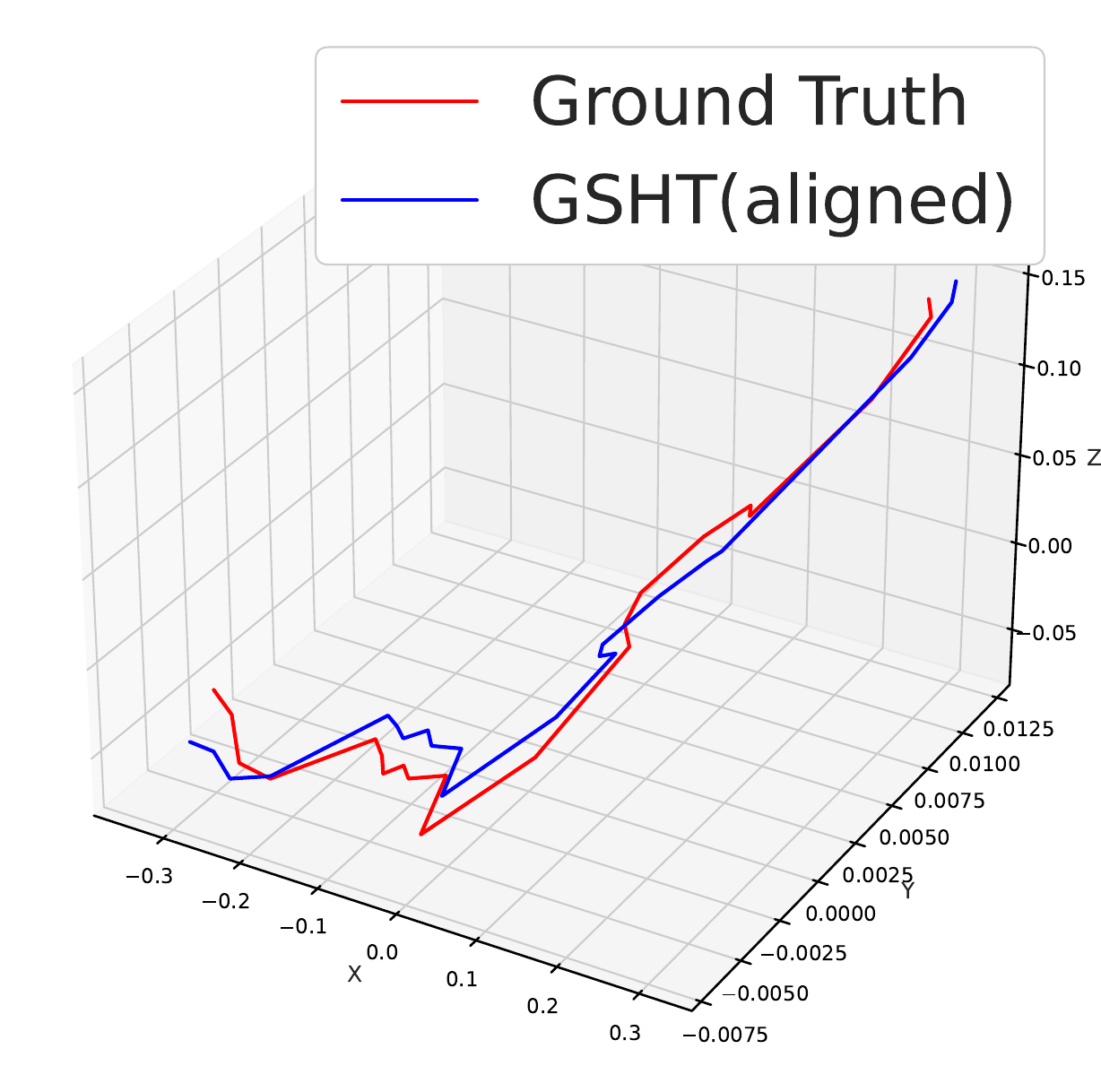} \\

% 第三行：Ours (注意路径根据你之前的代码可能是 JOGS)
\multirow{3}{*}[1.1cm]{\rotatebox{90}{\textbf{\small Ours}}} &
\includegraphics[width=0.155\linewidth]{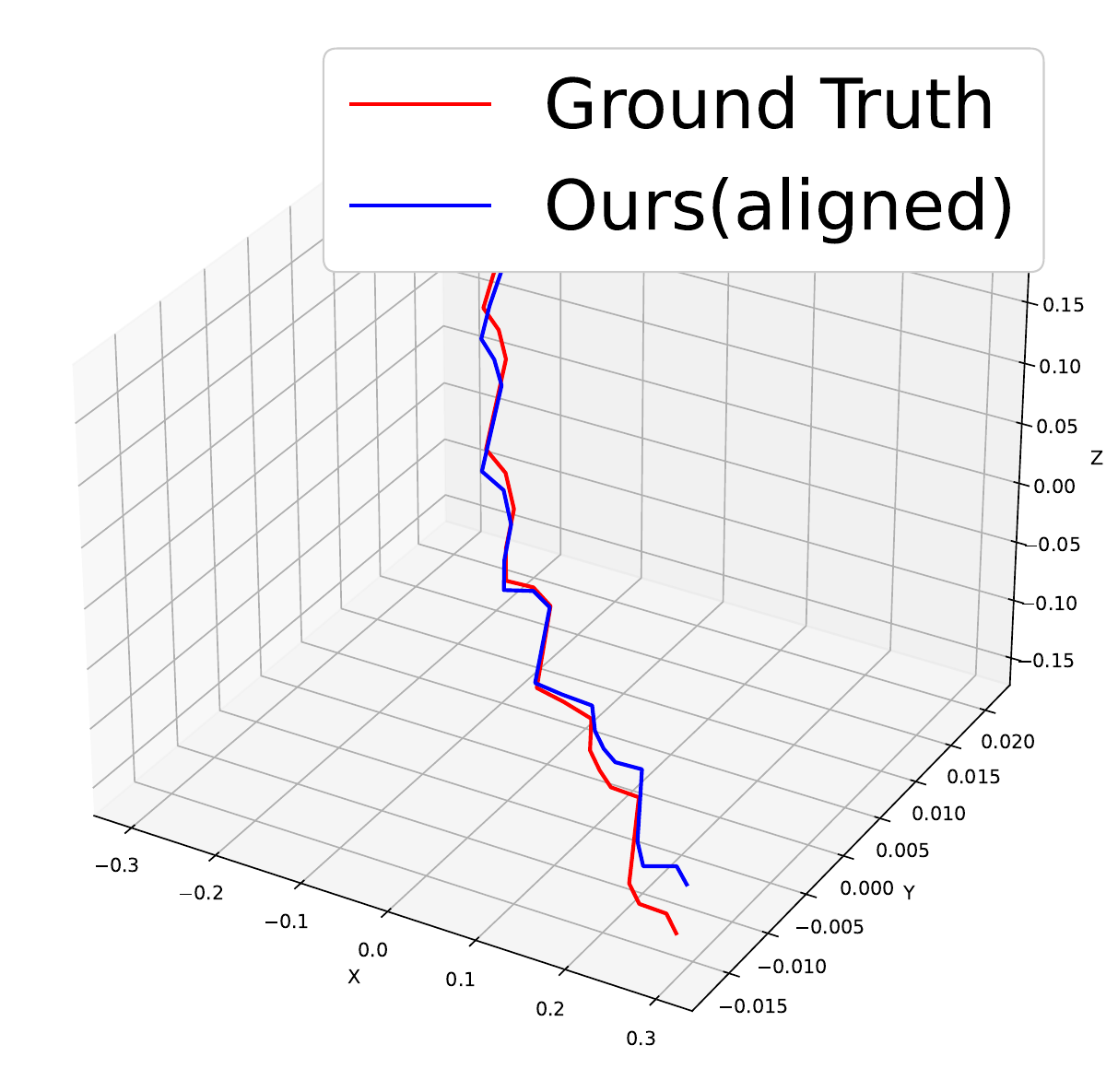} &
\includegraphics[width=0.155\linewidth]{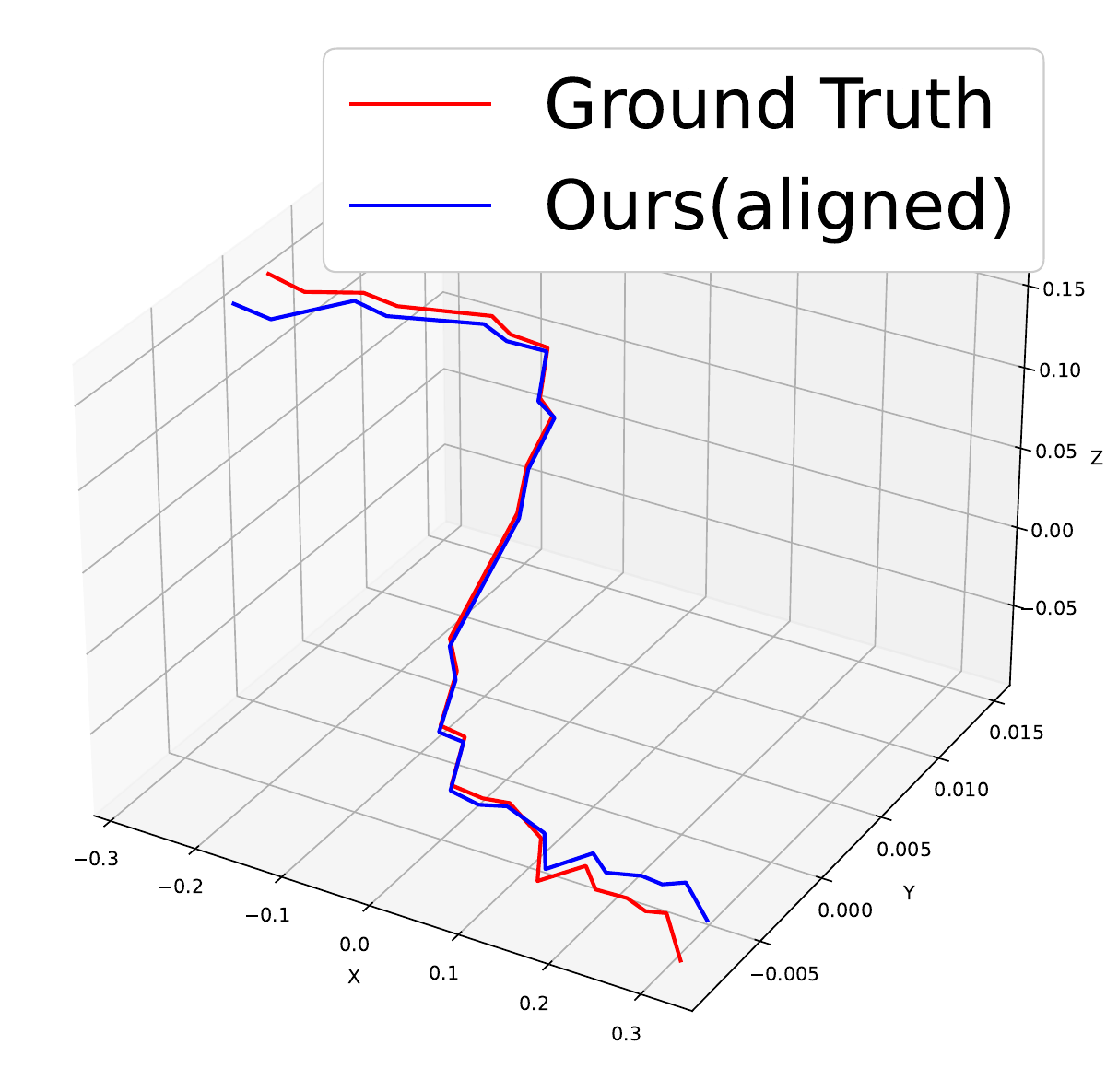} &
\includegraphics[width=0.155\linewidth]{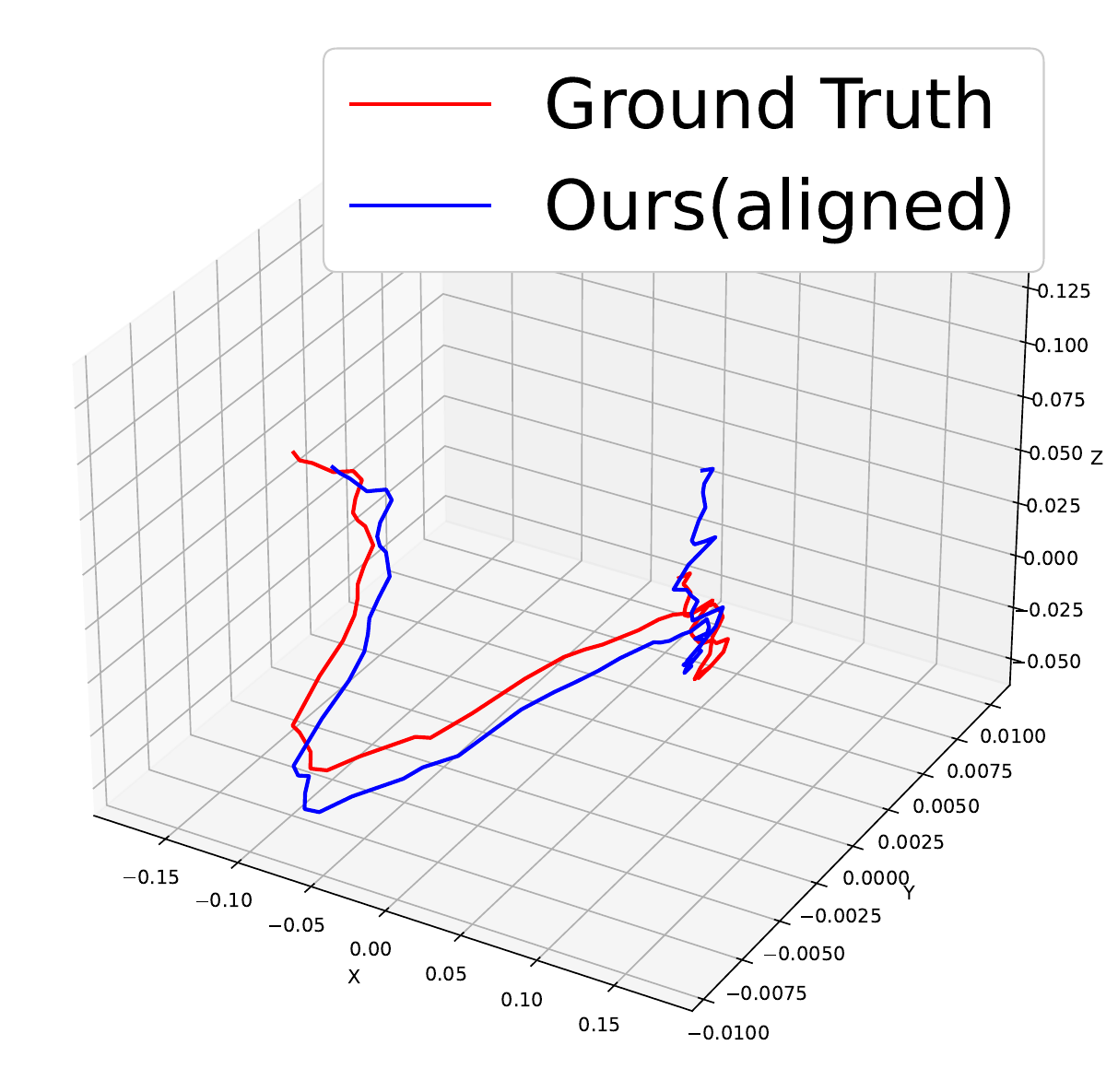} &
\includegraphics[width=0.155\linewidth]{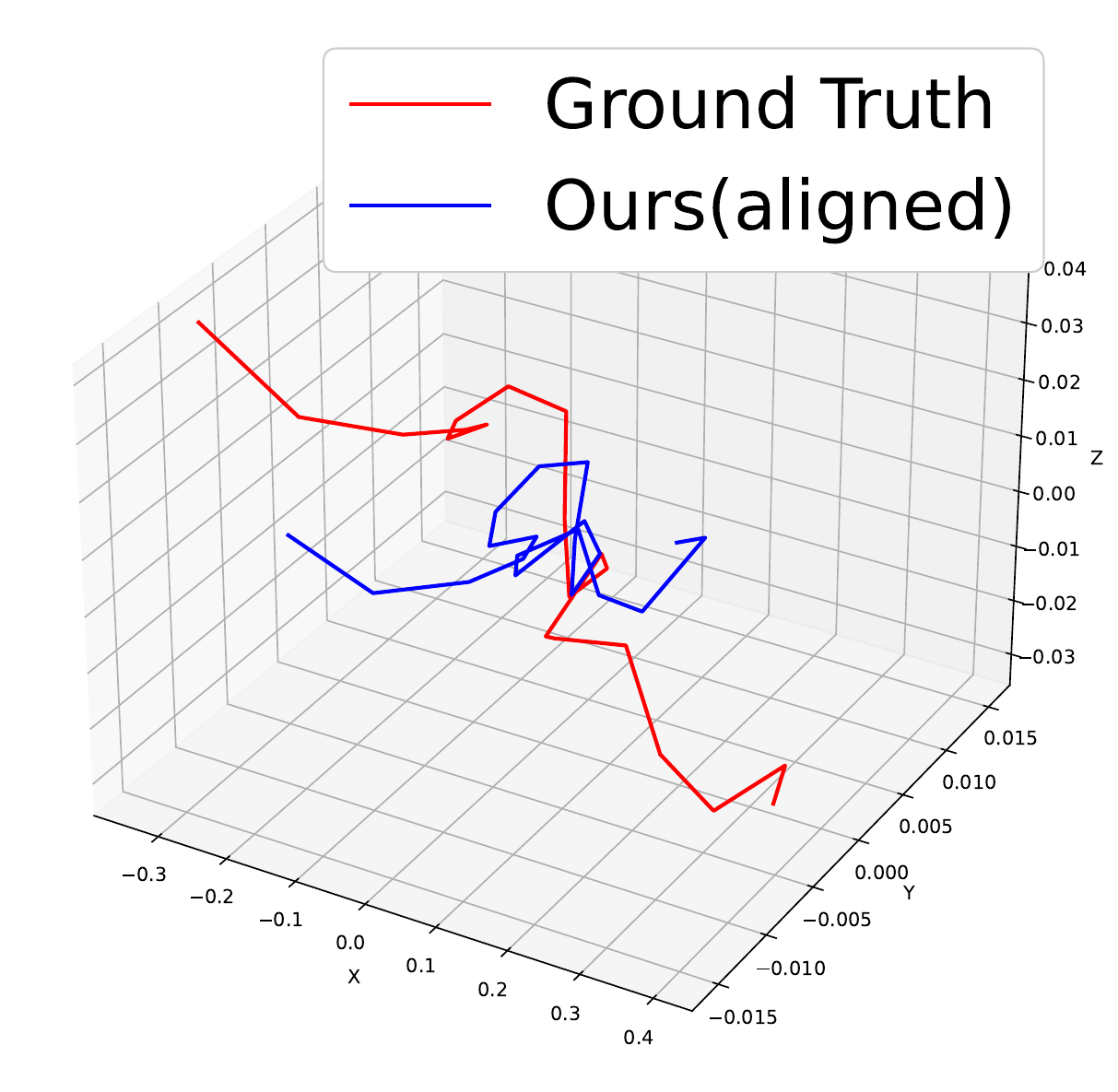} &
\includegraphics[width=0.155\linewidth]{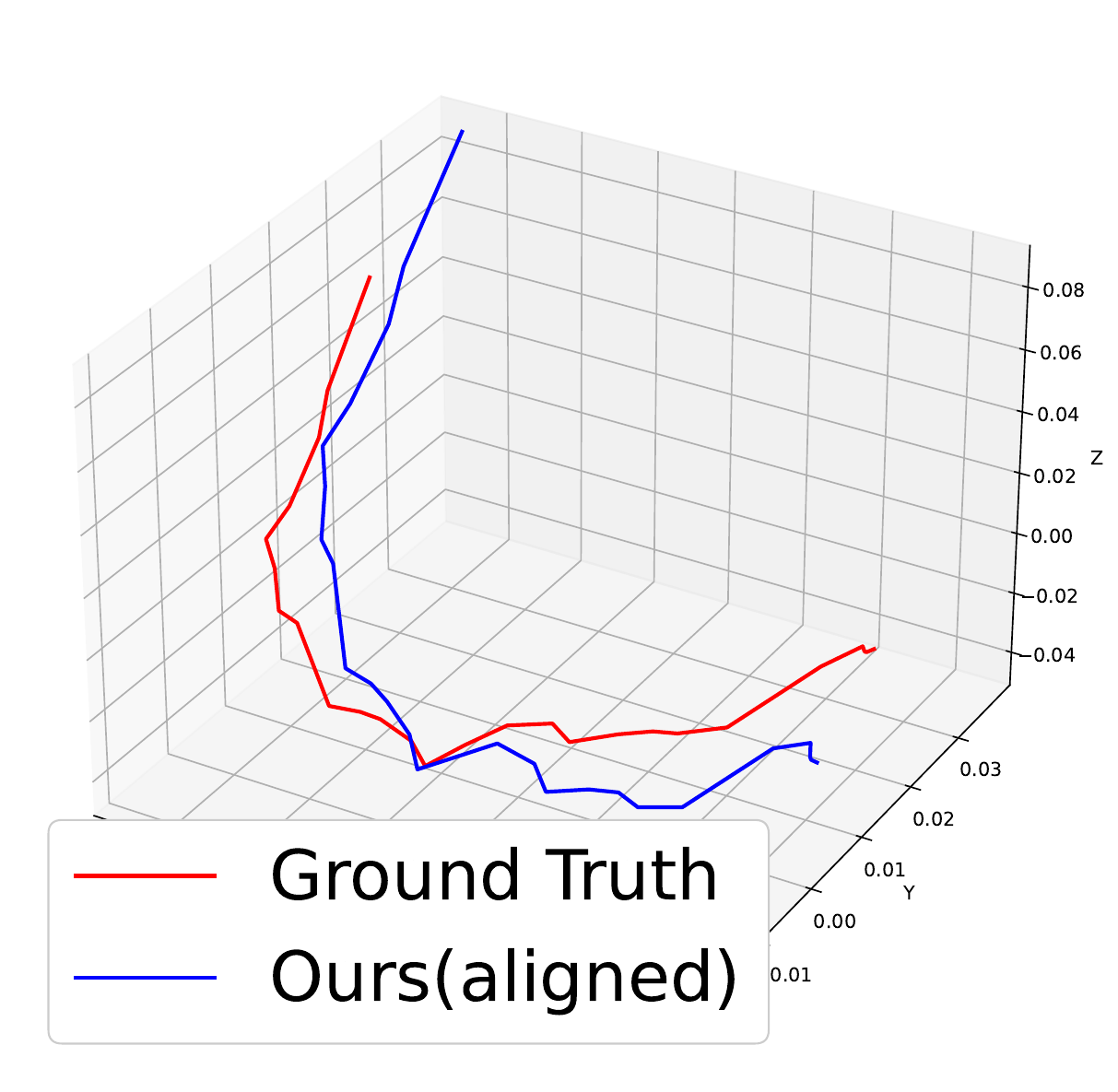} &
\includegraphics[width=0.155\linewidth]{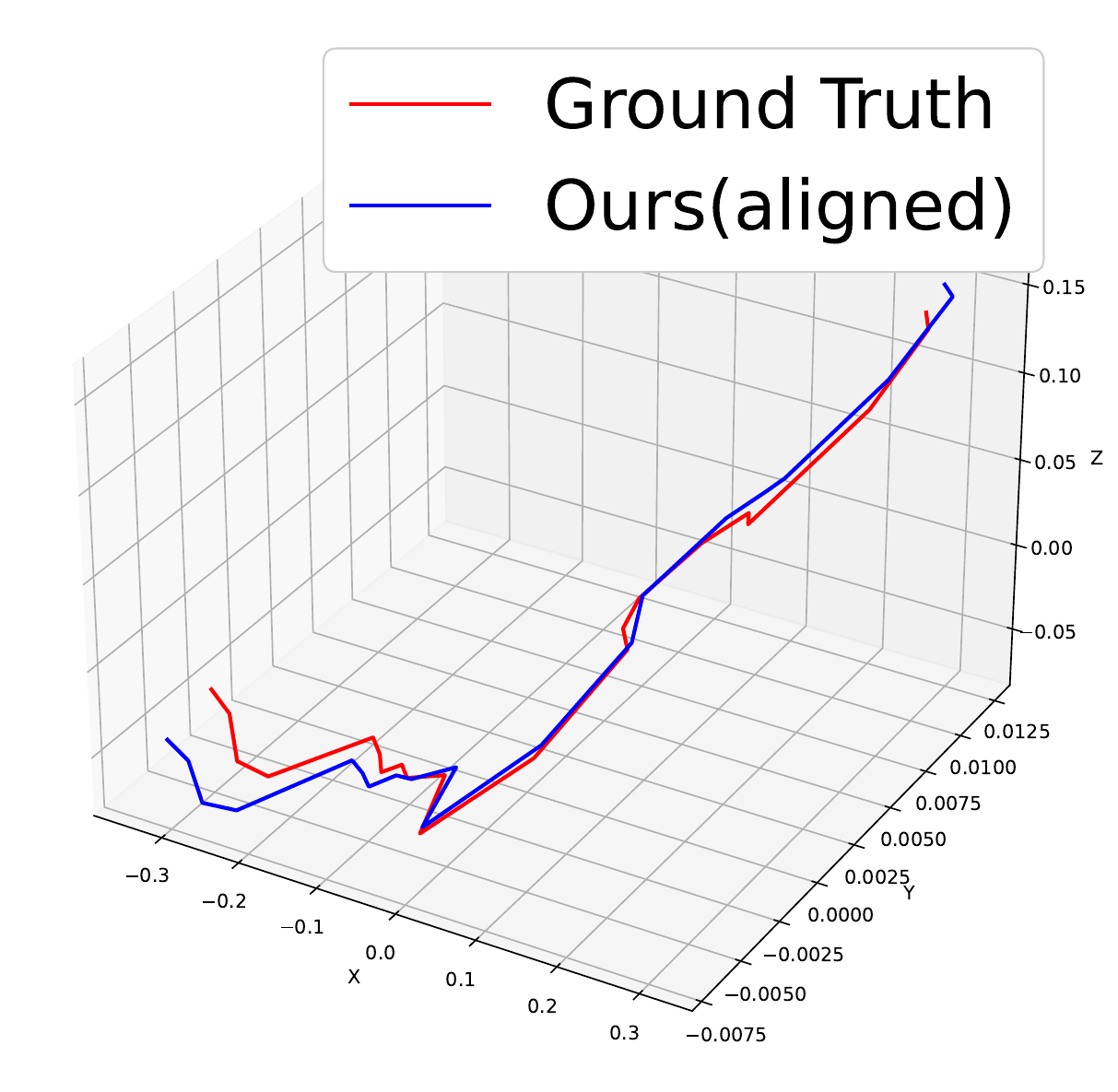} \\

& \scriptsize Ballroom & \scriptsize Barn & \scriptsize Church & \scriptsize Family & \scriptsize Francis & \scriptsize Horse \\
\end{tabular}

\caption{Trajectory comparison on the Tanks and Temples dataset. Each row represents a method (CFGS, GSHT and Ours), and each column represents a different scene.}
\label{fig:trajectory_all}
\vspace{-1em}
\end{figure*}

\begin{table*}[t] % 使用 table* 环境使其横跨两栏，[t] 表示放在页面顶部
\centering
\small
\renewcommand{\arraystretch}{1.2} % 稍微增加行间距，提升可读性
\setlength{\tabcolsep}{10pt} % 增加列间距，因为横跨两栏空间更充裕
\captionsetup{font=small}
\caption{\textbf{Pose estimation performance comparison} on Tanks and Temples dataset.}
\label{tab:pose_comparison_tant}

\begin{tabular}{l|ccc|ccc|ccc} % 将第一列改为左对齐（l），视觉上更整齐
\toprule[1.2pt]
\multirow{2}{*}{\textbf{Scene}} &
\multicolumn{3}{c|}{\textbf{RPE\textsubscript{trans} ↓}} &
\multicolumn{3}{c|}{\textbf{RPE\textsubscript{rot} ↓}} &  
\multicolumn{3}{c}{\textbf{ATE ↓}} \\
\cmidrule(lr){2-4} \cmidrule(lr){5-7} \cmidrule(l){8-10}
& CFGS & GSHT & Ours & CFGS & GSHT & Ours & CFGS & GSHT & Ours \\
\midrule[0.8pt]
Ballroom  & 2.759 & \underline{0.306} & \textbf{0.126} & 3.374 & \underline{0.076} & \textbf{0.035} & 0.196 & \textbf{0.004} & \underline{0.040} \\
Barn      & 6.915 & \underline{1.007} & \textbf{0.153} & 7.216 & \underline{0.202} & \textbf{0.063} & 0.190 & \underline{0.025} & \textbf{0.005} \\
Church    & 1.892 & \textbf{0.070} & \underline{0.110} & 12.061 & \underline{0.065} & \textbf{0.049} & 0.119 & \underline{0.006} & \textbf{0.018} \\
Family    & 1.838 & \underline{0.484} & \textbf{0.127} & 7.192 & \underline{0.126} & \textbf{0.030} & 0.169 & \textbf{0.007} & \underline{0.033} \\
Francis   & 4.141 & \underline{0.276} & \textbf{0.102} & 6.112 & \underline{0.566} & \textbf{0.057} & 0.194 & \underline{0.011} & \textbf{0.016} \\
Horse     & 8.963 & \underline{0.789} & \textbf{0.192} & 7.140 & \underline{0.159} & \textbf{0.026} & 0.205 & \underline{0.009} & \textbf{0.005} \\
Ignatius  & 8.785 & \underline{0.345} & \textbf{0.174} & 7.381 & \underline{0.059} & \textbf{0.052} & 0.206 & \textbf{0.011} & \underline{0.034} \\
Museum    & 8.224 & \underline{3.418} & \textbf{0.232} & 4.835 & \underline{2.912} & \textbf{0.039} & 0.227 & \textbf{0.057} & \underline{0.168} \\
\midrule[0.8pt]
Mean      & 5.440 & \underline{0.837} & \textbf{0.152} & 6.914 & \underline{0.521} & \textbf{0.044} & 0.188 & \textbf{0.016} & \underline{0.040} \\
\bottomrule[1.2pt]
\end{tabular}
\end{table*}
\subsection{Comparing with Baseline}
Our experimental framework is built upon the original 3DGS architecture~\cite{10.1145/3592433}. While the proposed modules are theoretically compatible with advanced 3DGS variants, our current implementation specifically adheres to the canonical formulation due to two methodological considerations: (1) make sure that the comparison with the original 3DGS can be made directly so that future generations can easily reproduce our work; and (2) isolating the performance impact of our contributions from other confounding factors. To maintain consistency, all architectural parameters strictly follow the original 3DGS configuration. This design choice facilitates direct comparability with COLMAP-based 3DGS baselines under identical experimental protocols.

For the COLMAP-free methods, NeRF-based approaches exhibit significantly longer training times and performance gaps compared to 3DGS variants, so we exclude them from comparison.   Our quantitative and qualitative comparisons emphasize Ground Truth, the proposed JOGS, 3DGS~\cite{10.1145/3592433}, CFGS~\cite{Fu_2024_CVPR} and GSHT~\cite{ji2024sfmfree3dgaussiansplatting}, of which the last two are also COLMAP-free methods.

\begin{table}[htbp]
\centering
\footnotesize
\renewcommand{\arraystretch}{1.0}  
\setlength{\tabcolsep}{5pt} 
\captionsetup{font=small}
\caption{\textbf{Pose estimation performance comparison} between our method and VGGT on the LLFF dataset.}
\resizebox{0.5\textwidth}{!}{
\captionsetup{skip=2pt}            
\begin{tabular}{c|cc|cc|cc}
\toprule[1.2pt]
\multirow{2}{*}{Scene } & 
\multicolumn{2}{c|}{RPE\textsubscript{trans} ↓} & 
\multicolumn{2}{c|}{RPE\textsubscript{rot} ↓} & 
\multicolumn{2}{c}{{ATE ↓}} \\
\cmidrule(lr){2-3} \cmidrule(lr){4-5} \cmidrule(l){6-7} 
& VGGT & Ours & VGGT & Ours & VGGT & Ours \\
\midrule[0.8pt]
\addlinespace[2pt]  % 减小行间距
Fern   & 0.285  & \textbf{0.146}  & 0.071&  \textbf{0.039}  & 0.038&\textbf{0.014}   \\
Flowers& 0.160& \textbf{0.100}  & \textbf{0.036}& 0.052  & \textbf{0.004} &0.005  \\
Horns      & 0.056& \textbf{0.051}  & \textbf{0.009} & 0.027 & \textbf{0.010} & 0.019 \\
Orchids       & 0.312& \textbf{0.135}  & \textbf{0.087}& 0.117  & 0.025& \textbf{0.018}  \\
Room       & 0.104&   \textbf{0.039}& 0.035&  \textbf{0.030}   & 0.017& \textbf{0.005}  \\
Trex      & 0.138&\textbf{0.084}   & 0.032&  \textbf{0.026}  & 0.011&  \textbf{0.006} \\
\midrule[0.8pt]

Mean      & {0.203}& \textbf{0.093}  &{0.059}& \textbf{0.049}  & {0.017}& \textbf{0.011}  \\
\bottomrule[1.2pt]
\end{tabular}
}
\vspace{0.5em}  
\label{table:vggt_pose}
\vspace{-1.5em}
\end{table}

\subsection{Novel View Synthesis Evaluation}
As shown in Tab.~\ref{table:nvs_tank_sub}, \ref{table:nvs_llff} and \ref{table:nvs_shiny}, both CFGS and GSHT suffer degraded reconstruction quality, primarily due to their reliance on temporal continuity—when pose changes become large (e.g., under frame-subsampling on Tanks and Temples), their performance deteriorates sharply, as shown by the sharp drop in PSNR. In contrast, our method is sequence-agnostic and remains robust even under aggressive subsampling. 
As illustrated in Fig.~\ref{fig:nvs_all_2}, our method generates sharper geometric features and more coherent textures, in contrast to the blurred reconstructions of CFGS and the fragmented surfaces of GSHT. Beyond holistic visual assessment, Fig.~\ref{fig:detail_compare} presents fine-grained comparisons of structural details.  Our method achieves superior fidelity in geometric preservation and texture reconstruction compared to baseline approaches.

In addition, it is noteworthy that in Fig.~\ref{fig:nvs_all_2}, the CFGS method produces noticeably blurred novel view synthesis images due to its inaccurate pose estimation. This issue becomes particularly pronounced in the detailed regions as illustrated in Fig.~\ref{fig:detail_compare}. As demonstrated in the figure, both CFGS and GSHT exhibitscale drift and spatial misalignment of the display units.
As shown in Fig.~\ref{fig:detail_compare} (the first row), 3DGS shows obvious blurring around high-frequency structures such as the edge of the display. This is primarily due to the lack of joint camera pose optimization during training, where even slight pose inaccuracies can be amplified during dense rendering, leading to structural blur and color artifacts.
We evaluate on the Shiny dataset, which features strong reflections and refractions. As shown in Tab.~\ref{table:nvs_shiny}, JOGS matches COLMAP+3DGS in overall metrics and significantly outperforms both COLMAP-free baselines. 
\subsection{Camera Pose Estimation}
% In Tab.~\ref{tab:pose_comparison}, we provide a quantitative comparison of camera pose estimation on the LLFF dataset. The estimated camera poses are first aligned in scale with the ground truth, following the alignment strategy proposed in GSHT, and evaluated in terms of ATE and RPE.
% As shown in Fig.~\ref{fig:pose_comparison}, our method produces significantly more accurate results than both CFGS and GSHT, demonstrating its effectiveness in reducing both relative and absolute pose errors.
In Tab.~\ref{tab:pose_comparison} and Tab.~\ref{tab:pose_comparison_tant}, we provide a quantitative comparison of camera pose estimation performance on the LLFF and Tanks and Temples datasets, respectively. Following the alignment strategy proposed in GSHT, the estimated camera trajectories are first aligned with the ground truth in scale before evaluation. We report the results in terms of ATE and RPE.
As shown in these results, our method achieves the lowest errors on most scenes. Qualitative results are visualized in Fig.~\ref{fig:pose_comparison} and Fig.~\ref{fig:trajectory_all}. Our method produces estimated trajectories that are significantly more consistent with the ground truth than the baselines, demonstrating its robustness and effectiveness in handling complex camera motions and large-scale environments.

Furthermore, we compare our method with VGGT~\cite{vggt}, a feed-forward neural network that reports strong performance across multiple 3D tasks, including camera parameter estimation. As shown in Table~\ref{table:vggt_pose}, on the LLFF dataset, our method achieves lower ATE and RPE on most scenes, with a lower mean error overall.

\subsection{Ablation Study}
To validate the necessity of our joint optimization framework, we conduct an ablation study comparing two variants: (1) Initialization-only (using initialized poses without iterative refinement during training) and (2) Full method (with alternating Gaussian and pose optimization).  

As shown in Tab.~\ref{table:ablation_dataset}, the full method outperforms the reduced initialization-only variant across all datasets.  \textit{Init} means working with only pose initialization without iterative refinement, while \textit{Full} means working with the full version of our method containing joint optimization of Gaussian points and camera poses.
The ablation study  demonstrates that our combined optimization framework effectively mitigates error accumulation and enhances the synthesis accuracy of novel view scenes by alternately updating Gaussian points and refining camera poses using LK3D.
\begin{table}[htbp]
\centering
\footnotesize
\renewcommand{\arraystretch}{1.0}  
\setlength{\tabcolsep}{5pt} 
\captionsetup{font=small}
\caption{\textbf{Ablation study} of joint optimization across three benchmark datasets.}
\resizebox{0.5\textwidth}{!}{
\captionsetup{skip=2pt}            
\begin{tabular}{c|cc|cc|cc}
\toprule[1.2pt]
\multirow{2}{*}{Dataset} & 
\multicolumn{2}{c|}{PSNR$\uparrow$} & 
\multicolumn{2}{c|}{SSIM$\uparrow$} & 
\multicolumn{2}{c}{LPIPS$\downarrow$} \\
\cmidrule(lr){2-3} \cmidrule(lr){4-5} \cmidrule(l){6-7} 
& Init & Full & Init & Full & Init & Full \\
\midrule[0.8pt]
\addlinespace[2pt]  % 减小行间距
LLFF-NeRF    & 25.25 & \textbf{25.39} & \textbf{0.81} & \textbf{0.81} & \textbf{0.20} & \textbf{0.20} \\
Tanks and Temples & 25.94 & \textbf{26.91} & 0.86 & \textbf{0.88} & 0.14 & \textbf{0.13} \\
Shiny        & 24.98 & \textbf{25.58} & 0.80 & \textbf{0.85} & 0.23 & \textbf{0.21} \\
\midrule[0.8pt]
Mean         & 25.39 & \textbf{25.96} & 0.82 & \textbf{0.85} & 0.19 & \textbf{0.18} \\
\bottomrule[1.2pt]
\end{tabular}
}
\vspace{0.5em}  
\label{table:ablation_dataset}
\vspace{-1.5em}
\end{table}

\section{Conclusion}
\label{sec:conclusion}

In this paper, we introduce a novel view synthesis framework that jointly optimize pose estimation and 3DGS, without requiring camera poses as inputs.  This framework outperforms state-of-the-art methods in both pose estimation accuracy and rendering quality, particularly under challenging conditions, by leveraging an alternating optimization strategy for 3D Gaussian representations and camera poses.

\vspace{1mm}\noindent\textbf{Limitations.} Despite the effectiveness in both camera pose estimation and rendering quality, our method requires an increased training time due to the increased pose refinement operation. We plan to address this issue by exploring parallel optimization strategy in the future work.

\section*{Acknowledgments}
This work is supported by the National Natural Science Foundation of China (No. 62376020).

\bibliographystyle{IEEEtran}
\bibliography{main.bib}

\end{document}